\documentclass{article}

\usepackage{microtype}
\usepackage{graphicx}
\usepackage{subfigure}
\usepackage{subcaption}
\usepackage{booktabs} 

\usepackage{hyperref}

\usepackage{tcolorbox}

\usepackage[accepted]{icml2026}

\usepackage{amsmath}
\usepackage{amssymb}
\usepackage{mathtools}
\usepackage{amsthm}

\def\R{\mathbb{R}}
\def\E{\mathbb{E}}

\def\F{{\rm F}}

\newcommand{\tr}{\mbox{tr}}

\usepackage[capitalize,noabbrev]{cleveref}

\theoremstyle{plain}
\newtheorem{theorem}{Theorem}[section]

\newtheorem{lemma}[theorem]{Lemma}

\theoremstyle{definition}
\newtheorem{definition}[theorem]{Definition}
\newtheorem{assumption}[theorem]{Assumption}
\theoremstyle{remark}
\newtheorem{remark}[theorem]{Remark}

\usepackage[textsize=tiny]{todonotes}

\icmltitlerunning{LiMuon: Light and Fast Muon Optimizer for Large Models}

\begin{document}

\twocolumn[
  \icmltitle{LiMuon: Light and Fast Muon Optimizer for Large Models}



  \icmlsetsymbol{equal}{*}

  \begin{icmlauthorlist}
    \icmlauthor{Feihu Huang}{1,2}
    \icmlauthor{Yuning Luo}{3}
    \icmlauthor{Songcan Chen}{1,2}
  \end{icmlauthorlist}

  \icmlaffiliation{1}{College of Computer Science and Technology,	Nanjing University of Aeronautics and Astronautics, Nanjing, China}
  \icmlaffiliation{2}{MIIT Key Laboratory of Pattern Analysis and Machine Intelligence, Nanjing, China}
  \icmlaffiliation{3}{College of Design and Engineering, National University of Singapore, Singapore}

  \icmlcorrespondingauthor{Feihu Huang}{huangfeihu2018@gmail.com}

  \icmlkeywords{Machine Learning, ICML}

  \vskip 0.3in
]



\printAffiliationsAndNotice{}  

\begin{abstract}
  Large models recently are widely applied in machine learning, so efficient training of large models has received widespread attention. More recently, the useful Muon optimizer is specifically designed for matrix-structured parameters of large models. Although some works have begun to study the Muon optimizer, the existing Muon and its variants still suffer from high sample complexity or high memory for large models.
To fill this gap, we propose a light and fast Muon (LiMuon) optimizer
for training large models, which builds on the momentum-based variance reduced technique and randomized Singular Value Decomposition (SVD). In particular, our LiMuon simultaneously has a lower memory and
lower sample complexity than the Muon and its variants. Moreover, we prove that our LiMuon with lower memory has a lower sample complexity of $O(\epsilon^{-3})$ for finding an $\epsilon$-stationary solution of non-convex stochastic optimization under the generalized smoothness condition. To further narrow practice and theory gap, we also prove that our LiMuon with Newton-Schulz steps has a lower sample complexity than the Muon with Newton-Schulz steps. 
 Numerical experimental results on
pre-training Mamba-130M, Qwen2.5-0.5B and ViT models demonstrate effectiveness of our LiMuon.
\end{abstract}

\section{Introduction}
Large models are widely used in many machine learning tasks such as language understanding and generation~\citep{achiam2023gpt,gu2023mamba,liu2024deepseek}, which consists of a massive amount of parameters. Since training large models requires significant computational
 resources, designing efficient optimizers for training such models has received widespread attention.
 Recently, adaptive gradient methods such as Adam~\citep{kingma2014adam}
 and AdamW~\citep{loshchilov2017decoupled} have become standard choice for training large-scale models,
 which use their ability to dynamically adjust learning rates based on the first-order
 and second-order moment estimates.

 Some optimization algorithms~\citep{Tuddenham2022,gupta2018shampoo,jordanmuon,an2025asgo} recently 
have made significant progress by
taking advantage of structures in large models, which have shown promising potential to outperform
these standard adaptive gradient methods. In fact, most of the parameters in neural networks are naturally
represented as matrices or tensors. For example, the matrix-structured parameters appear in the query, key,
and value matrices in attention mechanisms~\citep{vaswani2017attention}, and tensor-structured parameters appear in nuclear tensor of convolutional neural networks~\citep{krizhevsky2017imagenet}. A typical noteworthy optimization algorithm is the Muon optimizer~\citep{jordanmuon}, which updates matrix parameters with orthogonalized gradient momentum using Newton-Schulz iteration. It has demonstrated competitive empirical performance across diverse large language models (LLMs)~\citep{liu2025muon}.

\begin{table*}
  \centering
  \caption{ Comparison of the Muon optimizer and its variants  for finding an $\epsilon$-stationary point ($\E\|\nabla F(W)\|_\F \leq \epsilon$) of nonconvex stochastic optimization problem~(\ref{eq:p1}). Here $\hat{r} \ll \min(m,n)$ and $\chi_{q}=\frac{1}{1-\varepsilon_{q}}>1$, where $\varepsilon_{q}\in (0,1)$ is a polar approximation error.}
  \label{tab:1}
   \resizebox{0.99\textwidth}{!}{
\begin{tabular}{c|c|c|c|c|c}
  \hline
  \textbf{Algorithm} & \textbf{Reference} & \textbf{SFO Complexity} & \textbf{ State Memory} &  \textbf{Generalized Smoothness} &  \textbf{Newton-Schulz}
  \\ \hline
  Muon  & \cite{shen2025convergence} & $O(\epsilon^{-4})$  & $mn$ & &  \\   \hline
  SCG  & \cite{pethick2025training} & $O(\epsilon^{-4})$  & $mn$ & & \\  \hline
  Gluon & \cite{riabinin2025gluon} &  $O(\epsilon^{-4})$ & $mn$ & $\surd$ & \\  \hline
  GGNC & \cite{pethick2025generalized} &  $O(\epsilon^{-4})$ & $mn$ & $\surd$ & \\  \hline
  SUMO  & \cite{refael2025sumo} &   $O(\epsilon^{-4})$& $(m+n)\hat{r}$ & & \\  \hline
  Muon$^{++}$  & \cite{sfyraki2025lions} & $O(\epsilon^{-3})$  & $mn$ & & \\  \hline
  LiMuon  & Ours &  \textcolor{red}{$O(\epsilon^{-3})$ } & \textcolor{red}{$(m+n)\hat{r}$} &  \textcolor{red}{$\surd$}  & \\  \hline \hline
  Muon  & \cite{kim2026convergence} & $O(\chi_{q}^4\epsilon^{-4})$  & $mn$ & & $\surd$ \\  \hline
  LiMuon  & Ours &  \textcolor{red}{$O(\chi_{q}^3\epsilon^{-3})$ } & \textcolor{red}{$(m+n)\hat{r}$} &  \textcolor{red}{$\surd$}  & \textcolor{red}{$\surd$} \\  \hline
\end{tabular}
 }
\end{table*}

Recently, some works have begun to studying the convergence properties
of the Muon optimizer relying on an exact SVD-based polar factor. For example,
\cite{shen2025convergence,li2025note} proved that the Muon optimizer has a
Stochastic First-order Oracle (SFO) complexity
of $O(\epsilon^{-4})$ for finding an $\epsilon$-nuclear-norm stationary point of
nonconvex stochastic optimization.
\cite{pethick2025training} studied the stochastic conditional gradient (a.k.a., stochastic Frank-Wolfe)~\citep{hazan2016variance} methods
by leveraging the linear minimization oracle (LMO)
over a norm-ball, which includes the Muon optimizer, and proved that it also has a SFO complexity
of $O(\epsilon^{-4})$ for finding an $\epsilon$-nuclear-norm stationary point.
Further, \cite{sfyraki2025lions} showed that the Muon with weight decay can be viewed as special
 instances of a stochastic Frank-Wolfe algorithm, and established its convergence guarantees
 in terms of the Frank-Wolfe gap, which implies convergence to a KKT point of the Muon under
 a norm constraint for non-convex stochastic optimization. \cite{chen2025muon} provided theoretical results of the Lion-$\mathcal{K}$ to show that the Muon with decoupled weight decay implicitly solves an optimization problem that enforces a constraint on the spectral norm of weight matrices.
 Meanwhile, \cite{kovalev2025understanding} studied the convergence properties of the stochastic non-Euclidean
trust-region gradient method with momentum, which recovers the Muon optimizer as a special case. \cite{lau2025polargrad} showed that the Muon is a class of matrix-gradient optimizers from a unifying preconditioning perspective.

Subsequently, \cite{riabinin2025gluon} studied convergence properties of Muon optimizer under the layer-wise $(L_0,L_1)$-smoothness, which is a variant of $(L_0,L_1)$-smoothness~\citep{zhang2019gradient}.  Meanwhile, \cite{pethick2025generalized} also studied the convergence properties of generalized gradient norm clipping methods under the $(L_0,L_1)$-smoothness. 
To accelerate Muon, \cite{sfyraki2025lions}
proposed a 
variance-reduced Muon (i.e., Muon$^{++}$) algorithm based on the momentum-based variance reduced technique~\citep{cutkosky2019momentum,tran2022hybrid}. To obtain memory-efficient LLM training,
\cite{refael2025sumo} proposed a subspace-aware moment-orthogonalization (SUMO) optimizer based on the Muon. 
So far, although some variants of Muon optimizer have been developed, these methods still suffer from high gradient complexity or high memory in training large models (see Table~\ref{tab:1}). 

 To fill this gap, we propose a light and fast Muon optimizer
for training large models, which builds on the momentum-based variance reduced technique~\citep{cutkosky2019momentum,tran2022hybrid} and randomized Singular Value Decomposition (SVD)~\citep{halko2011finding}.
In summary, our main contributions are as follows:
\begin{itemize}
\item[1)] We propose a light and fast Muon (i.e., LiMuon) optimizer
for the matrix-structured parameters of large models, which \emph{simultaneously} has a lower memory and
lower sample complexity than the Muon optimizer (see Table~\ref{tab:1}).
\item[2)] We provide a solid convergence analysis for our LiMuon optimizer, and prove that our LiMuon obtains a lower SFO (or sample) complexity of $O(\epsilon^{-3})$ with a lower memory for finding an $\epsilon$-stationary point of nonconvex stochastic optimization under the generalized smoothness condition. 
\item[3)] We also prove that our LiMuon with Newton-Schulz steps also obtains a lower SFO complexity of $O(\chi_{q}^3\epsilon^{-3})$ under a lower memory than $O(\chi_{q}^4\epsilon^{-4})$ of Muon with Newton-Schulz~\citep{kim2026convergence}.
\item[4)] We conduct some numerical experiments on pre-training Mamba-130M, Qwen2.5-0.5B and ViT models to verify efficiency of our LiMuon optimizer.
\end{itemize}
From Table~\ref{tab:1}, our LiMuon optimizer can simultaneously obtain a lower SFO complexity and a lower state memory. Note that although the SUMO~\citep{refael2025sumo} also obtains a lower state memory, it relies on a strict condition that the objective function is bounded by some positive constants.

Here we need to emphasize that our LiMuon is \textbf{the first study} of variance-reduced Muon with reduced memory cost. Moreover, we are \textbf{the first} to introduce a novel low-rank approximated STORM-like gradient estimator. In theory, our LiMuon is \textbf{the first study} of convergence properties on variance-reduced Muon under the generalized smoothness condition.
Moreover, our LiMuon also is \textbf{the first study} of convergence properties on variance-reduced Muon with 
Newton-Schulz steps.

\vspace*{-8pt}
\section*{Notation}
For all matrices $A,B\in \R^{m\times n}$, the Frobenius inner product is defined
as $\langle A,B\rangle = \tr(A^\top B)$ and the Frobenius norm is defined as $\|A\|_\F=\sqrt{\tr(A^\top A)}$,
where $\tr(\cdot)$ denotes the trace of a square matrix. 
$\|A\|_*$ and $\|A\|_{op}$ denote the nuclear and spectral (operator) norms of matrix $A$, respectively. $\R^+$ denotes the set of positive real numbers.
For a matrix $A\in \R^{m\times n}$, $\|A\|_*$ denotes its nuclear norm. $A^\top$
denotes the transpose of matrix $A$.
$I$ denotes an identity matrix.
$x_t=O(y_t)$ denotes that $x_t \leq c y_t$ for some constant $c>0$.
\vspace*{-8pt}
\section{Preliminaries}
In the paper, we study the Muon optimizer to solve the following
nonconvex stochastic optimization problem:
\begin{align}\label{eq:p1}
 \min_{W\in \R^{m\times n}} \E_{\xi\sim \mathcal{D}}[f(W;\xi)],
\end{align}
where $F(W)=\E_{\xi\sim \mathcal{D}}[f(W;\xi)]: \R^{m\times n}\rightarrow \R^+$ denotes a loss function
of training large models such as deep neural networks, which generally is nonconvex.
Here $W\in \R^{m\times n}$ denotes parameter matrix, and $\xi$ is a random variable drawn from 
some fixed but unknown data distribution $\mathcal{D}$.

Muon~\citep{jordanmuon} optimizer can be written as: for all $t\geq1$
\begin{align}
 & G_t = \nabla f(W_t;\xi_t),  \nonumber \\
 & B_t = \mu B_{t-1} +G_t,   \nonumber \\
 & O_t = \arg\min \{ \|O-B_t\|_\F: O^\top O=I, \ or \ O O^\top =I \}, \label{eq:svd} \\
 & W_{t+1} = W_t -\eta_t O_t, \nonumber
\end{align}
where momentum parameter $\mu>0$ and learning rate $\eta_t>0$. Here (\ref{eq:svd}) can be equivalently written as $O_t= UV^\top$, where $B_t=U\Sigma V^\top$ is the singular value
decomposition (SVD).
Since SVD is usually expensive in practice, especially when the
 dimensions are large, a popular implementation of Muon uses the Newton-Schulz
 steps~\citep{jordanmuon,bernstein2024old} to approximate the orthogonalization process.

\begin{algorithm}
	\caption{\textbf{LiMuon} Optimizer}
	\label{alg:1}
	\begin{algorithmic}[1]
		\STATE \textbf{Input}: $\eta_t>0$, $\beta_t\in [0,1)$, $0<\hat{r} <r=\min(m,n)$ and $s\geq 2$;
		\STATE \textbf{Initialize:} $W_0\in \R^{m\times n}$, $\xi_0\sim \mathcal{D}$ and $M_0=\nabla f(W_0;\xi_0)$;
		\FOR{$t = 0, 1, \ldots, T-1$}
		\STATE $(U_{t}, \Sigma_{t}, V_{t}) = \text{SVD}(M_t)$ ;
		\STATE $W_{t+1} = W_t - \eta_t U_{t}V_{t}^\top$;
		\STATE  Draw a sample $\xi_{t+1} \sim \mathcal{D}$;
		\STATE $M_{t+1} = \nabla f(W_{t+1};\xi_{t+1}) + (1-\beta_{t+1})(M_t -\nabla f(W_t;\xi_{t+1}))$ (\textbf{Option} \#1);
		\STATE $(\hat{U}_{t}, \hat{S}_{t}, \hat{V}_{t}) = \text{RSVD}(M_t,\hat{r},s)$ (from Algorithm~\ref{alg:2}), and
		$\hat{M}_t=\hat{U}_{t} \hat{S}_{t} \hat{V}_{t}^T$;
		\STATE $M_{t+1} = \nabla f(W_{t+1};\xi_{t+1}) + (1-\beta_{t+1})(\textcolor{blue}{\hat{M}_t} -\nabla f(W_t;\xi_{t+1}))$ (\textbf{Option} \#2).
		\ENDFOR
		\STATE \textbf{Output}: $W_T$.
	\end{algorithmic}
\end{algorithm}

\begin{algorithm}
	\caption{\textbf{RSVD}($A,\hat{r},s$) }
	\label{alg:2}
	\begin{algorithmic} [1]
		\STATE \textbf{Input}: Matrix \(A \in \mathbb{R}^{m \times n}\), target rank \( \hat{r}>0 \),
		oversampling parameter \(s>0 \);
		\STATE \textbf{Output}: $ U, \Sigma, V$;
		\STATE \( l = \hat{r} + s \), and generate a random Gaussian matrix \( \Omega \in \mathbb{R}^{n \times l} \);
		\STATE \( Y = A \Omega \in \mathbb{R}^{m \times l} \), and compute the QR decomposition \( Y = QR \);
		\STATE \( B = Q^\top A \in \mathbb{R}^{l \times n} \), and compute SVD of the small matrix: \( (\tilde{U}, \Sigma, V) = \text{SVD}(B) \), and \( U = Q \tilde{U} \).
	\end{algorithmic}
\end{algorithm}

\begin{figure}
  \centering
  \includegraphics[width=0.48\textwidth]{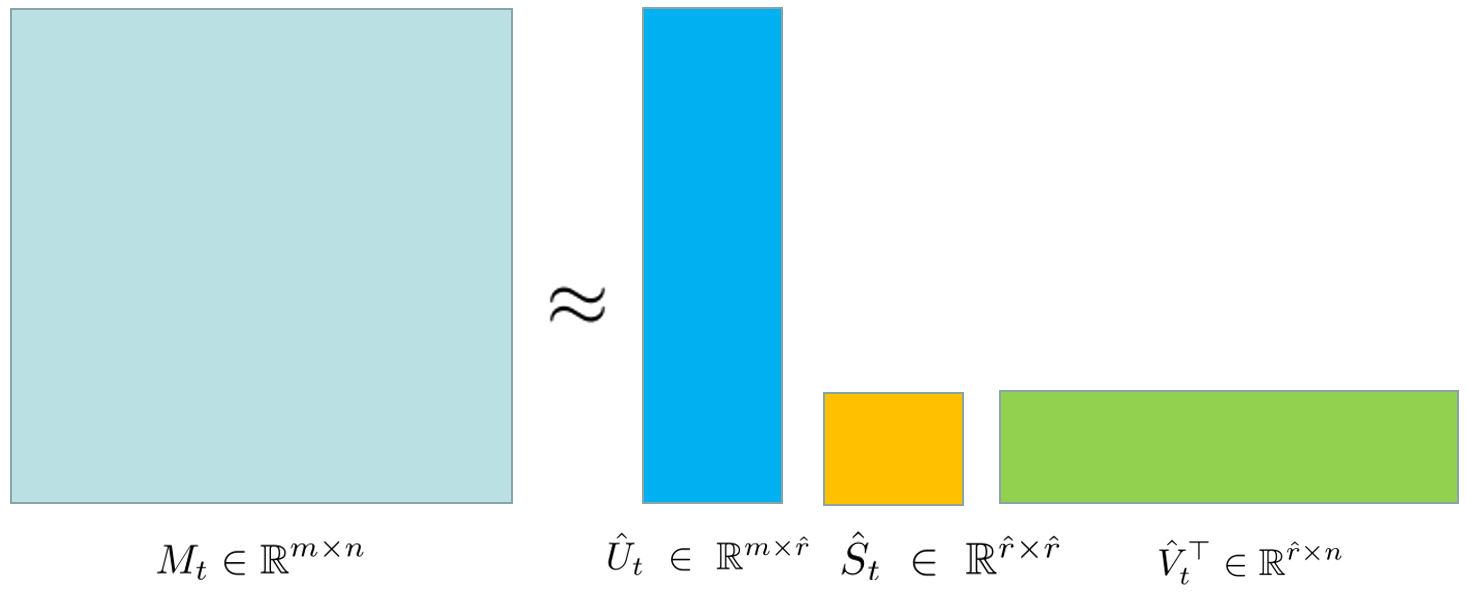}\\
  \caption{Low Rank Compression of Momentum $M_t$. }
  \label{fig:0}
\end{figure}

\section{ Our LiMuon Optimizer}
In the section, we propose a light and fast Muon (LiMuon) optimizer, which builds on the momentum-based variance reduced technique of STORM~\citep{cutkosky2019momentum} and randomized SVD~\citep{halko2011finding}. Algorithm~\ref{alg:1} provides an algorithmic framework of our LiMuon optimizer.

At the line 4 of Algorithm~\ref{alg:1}, we use the SVD to obtain the orthogonalization of stochastic gradient estimate $M_t$. At the line 7 of Algorithm~\ref{alg:1}, we use
 the momentum-based variance reduced technique of STORM to obtain stochastic gradient estimate
 $$M_{t+1} = \nabla f(W_{t+1};\xi_{t+1}) + (1-\beta_{t+1})(M_t -\nabla f(W_t;\xi_{t+1})).$$
 Here we still store full-rank 
 gradient estimate $M_t\in \R^{m\times n}$,
 our LiMuon algorithm with \textbf{Option} \#1 could not reduce state memory.
 
 Compared with the Muon$^{++}$~\citep{sfyraki2025lions} algorithm, our LiMuon (Option \#1) algorithm does not rely on gradient clipping, and uses less tuning parameter (without clipping parameter) in algorithmic framework. Due to its simplicity, our LiMuon algorithm has better performances than the Muon$^{++}$ in the experiments (Please see the following experimental results). 

At the line 8 of Algorithm~\ref{alg:1}, we use randomized SVD to obtain an approximated estimate $\hat{M}_t = \hat{U}_{t} \hat{S}_{t} \hat{V}_{t}^\top$, where $\hat{U}_t\in \R^{m\times \hat{r}}$, $\hat{S}_t\in \R^{\hat{r}\times \hat{r}}$ and $\hat{V}_t\in \R^{n\times \hat{r}}$. The randomized SVD (RSVD) algorithm is shown in Algorithm~\ref{alg:2}.  \textbf{Note that}:
we store the matrices $\hat{U}_t\in \R^{m\times \hat{r}}$, $\hat{S}_t\in \R^{\hat{r}\times \hat{r}}$ and $\hat{V}_t\in \R^{n\times \hat{r}}$ instead of the matrix
$M_t\in \R^{m\times n}$ (Please see Figure~\ref{fig:0}). Since
$\hat{r}\ll \min(m,n)$, we have $m\hat{r}+n\hat{r} +\hat{r}^2 \ll mn$. Thus, our LiMuon algorithm with \textbf{Option \#2} requires lower state memory than that of LiMuon algorithm with \textbf{Option} \#1.
Although this RSVD introduces additional computational cost, it significantly reduces state memory.

At the line 9 of Algorithm~\ref{alg:1}, we use a low-rank STORM-like gradient estimate:
\begin{align}
M_{t+1} \!=\! \nabla f(W_{t+1};\xi_{t+1}) \!+ (1-\beta_{t+1})(\textcolor{blue}{\hat{M}_t} \! -\! \nabla f(W_t;\xi_{t+1})), \nonumber
\end{align}
where $\hat{M}_t = \hat{U}_{t} \hat{S}_{t} \hat{V}_{t}^\top$ is a low-rank approximated estimate of
momentum $M_t$.

Algorithm~\ref{alg:3} provides an algorithmic framework of our LiMuon optimizer with Newton-Schulz steps. Since the exact SVD is expensive, our Algorithm~\ref{alg:3}  uses a small and fixed number of  Newton-Schulz steps to approximate the orthogonalization process as in~\citep{jordanmuon}. Here our Algorithm~\ref{alg:3} uses the matrix $O_t$ instead of the exact SVD-based polar factor $P_t=U_tV_t^{\top}$ used in Algorithm~\ref{alg:1}. In practice, 
Muon~\citep{jordanmuon} uses the polynomial \(p_{2} (z)= 3.4445-4.7750z+2.0315z^2\).

\begin{algorithm}
	\caption{\textbf{LiMuon} Optimizer with Newton-Schulz}
	\label{alg:3}
	\begin{algorithmic}[1]
		\STATE \textbf{Input}: $\eta_t>0$, $\beta_t\in [0,1)$, $0<\hat{r} <r=\min(m,n)$, $s\geq 2$ and $q\geq 1$;
		\STATE \textbf{Initialize:} $W_0\in \R^{m\times n}$, $\xi_0\sim \mathcal{D}$ and $M_0=\nabla f(W_0;\xi_0)$;
		\FOR{$t = 0, 1, \ldots, T-1$}
		\STATE $O_t = \text{Newton-Schulz}(M_t,q)$ (from Algorithm~\ref{alg:4});
		\STATE $W_{t+1} = W_t - \eta_t O_t$;
		\STATE  Draw a sample $\xi_{t+1} \sim \mathcal{D}$;
		\STATE $M_{t+1} = \nabla f(W_{t+1};\xi_{t+1}) + (1-\beta_{t+1})(M_t -\nabla f(W_t;\xi_{t+1}))$ (\textbf{Option} \#1);
		\STATE $(\hat{U}_{t}, \hat{S}_{t}, \hat{V}_{t}) = \text{RSVD}(M_t,\hat{r},s)$ (from Algorithm~\ref{alg:2}), and
		$\hat{M}_t=\hat{U}_{t} \hat{S}_{t} \hat{V}_{t}^T$;
		\STATE $M_{t+1} = \nabla f(W_{t+1};\xi_{t+1}) + (1-\beta_{t+1})(\textcolor{blue}{\hat{M}_t} -\nabla f(W_t;\xi_{t+1}))$ (\textbf{Option} \#2).
		\ENDFOR
		\STATE \textbf{Output}: $W_T$.
	\end{algorithmic}
\end{algorithm}

\begin{algorithm}
	\caption{\textbf{ Newton-Schulz}($M,q$) }
	\label{alg:4}
	\begin{algorithmic} [1]
		\STATE \textbf{Input}: Matrix \(M \in \mathbb{R}^{m \times n}\), step size \( q\geq 1 \), and 
		 polynomial \(p_{\kappa}(\cdot) \), where $\kappa$ is degree;
		\STATE \textbf{Initialize:} $X_0=M/\alpha$ with $\alpha =\max(1,\|M\|_\F)$;
	    \FOR{$j = 1, 2,\ldots, q$}
		\STATE $X_j = p_{\kappa}(X_{j-1}X_{j-1}^\top)X_{j-1}$;
		\ENDFOR
		\STATE $O=X_q$;
		\STATE \textbf{Output}: $O$.
	\end{algorithmic}
\end{algorithm}

\section{ Convergence Analysis}
In this section, we provide convergence analysis for our LiMuon algorithm with \textbf{exact SVD} (\emph{using in orthogonalization process}) and \textbf{Newton-Schulz steps}, respectively. All detailed proofs are provided in the following Appendix.
We first give the definition of $\epsilon$-Frobenius norm and $\epsilon$-nuclear norm stationary point, respectively.

\begin{definition}\label{def1}
	If $W^*$ be an $\epsilon$-Frobenius norm stationary point of $F(W)$ such as
	$$\|\nabla F(W^*)\|_\F\leq\epsilon, \ \epsilon > 0.$$
\end{definition} 

\begin{definition}\label{def2}
	If $W^*$ be an $\epsilon$-nuclear norm stationary point of $F(W)$ such as
	$$\|\nabla F(W^*)\|_*\leq\epsilon, \ \epsilon > 0.$$
\end{definition} 

\subsection{ Convergence Analysis of LiMuon \textcolor{blue}{with Exact SVD}}
In this subsection, we provide convergence analysis for our LiMuon algorithm with exact SVD under \textbf{Option \#1
and \#2}, respectively.
We first give some mild assumptions.

\begin{assumption}[\textbf{Frobenius-Norm Lipschitz Smoothness}]\label{ass:s1}
Functions $F(W)$ and $f(W;\xi)$ for all $\xi\sim \mathcal{D}$ are $L$-Frobenius norm Lipschitz smooth, if for any $W, W'\in \R^{m\times n}$, we have
\begin{align}
    \|\nabla F(W)-\nabla F(W')\|^2_\F\leq L^2\|W-W'\|^2_\F, \nonumber \\
    \|\nabla f(W;\xi)-\nabla f(W';\xi)\|^2_\F\leq L^2\|W-W'\|^2_\F. \nonumber
\end{align}
\end{assumption}

\begin{assumption}[\textbf{Frobenius-Norm Generalized Smoothness}]\label{ass:s2}
Functions $F(W)$ and $f(W;\xi)$ for all $\xi\sim \mathcal{D}$ are $(L_0,L_1)$-Frobenius norm smooth, if for any $W, W'\in \R^{m\times n}$, we have
\begin{align}
&\|\nabla F(W)-\nabla F(W')\|_\F^2   \nonumber \\
& \leq (L_0^2+L_1^2\|\nabla F(W)\|_\F^2)\|W-W'\|_\F^2, \nonumber \\
&\E\|\nabla f(W;\xi)-\nabla f(W';\xi)\|_\F^2 \nonumber \\
& \leq  (L_0^2+L_1^2(\E\|\nabla F(W)\|_\F)^2)\|W-W'\|_\F^2. \nonumber
\end{align}
\end{assumption}

Assumption~\ref{ass:s1} is the standard Frobenius norm smoothness, which is a natural extension of the conventional $l_2$-norm smoothness for functions with vector parameter to functions with matrix parameter~\citep{beck2017first}. Assumption~\ref{ass:s2} is a Frobenius norm $(L_0,L_1)$-smoothness, which can be seen as a natural extension of the conventional $l_2$-norm $(L_0,L_1)$-smoothness for functions with vector parameter to functions with matrix parameter~\citep{zhang2019gradient}.

For a stochastic setting, we also use the following standard bounded variance assumption.
\begin{assumption}[Bounded Variance]\label{ass:v}
$\nabla f(W;\xi)$ is an unbiased stochastic estimator of the true gradient $\nabla F(W)$ and has a bounded variance, i.e., $\E[\nabla f(W;\xi)]=\nabla F(W)$ and
\begin{align}
 \E\|\nabla f(W;\xi)-\nabla F(W)\|_\F^2]\leq \sigma^2. \nonumber
\end{align}
\end{assumption}

\begin{assumption} \label{ass:l}
 The objective function $F(W)$ is lower bounded, i.e.,  $F^* = \inf_{W\in
\mathbb{R}^{m\times n}} F(W)>-\infty$.
\end{assumption}

Assumption~\ref{ass:v} is the standard bounded variance condition of unbiased stochastic gradient, which is a natural extension of the conventional $l_2$-norm smoothness for functions with vector parameters to functions with matrix parameters~\citep{shen2025convergence,li2025note}. Assumption~\ref{ass:l} guarantees feasibility of Problem~(\ref{eq:p1}). 

\subsubsection{Convergence Analysis of LiMuon Algorithm with  \textcolor{blue}{Exact SVD and Option \#1 under Lipschitz Smoothness}}

\begin{theorem} \label{th:1}
Under the Assumptions~\ref{ass:s1},~\ref{ass:v},~\ref{ass:l}, the sequence $\{W_t\}_{t=0}^T$ be generated
from Algorithm \ref{alg:1} with \textbf{Option} \#1. Let $\beta_t=\beta \in (0,1)$ and $\eta_t=\eta$ for all $t\geq 0$, we have
\begin{align}
\frac{1}{T+1}\sum_{t=0}^{T} &\E \|\nabla F(W_t)\|_* \leq \frac{F(W_0)- F^*}{T\eta}  + \frac{r L\eta}{2}   \nonumber \\
 & \quad + 2\sqrt{r}\big(\frac{\sigma}{T\beta} + \sqrt{\frac{\beta}{2-\beta}}\sigma + \sqrt{\frac{r(1-\beta)^2}{(2-\beta)\beta}}L\eta \big), \nonumber
\end{align}
 where $r= \min(m,n)$.
\end{theorem}

\begin{remark}
From the above Theorem~\ref{th:1},
let $\eta=O(\frac{1}{T^{2/3}})$ and $\beta=O(\frac{1}{T^{2/3}})$, we have
\begin{align}
\min_{0\leq t\leq T} \E \|\nabla F(W_t)\|_* &\leq \frac{1}{T+1}\sum_{t=0}^{T} \E \|\nabla F(W_t)\|_*
 \nonumber \\
 & \leq O\big(\frac{1}{T^{1/3}}\big) \leq \epsilon. \nonumber
\end{align}
Thus, our LiMuon has a fast convergence rate of $O(\frac{1}{T^{1/3}})$. 
Since our LiMuon algorithm only requires one sample at each iteration, i.e., batch size $b=1$,
so it has a lower sample (or SFO) complexity of $bT = O(\epsilon^{-3})$ for finding an
$\epsilon$-nuclear-norm stationary point of the problem~(\ref{eq:p1}). Since $\|\nabla F(W_t)\|_\F \leq \|\nabla F(W_t)\|_*$ for all $W_t$, our LiMuon algorithm also has a lower SFO complexity of $O(\epsilon^{-3})$ for finding an $\epsilon$-Frobenius-norm stationary point of the problem~(\ref{eq:p1}).
\end{remark}

\subsubsection{Convergence Analysis of LiMuon Algorithm with   \textcolor{blue}{Exact SVD and Option \#1 under generalized smoothness}}

\begin{theorem} \label{th:2}
Under the Assumptions~\ref{ass:s2},~\ref{ass:v},~\ref{ass:l}, the sequence $\{W_t\}_{t=0}^T$ be generated
from Algorithm \ref{alg:1} with \textbf{Option} \#1. Let $\beta_t=\beta \in (0,1)$ and $\eta_t=\eta \leq \min\big(\frac{1}{2L_1r},\frac{\beta}{8L_1r(1-\beta)}\big)$ for all $t\geq 0$, we have
\begin{align}
& \frac{1}{T+1}\sum_{t=0}^{T}  \E \|\nabla F(W_t)\|_* \leq  \frac{2(F(W_0)- F^*)}{T\eta}  + r L_0\eta
\nonumber \\
&\quad + \frac{4\sqrt{r}\sigma}{T\beta} + 4\sqrt{r}\sqrt{\frac{\beta}{2-\beta}}\sigma
 + 4L_0\eta r\sqrt{\frac{(1-\beta)^2}{(2-\beta)\beta}}, \nonumber
\end{align}
 where $r= \min(m,n)$.
\end{theorem}

\begin{remark}
From the above Theorem~\ref{th:2},
let $\eta=O(\frac{1}{T^{2/3}})$ and $\beta=O(\frac{1}{T^{2/3}})$, we have
\begin{align}
 &\frac{1}{T+1}\sum_{t=0}^{T} \E \|\nabla F(W_t)\|_*
\leq O\big(\frac{1}{T^{1/3}}\big) \leq \epsilon. \nonumber
\end{align}
Thus, our LiMuon has a fast convergence rate of $O(\frac{1}{T^{1/3}})$. 
Since our LiMuon algorithm only requires a sample at each iteration, i.e., batch size $b=1$,
so it has a lower sample (or gradient) complexity of $bT = O(\epsilon^{-3})$ for finding an
$\epsilon$-nuclear-norm stationary solution of the problem~(\ref{eq:p1}) under
generalized smoothness condition.
\end{remark}

\subsubsection{Convergence Analysis of LiMuon Algorithm with   \textcolor{blue}{Exact SVD and Option \#2 under Lipschitz Smoothness}}

\begin{theorem} \label{th:3}
Under the Assumptions~\ref{ass:s1},~\ref{ass:v},~\ref{ass:l}, the sequence $\{W_t\}_{t=0}^T$ be generated
from Algorithm \ref{alg:1} with \textbf{Option} \#2. Let $\beta_t=\beta \in (0,1)$, $\eta_t=\eta$ for all $t\geq 0$, $s\geq 2$, $\hat{r}\geq 2$, $s+\hat{r}\leq r=\min(m,n)$, and $\gamma \leq \frac{\beta}{3\sqrt{r}(1-\beta)}$, we have
\begin{align}
	& \frac{1}{T+1}\sum_{t=0}^{T}\E \|\nabla F(W_t)\|_* \leq \frac{2(F(W_0)- F^*)}{T\eta}  + r L\eta \nonumber \\
	& \quad + 3\sqrt{r}\big(\frac{\sigma}{T\beta} + \sqrt{\frac{\beta}{2-\beta}}\sigma + \sqrt{\frac{r(1-\beta)^2}{(2-\beta)\beta}}L\eta \big), \nonumber
\end{align}
where $\gamma=\tau\left(1 + \frac{\hat{r}}{s - 1} \right)^{\frac{1}{2}}$ with $\tau\in(0,1)$.
\end{theorem}

\begin{remark}
From the above Theorem~\ref{th:3},
let $\eta=O(\frac{1}{T^{2/3}})$ and $\beta=O(\frac{1}{T^{2/3}})$, we have
\begin{align}
 \frac{1}{T+1}\sum_{t=0}^{T}\E\|\nabla F(W_t)\|_*
    \leq O\big(\frac{1}{T^{1/3}}\big) \leq \epsilon. \nonumber
\end{align}
Since our LiMuon algorithm only requires a sample at each iteration, i.e., batch size $b=1$,
so it has a lower sample (or gradient) complexity of $bT = O(\epsilon^{-3})$ for finding an
$\epsilon$-nuclear-norm stationary solution of the problem~(\ref{eq:p1}).
From this result, our LiMuon algorithm with
lower memory cost still obtains a lower sample complexity of $O(\epsilon^{-3})$.
\end{remark}

\subsubsection{Convergence Analysis of LiMuon Algorithm with   \textcolor{blue}{Exact SVD and Option \#2 under generalized smoothness}}

\begin{theorem} \label{th:4}
Under the Assumptions~\ref{ass:s2},~\ref{ass:v},~\ref{ass:l}, the sequence $\{W_t\}_{t=0}^T$ be generated
from Algorithm \ref{alg:1} with \textbf{Option} \#2. Let $\beta_t=\beta \in (0,1)$ and $\eta_t=\eta \leq \min\big(\frac{1}{4L_1r}, \frac{\beta}{12L_1r(1-\beta)} \big)$ for all $t\geq 0$, and $s\geq 2$, $\hat{r}\geq 2$, $s+\hat{r}\leq r$, and $\gamma \leq \frac{\beta}{3\sqrt{r}(1-\beta)}$, we have
\begin{align}
	& \frac{1}{T+1}\sum_{t=0}^{T}\E \|\nabla F(W_t)\|_* \leq  \frac{4(F(W_0)- F^*)}{T\eta}  + 2r L_0\eta  \nonumber \\
	& \quad + \frac{6\sqrt{r}\sigma}{T\beta} + 6\sqrt{r}\sqrt{\frac{\beta}{2-\beta}}\sigma
	+ 6L_0\eta r\sqrt{\frac{(1-\beta)^2}{(2-\beta)\beta}}, \nonumber
\end{align}
where $\gamma=\tau\left(1 + \frac{\hat{r}}{s - 1} \right)^{\frac{1}{2}}$ and $\tau\in(0,1)$.
\end{theorem}

\begin{remark}
From the above Theorem~\ref{th:4},
let $\eta=O(\frac{1}{T^{2/3}})$ and $\beta=O(\frac{1}{T^{2/3}})$, we have
\begin{align}
   \frac{1}{T+1}\sum_{t=0}^{T}\E \|\nabla F(W_t)\|_*
     \leq  O\big(\frac{1}{T^{1/3}}\big) \leq \epsilon. \nonumber
\end{align}
Our LiMuon algorithm with
lower memory cost still obtains a lower sample complexity of $O(\epsilon^{-3})$ under
generalized smoothness condition.
\end{remark}

\subsection{ Convergence Analysis of LiMuon \textcolor{blue}{with Newton-Schulz}}
In this subsection, we provide convergence analysis for our LiMuon algorithm with Newton-Schulz under \textbf{Option \#1
and \#2}, respectively. We first give some mild assumptions as in~\citep{kim2026convergence}.
.

\begin{assumption}[\textbf{Nuclear-Norm Lipschitz Smooth}]\label{ass:ss1}
	Functions $F(W)$ and $f(W;\xi)$ for all $\xi\sim \mathcal{D}$ are $L$-nuclear norm Lipschitz smooth, if for any $W, W'\in \R^{m\times n}$, we have
	\begin{align}
		\|\nabla F(W)-\nabla F(W')\|^2_*\leq L^2\|W-W'\|^2_{op}, \nonumber \\
		\|\nabla f(W;\xi)-\nabla f(W';\xi)\|^2_* \leq L^2\|W-W'\|^2_{op}. \nonumber
	\end{align}
\end{assumption}

\begin{assumption}[\textbf{Nuclear-Norm Generalized Smoothness}]\label{ass:ss2}
	Functions $F(W)$ and $f(W;\xi)$ for all $\xi\sim \mathcal{D}$ are $(L_0,L_1)$-nuclear norm  smooth, if for any $W, W'\in \R^{m\times n}$, we have
	\begin{align}
		& \|\nabla F(W)-\nabla F(W')\|_*^2  \nonumber \\
		& \leq 
		(L_0^2+L_1^2\|\nabla F(W)\|_*^2)\|W-W'\|_{op}^2, \nonumber \\
		& \E\|\nabla f(W;\xi)-\nabla f(W';\xi)\|_*^2 
		\nonumber \\
		&\leq 
		(L_0^2+L_1^2(\E\|\nabla F(W)\|_*)^2)\|W-W'\|_{op}^2. \nonumber
	\end{align}
\end{assumption}

Meanwhile, following~\cite{kim2026convergence}, we can define a polar approximation error, which is the error between the exact polar factor $P_t=U_tV_t^\top$ with $[U_t,\Sigma_t,V_t]=\text{SVD}(M_t)$) and the actual step $O_t$ generated by $q$-step Newton-Schulz steps, i.e., $O_t =\text{Newton-Schulz}(M_t,q)$. This polar approximation error is defined as 
\begin{align}
	\varepsilon_{q,t}=\|O_t-P_t\|_{op} \quad \varepsilon_{q} = \sup_{t} \varepsilon_{q,t}.
\end{align}
Following Theorem 2 of ~\cite{kim2026convergence}, we can set $\varepsilon_{q}\in (0,1)$. 
Since $P_t=U_tV_t^\top$, $\varepsilon_{q,t}=\|O_t-P_t\|_{op}$ with $\varepsilon_{q}=\sup_{t} \varepsilon_{q,t}$, we have 
\begin{align}
	\|P_t\|_{op}=1, \quad \|O_t\|_{op} \leq 1+\varepsilon_{q,t}\leq 1+\varepsilon_{q}.
\end{align}

\subsubsection{Convergence Analysis of LiMuon Algorithm with  \textcolor{blue}{Newton-Schulz and Option \#1 under Lipschitz Smoothness}}

\begin{theorem} \label{th:5}
	Under the above Assumptions~\ref{ass:ss1},~\ref{ass:v},~\ref{ass:l}, the sequence $\{W_t\}_{t=0}^T$ be generated
	from Algorithm \ref{alg:3} with \textbf{Option} \#1. Let $\beta_t=\beta \in (0,1)$ and $\eta_t=\eta$ for all $t\geq 0$, we have
	\begin{align}
		& \frac{1}{T+1}\sum_{t=0}^{T}\E \|\nabla F(W_t)\|_* \leq \frac{F(W_0)- F^*}{T\eta\chi_q^{-1}} + \frac{ L\eta(1+\varepsilon_{q})^2}{2\chi_q^{-1}}  
		\nonumber \\
		& +2\chi_q\Big( \frac{\sqrt{r}\sigma}{T\beta} + \sqrt{r}\sqrt{\frac{\beta}{2-\beta}}\sigma + \sqrt{\frac{r(1-\beta)^2}{(2-\beta)\beta}}L\eta (1+\varepsilon_{q})\Big), \nonumber
	\end{align}
	where $\chi_q= \frac{1}{1-\varepsilon_q}>1$, and $\varepsilon_{q}\in (0,1)$ is a polar approximation error.
\end{theorem}

\begin{remark}
	From the above Theorem~\ref{th:5},
	let $\eta=O(\frac{1}{T^{2/3}})$ and $\beta=O(\frac{1}{T^{2/3}})$, we have
	\begin{align}
		\frac{1}{T+1}\sum_{t=0}^{T}\E \|\nabla F(W_t)\|_*
		\leq  O\big(\frac{\chi_q}{T^{1/3}}\big) \leq \epsilon. \nonumber
	\end{align}
	Our LiMuon algorithm obtains a lower sample complexity of $O(\chi_q^3\epsilon^{-3})$ under
	Lipschitz smooth condition than $O(\chi_q^4\epsilon^{-4})$ of the Muon optimizer with Newton-Schulz~\citep{kim2026convergence}.
\end{remark}

\subsubsection{Convergence Analysis of LiMuon Algorithm with  \textcolor{blue}{Newton-Schulz and Option \#1 under generalized smoothness}}

\begin{theorem} \label{th:6}
	Under the Assumptions~\ref{ass:ss2},~\ref{ass:v},~\ref{ass:l}, the sequence $\{W_t\}_{t=0}^T$ be generated
	from Algorithm \ref{alg:3} with \textbf{Option} \#1. Let $\beta_t=\beta \in (0,1)$ and $\eta_t=\eta \leq \min\big(\frac{1-\varepsilon_{q}}{2L_1(1+\varepsilon_{q})^2},\frac{\beta(1-\varepsilon_{q})}{8L_1\sqrt{r}(1+\varepsilon_{q})(1-\beta)}\big)$ for all $t\geq 0$, we have
	\begin{align}
	&  \frac{1}{T+1}\sum_{t=0}^{T}\E \|\nabla F(W_t)\|_* \! \leq \! \frac{2((W_0)- F^*)}{(T+1)\eta\chi_q^{-1}} \!  + \frac{L_0\eta(1+\varepsilon_{q})^2}{\chi_q^{-1}} \nonumber \\
	& + 4\chi_{q}\Big(\frac{\sqrt{r}\sigma}{T\beta} + \sqrt{\frac{r\beta}{2-\beta}}\sigma + \sqrt{\frac{r(1-\beta)^2}{(2-\beta)\beta}}L_0\eta (1+\varepsilon_{q})\Big), \nonumber
	\end{align}
	where $\chi_q= \frac{1}{1-\varepsilon_q}>1$.
\end{theorem}

\begin{remark}
	From the above Theorem~\ref{th:6},
	let $\eta=O(\frac{1}{T^{2/3}})$ and $\beta=O(\frac{1}{T^{2/3}})$, we have
	\begin{align}
		\frac{1}{T+1}\sum_{t=0}^{T}\E \|\nabla F(W_t)\|_*
		\leq  O\big(\frac{\chi_q}{T^{1/3}}\big) \leq \epsilon. \nonumber
	\end{align}
	Our LiMuon algorithm obtains a lower sample complexity of $O(\chi_q^3\epsilon^{-3})$ under
	generalized smoothness condition than $O(\chi_q^4\epsilon^{-4})$ of the Muon optimizer with Newton-Schulz~\citep{kim2026convergence}.
\end{remark}

\subsubsection{Convergence Analysis of LiMuon Algorithm with  \textcolor{blue}{Newton-Schulz and Option \#2 under Lipschitz Smoothness}}

\begin{theorem} \label{th:7}
	Under the Assumptions~\ref{ass:ss1},~\ref{ass:v},~\ref{ass:l}, the sequence $\{W_t\}_{t=0}^T$ be generated
	from Algorithm \ref{alg:3} with \textbf{Option} \#2. Let $\beta_t=\beta \in (0,1)$, $\eta_t=\eta$ for all $t\geq 0$, 
	$s\geq 2$, $\hat{r}\geq 2$, $\hat{r}+s\leq r=\min(m,n)$, and $\gamma \leq  \frac{\beta(1-\varepsilon_{q})}{(3+\varepsilon_{q})\sqrt{r}(1-\beta)}$, we have
	\begin{align}
	\frac{1}{T+1}& \sum_{t=0}^{T} \E \|\nabla F(W_t)\|_* \leq \frac{2(F(W_0)- F^*)}{T\eta\chi_q^{-1}}  \nonumber \\
	& + \frac{\eta L(1+ \varepsilon_{q})^2}{\chi_q^{-1}}+ \frac{3+\varepsilon_{q}}{1-\varepsilon_{q}}\Big( \frac{\sqrt{r}\sigma}{T\beta} + \sqrt{r}\sqrt{\frac{\beta}{2-\beta}}\sigma  \nonumber \\
	&\quad + \sqrt{r}\sqrt{\frac{(1-\beta)^2}{(2-\beta)\beta}}L\eta (1+\varepsilon_{q}) \Big), \nonumber
	\end{align}
	where $\gamma=\tau\left(1 + \frac{\hat{r}}{s - 1} \right)^{\frac{1}{2}}$ with $\tau\in(0,1)$, and $\chi_q= \frac{1}{1-\varepsilon_q}>1$.
\end{theorem}

\begin{remark}
	From the above Theorem~\ref{th:7},
	let $\eta=O(\frac{1}{T^{2/3}})$ and $\beta=O(\frac{1}{T^{2/3}})$, we have
	\begin{align}
		\frac{1}{T+1}\sum_{t=0}^{T}\E \|\nabla F(W_t)\|_*
		\leq  O\big(\frac{\chi_q}{T^{1/3}}\big) \leq \epsilon. \nonumber
	\end{align}
	Our LiMuon algorithm with
	lower memory cost still obtains a lower sample complexity of $O(\chi_q^3\epsilon^{-3})$ under
	Lipschitz smooth condition than $O(\chi_q^4\epsilon^{-4})$ of the Muon optimizer with Newton-Schulz~\citep{kim2026convergence}.
\end{remark}

\subsubsection{Convergence Analysis of LiMuon Algorithm with  \textcolor{blue}{Newton-Schulz and Option \#2 under generalized smoothness}}

\begin{theorem} \label{th:8}
	Under the Assumptions~\ref{ass:ss2},~\ref{ass:v},~\ref{ass:l}, the sequence $\{W_t\}_{t=0}^T$ be generated
	from Algorithm \ref{alg:3} with \textbf{Option} \#2. Let $\beta_t=\beta \in (0,1)$, and $\eta_t=\eta \leq \min\big(\frac{1-\varepsilon_{q}}{4L_1(1+\varepsilon_{q})^2},\frac{(1-\varepsilon_{q})\beta}{4L_1\sqrt{r}(1-\beta)(1+\varepsilon_{q})(3+\varepsilon_{q})}\big)$ for all $t\geq 0$, $s\geq 2$, $\hat{r}\geq 2$, $\hat{r}+s\leq r$ and $\gamma \leq \frac{(1-\varepsilon_{q})\beta}{(3+\varepsilon_{q})\sqrt{r}(1-\beta)}$, we have
	\begin{align}
		& \frac{1}{T+1}\sum_{t=0}^{T}\E \|\nabla F(W_t)\|_*
		 \leq  \frac{4(F(W_0)- F^*)}{(T+1)\eta\chi_{q}^{-1}}     \nonumber \\
		& + \frac{2L_0\eta(1+\varepsilon_{q})^2}{\chi_{q}^{-1}} + \frac{2(3+\varepsilon_{q})}{\chi_{q}^{-1}}\Big(\frac{\sqrt{r}\sigma}{(T+1)\beta} + \sqrt{r}\sqrt{\frac{\beta}{2-\beta}}\sigma \nonumber \\
		& +\sqrt{r}\sqrt{\frac{(1-\beta)^2}{(2-\beta)\beta}}L_0\eta (1+\varepsilon_{q}) \Big),  \nonumber
	\end{align}
	where $\gamma=\tau\left(1 + \frac{\hat{r}}{s - 1} \right)^{\frac{1}{2}}$ with $\tau\in(0,1)$, and $\chi_q= \frac{1}{1-\varepsilon_q}>1$.
\end{theorem}

\begin{remark}
	From the above Theorem~\ref{th:8},
	let $\eta=O(\frac{1}{T^{2/3}})$ and $\beta=O(\frac{1}{T^{2/3}})$, we have
	\begin{align}
		\frac{1}{T+1}\sum_{t=0}^{T}\E \|\nabla F(W_t)\|_*
		\leq  O\big(\frac{\chi_q}{T^{1/3}}\big) \leq \epsilon. \nonumber
	\end{align}
	Our LiMuon algorithm with
	lower memory cost still obtains a lower sample complexity of $O(\chi_q^3\epsilon^{-3})$ under
	generalized smoothness condition than $O(\chi_q^4\epsilon^{-4})$ of the Muon optimizer with Newton-Schulz~\citep{kim2026convergence}.
\end{remark}

\begin{table}[!ht]
	\centering
	\caption{Performance comparison at Mamba-130M}
	\label{tab:mamba}
	\resizebox{0.49\textwidth}{!}{
		\begin{tabular}{c|c|c|c}
			\toprule
			\textbf{Method} & {\textbf{Memory (GB)}} $\downarrow$ & \textbf{Training PPL} $\downarrow$ & \textbf{Validation PPL} $\downarrow$ \\
			\midrule
			Adam & 22.92 & 480.33 & 526.19 \\
			AdamW & 22.92 & 253.75 & 266.43 \\
			Lion & 22.56 & 75.25 & 83.82 \\
			SUMO & 20.10 & 111.56 & 118.52 \\
			Muon & 22.20 & 62.49 & 71.27 \\
			Muon++ & 22.35 & 51.97 & 56.79 \\
			\midrule
			\textbf{LiMuon(rank=4)} & \textbf{18.86} & 71.46 & 79.12 \\
			\textbf{LiMuon(rank=8)} & \textbf{20.25} & 59.07 & 62.23 \\
			\textbf{LiMuon(rank=16)} & 22.31 & \textbf{50.26} & \textbf{53.84} \\
			\textbf{LiMuon(full rank)} & 22.80 & \textbf{47.08} & \textbf{47.78} \\
			\bottomrule
		\end{tabular}
	}
\end{table}

\begin{figure}[ht]
	\centering
	\subfigure{\includegraphics[width=0.235\textwidth]{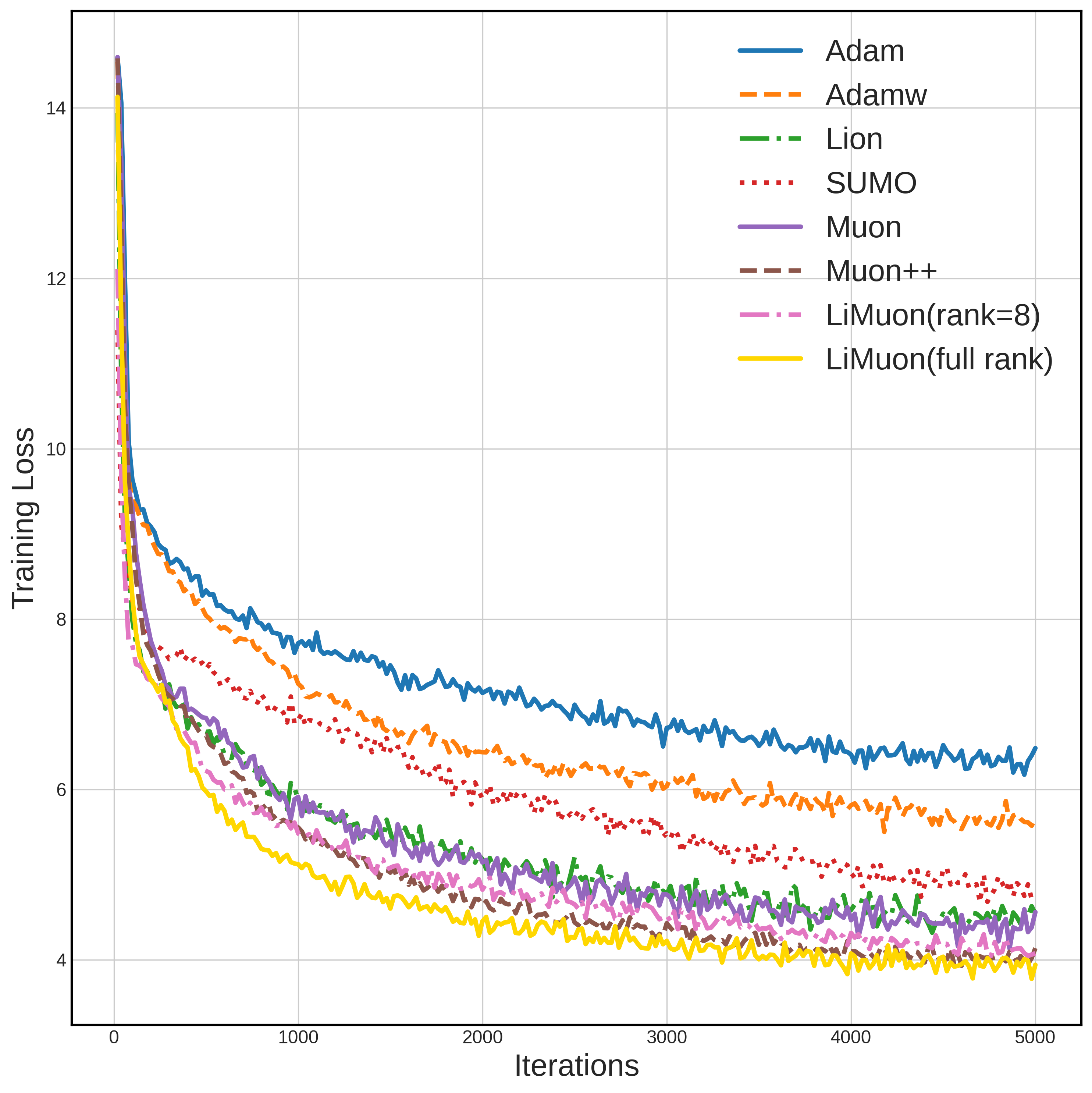}}
	\hfill
	\subfigure{\includegraphics[width=0.235\textwidth]{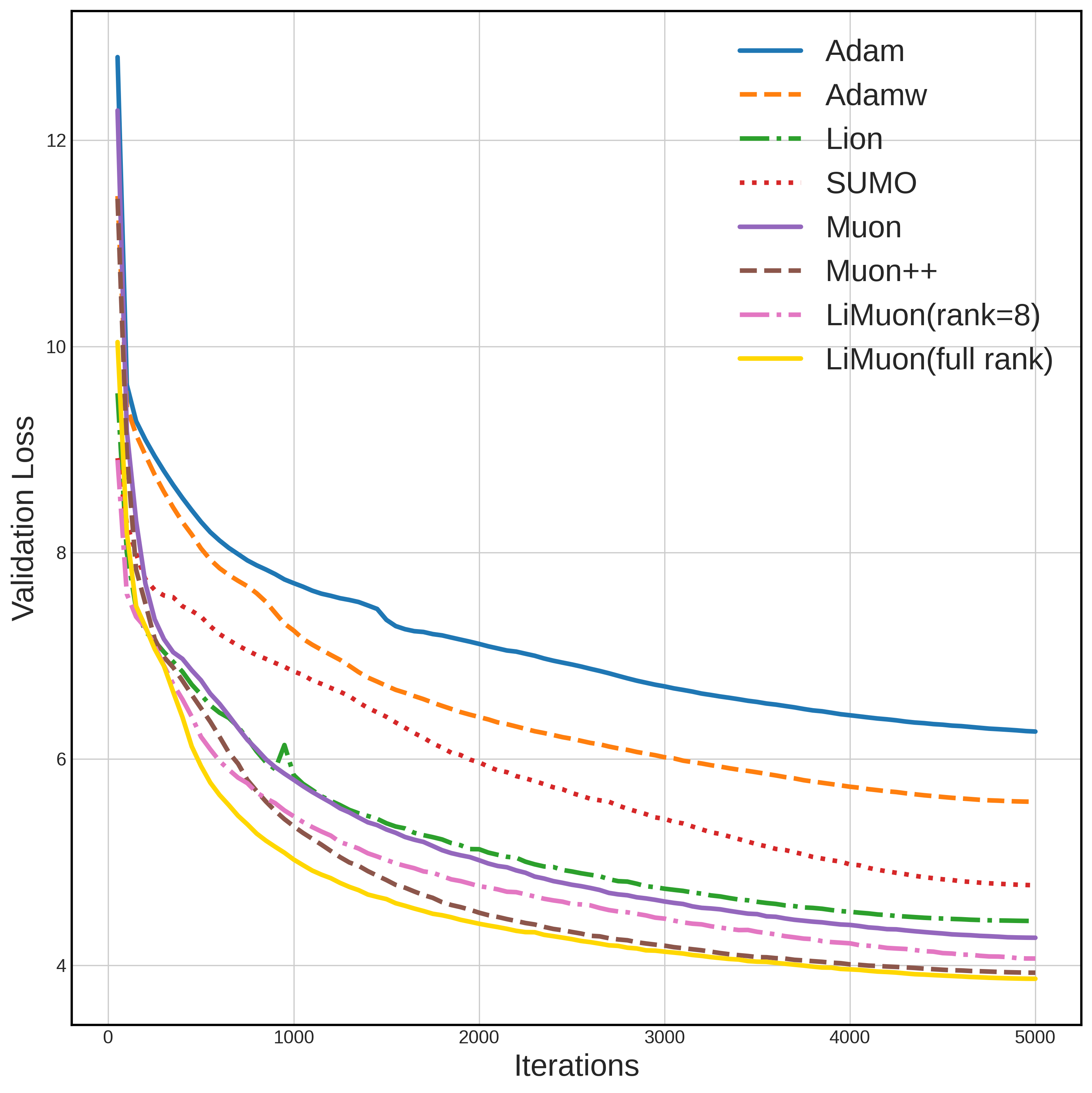}}
	\hfill
	\caption{Performance comparison at Mamba-130M }
	\label{fig:mamba}
\end{figure}

\section{Numerical Experiments}
In the section, we conduct some numerical experiments to demonstrate effectiveness of
our LiMuon optimizer on pre-training large models including Mamba-130M~\citep{gu2023mamba}, Qwen2.5-0.5B~\citep{qwen2.5} and ViT~\citep{dosovitskiy2020image}.
Specifically, we pre-train Mamba-130M on the wikitext-103~\citep{merity2016pointer} dataset, and pre-train Qwen2.5-0.5B on the MiniPile~\citep{kaddour2023minipile} dataset, and
pre-train ViT on the tiny-ImageNet~\citep{le2015tiny} dataset.

In the experiments, we compare our LiMuon optimizer with some typical optimizers including Adam~\citep{kingma2014adam}, AdamW~\citep{loshchilov2017decoupled}, Lion~\citep{chen2023symbolic}, Muon~\citep{jordanmuon}, Muon$^{++}$~\citep{sfyraki2025lions} and SUMO~\citep{refael2025sumo}.
Since the SCG~\citep{pethick2025training}, Gluon~\citep{riabinin2025gluon} and GGNC~\citep{pethick2025generalized} in Table~\ref{tab:1} mainly focus on the convergence analysis of the Muon optimizer, these algorithms basically follow the Muon optimizer.
Thus, we use the standard Muon optimizer instead of the SCG, Gluon and GGNC optimizer as a comparison.
All the experiments were conducted on the NVIDIA A100-SXM4-80GB.

\begin{table}[!ht]
    \centering
    \caption{Performance comparison at Qwen-2.5-0.5B}
    \label{tab:qwen}
     \resizebox{0.49\textwidth}{!}{
    \begin{tabular}{c|c|c|c}
        \toprule
        \textbf{Method} & {\textbf{Memory (GB)}} $\downarrow$ & \textbf{Training PPL} $\downarrow$ & \textbf{Validation PPL} $\downarrow$ \\
        \midrule
        Adam & 55.24 & 91.84 & 181.27 \\
        AdamW & 55.26 & 57.97 & 113.25 \\
        Lion & 54.08 & 48.42 & 113.92 \\
        SUMO & 52.73 & 41.26 & 88.73 \\
        Muon & 54.14 & 35.52 & 67.60 \\
        Muon++ & 54.30 & 35.16 & 82.26 \\
        \midrule
        \textbf{LiMuon(rank=4)} & \textbf{50.86} & 41.46 & 92.06 \\
        \textbf{LiMuon(rank=8)} & \textbf{52.55} & 38.86 & \textbf{54.87} \\
        \textbf{LiMuon(rank=16)} & 54.21 & \textbf{33.43} & \textbf{46.77} \\
        \textbf{LiMuon(full rank)} & 55.15 & \textbf{30.87} & \textbf{40.83} \\
        \bottomrule
    \end{tabular}
    }
\end{table}

\begin{figure}[ht]
\centering
  \subfigure{\includegraphics[width=0.235\textwidth]{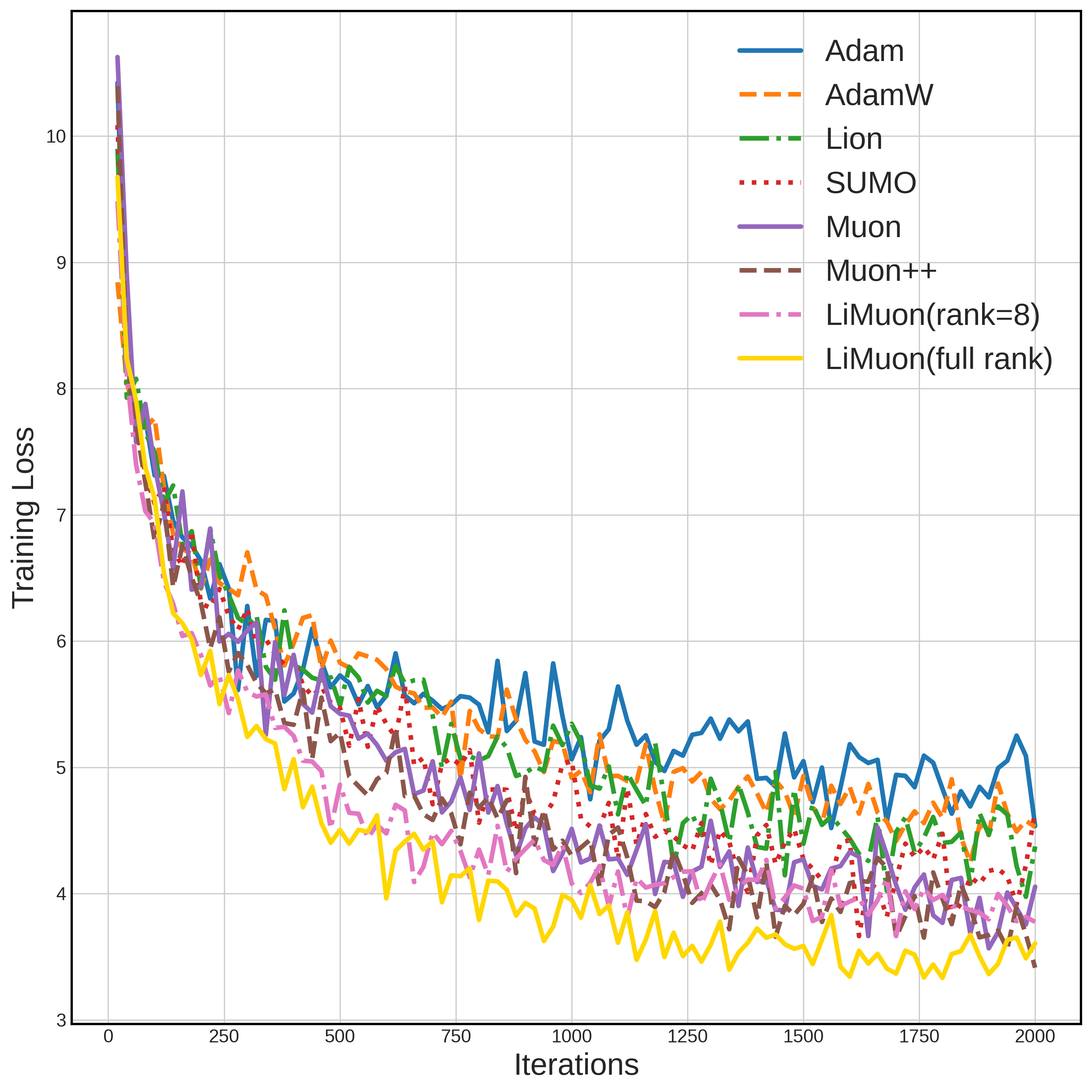}}
  \hfill
  \subfigure{\includegraphics[width=0.235\textwidth]{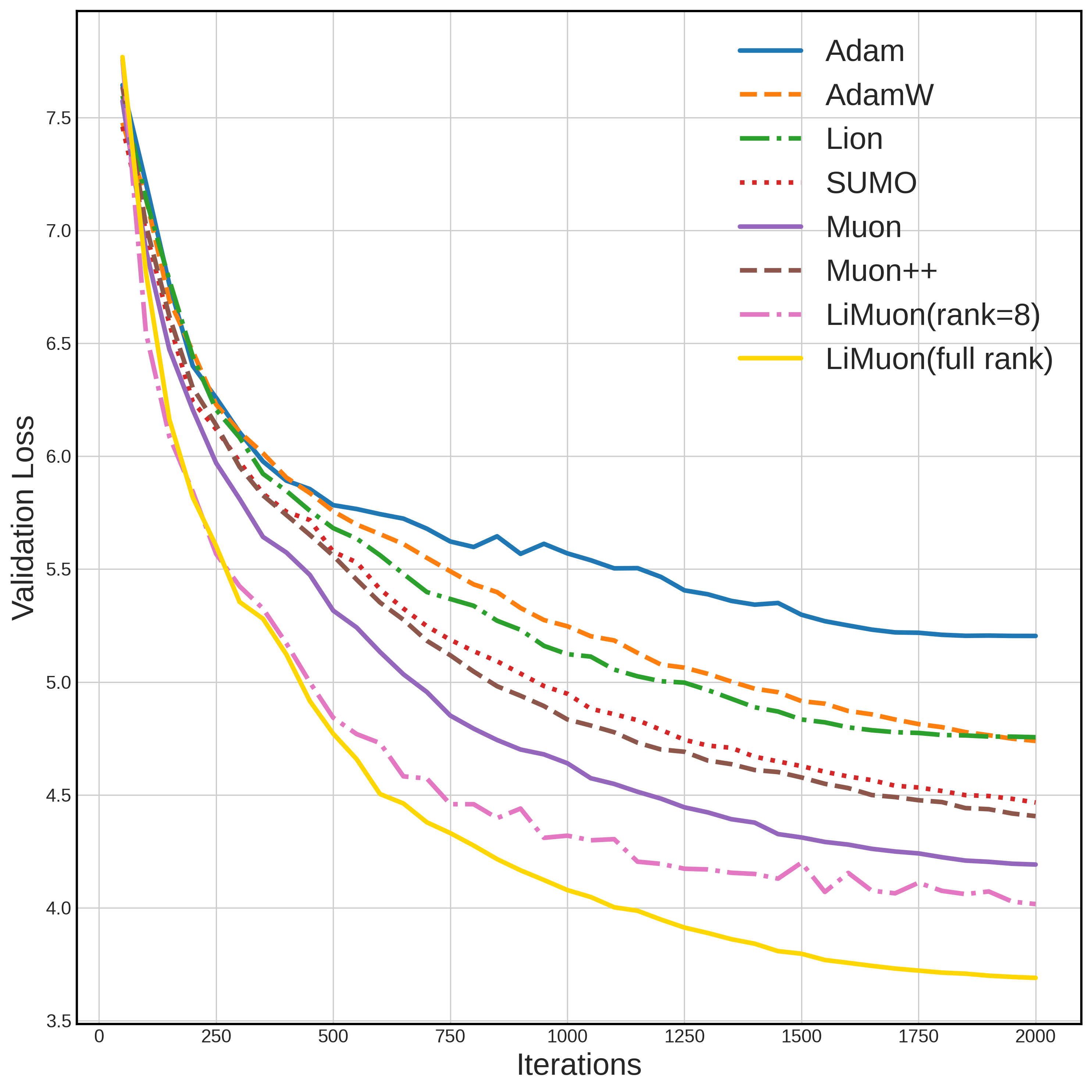}}
  \hfill
\caption{Performance comparison at Qwen-2.5-0.5B. }
\label{fig:qwen}
\end{figure}

\begin{table*}[!ht]
\centering
\caption{Performance comparison at ViT}
\label{tab:vit}
\resizebox{0.78\textwidth}{!}{
\begin{tabular}{lccccc}
\toprule
Method & Memory (GB) $\downarrow$ & \begin{tabular}[c]{@{}c@{}}Training Top-1 \\ Accuracy(\%) $\uparrow$ \end{tabular} & \begin{tabular}[c]{@{}c@{}}Training Top-5 \\ Accuracy(\%) $\uparrow$ \end{tabular} & \begin{tabular}[c]{@{}c@{}}Validation Top-1 \\ Accuracy(\%) $\uparrow$ \end{tabular} & \begin{tabular}[c]{@{}c@{}}Validation Top-5 \\ Accuracy(\%) $\uparrow$ \end{tabular} \\ \midrule
Adam             & 5.57 & 37.47 & 65.45 & 32.91 & 59.96 \\
AdamW            & 5.58 & 38.41 & 65.56 & 35.85 & 63.96 \\
Lion             & 5.44 & 40.97 & 68.55 & 40.21 & 68.02 \\
SUMO             & 5.31 & 51.23 & 78.08 & 44.23 & 70.19 \\
Muon             & 5.50 & 78.96 & 93.64 & 47.87 & 73.17 \\
Muon++           & 5.52 & 60.44 & 83.52 & 43.62 & 69.94 \\
\midrule
LiMuon (rank=4)  & \textbf{5.10} & 64.73 & 85.19 & 44.35 & 70.86 \\
LiMuon (rank=8)  & \textbf{5.28} & 71.94 & 90.06 & 46.75 & 72.17 \\
LiMuon (rank=16) & 5.44 & \textbf{80.58} & \textbf{93.91} & \textbf{47.98} & \textbf{73.22} \\
LiMuon (full rank) & 5.53 & \textbf{81.99} & \textbf{94.64} & \textbf{48.04} & \textbf{73.58} \\ \bottomrule
\end{tabular}
}
\end{table*}

\begin{table}[t]
	\caption{Runtime analysis on ViT.
		}
	\label{tab:runtime}
	\centering
	\begin{tabular}{lcc}
		\toprule
		\textbf{Optimizer}
		& \textbf{Time.\ Step}  $\downarrow$
		& \textbf{Time to Target} $\downarrow$ \\
		\midrule
		\multicolumn{3}{l}{\textit{Baseline optimizers}} \\
		Adam               & $5.58 ms$ & $15min 15s$ \\
		AdamW              & $5.66 ms$ & $13min 11s$ \\
		Lion               & $4.81 ms$ & $12min 37s$ \\
		\midrule
		Muon               & $23.43 ms$ & $7min 6s$ \\
		Muon++             & $24.50 ms$ & $9 min 54s$ \\
		SUMO              & $22.08 ms$ & $11min 20s$ \\
		\midrule
		\multicolumn{3}{l}{\textit{LiMuon (ours)}} \\
		LiMuon ($\hat{r}{=}4$)   & $24.75ms $ & $12min 4s$ \\
		LiMuon ($\hat{r}{=}8$)   & $25.49ms $ & $10min 29s$ \\
		LiMuon ($\hat{r}{=}16$)  & $26.87ms $ & $10min 51s$ \\
		LiMuon (full rank) & $23.37ms $ & $\textbf{7min 1s}$ \\
		\bottomrule
	\end{tabular}
\end{table}

\subsection{Pre-training Mamba-130M Model}
In this subsection, we evaluate our LiMuon optimizer against baseline methods on the Mamba-130M~\citep{gu2023mamba} model, a state space model (SSM) architecture with approximately 130 million parameters. Unlike Transformer-based models that rely on attention mechanisms with quadratic complexity, Mamba employs selective state spaces for linear-time sequence modeling, representing an emerging paradigm in efficient language modeling. We train the Mamba-130M model from scratch on the WikiText-103~\citep{merity2016pointer} dataset, which contains over 100 million tokens.

In the experiment, training Mamba-130M is conducted for 5,000 iterations with a global batch size of 64 and a sequence length of 256. All algorithms are individually tuned for optimal performance on this architecture. The detailed hyper-parameters used in all algorithms are given in the following Appendix~\ref{sec:hypara}. 
 We evaluate LiMuon under both full-rank and low-rank settings with $\hat{r} \in \{4, 8, 16\}$. For all Muon and its variants, we employ Newton-Schulz iteration with 10 steps, and use Nesterov momentum. In the experiments, we choose oversampling parameter $s=5$ or $s=10$, which is a standard recommendation in randomized SVD~\citep{halko2011finding}.
 
As shown in Figure~\ref{fig:mamba} and Table~\ref{tab:mamba}, our LiMuon optimizer demonstrates strong performance on this SSM-based architecture. Our full-rank LiMuon achieves the lowest perplexity (PPL), while providing more stable training dynamics. Notably, our low-rank LiMuon  (rank=8) achieves competitive language modeling performance with substantially reduced memory footprint, making them particularly suitable for training larger SSM models under memory constraints. These results demonstrate that LiMuon generalizes effectively beyond Transformer architectures to emerging efficient sequence models.
\subsection{Pre-training Qwen2.5-0.5B Model}
In this subsection, we extend our evaluation to the Qwen2.5-0.5B~\citep{qwen2.5} model, an advanced Transformer-based large language model with approximately 494 million parameters. Based on a deep architecture of 24 transformer layers with a hidden size of 896, the model utilizes 14 attention heads and 2 key-value heads, adopting a modern design incorporating Grouped Query Attention (GQA) and SwiGLU activation functions to achieve high efficiency and performance. In the experiment, we train the Qwen2.5-0.5B model from scratch on the MiniPile~\citep{kaddour2023minipile} dataset, a curated diverse text corpus containing approximately
300 million tokens.

In the experiment, training Qwen2.5-0.5B is conducted for 2000 iterations with an effective global batch size of 16 and a sequence length of 1024. All algorithms are individually tuned for optimal performance. The detailed hyper-parameters used in all algorithms are given in the following Appendix~\ref{sec:hypara}. 
We evaluate LiMuon under both full-rank and low-rank settings with $\hat{r} \in \{4, 8, 16\}$. For all Muon and its variants, we employ Newton-Schulz iteration with 10 steps, and use Nesterov momentum.

As detailed in Figure~\ref{fig:qwen} and Table~\ref{tab:qwen}, our LiMuon optimizer demonstrates superior performance on this modern LLM architecture. Our full-rank LiMuon achieves an optimal performance, establishing a new benchmark, while the low-rank LiMuon (rank=8) significantly reduce memory cost during optimization without compromising training stability. This result highlights versatility of our LiMuon in handling advanced Transformer architectures with complex components like GQA.
\vspace*{-6pt}
\subsection{Pre-training ViT Model}
In this subsection, we extend our evaluation to the Vision Transformer (ViT) model, a widely adopted Transformer-based architecture for image classification with approximately 22 million parameters. The model follows the standard ViT configuration with a hidden size of 384, 12 transformer layers, 6 attention heads, and a patch size of $16\times16$. We train this ViT model from scratch on the Tiny-ImageNet~\citep{le2015tiny} dataset, which contains 100,000 training images and 10,000 validation images across 200 classes, with images resized to $224\times 224$ to match the model's input requirements.

In the experiment, training the ViT is conducted for 10,000 iterations with a batch size of 128. All algorithms are individually tuned for optimal performance. The detailed hyper-parameters used in all algorithms are given in the following Appendix~\ref{sec:hypara}. 
We evaluate LiMuon under both full-rank and low-rank settings with $\hat{r} \in \{4, 8, 16\}$. For all Muon and its variants, we employ Newton-Schulz iteration with 10 steps, and use Nesterov momentum. 

As detailed in Table~\ref{tab:vit}, our LiMuon optimizer demonstrates competitive performance on this standard vision architecture. Our full-rank LiMuon achieves the highest classification accuracy, while our low-rank LiMuon (rank=8) significantly reduce memory cost during optimization without compromising training stability.
These results highlight versatility of our LiMuon optimizer, confirming its effectiveness not only in Large Language Models but also in Computer Vision architectures trained from scratch.

In Table~\ref{tab:runtime}, \text{Time. Step} is the wall-clock time of updating variable, which does not include time of computing gradients via backpropagation. \text{Time to Target} is the wall-clock time to first reach a target loss, which include time of computing gradients via backpropagation. Here we give a target loss $L^*= 3.70$ (an example, the other target losses have similar conclusions). From Table~\ref{tab:runtime}, our full rank LiMuon uses the shortest amount of time to reach the target loss.
Meanwhile, our low-rank LiMuon (rank=8) uses a comparable time to reach the target loss under less memory usage.
\vspace*{-6pt}
\section{ Conclusion}
In the paper, we proposed a light and fast Muon (LiMuon) optimizer, which \emph{simultaneously} has a lower memory and
lower sample complexity than the Muon and its variants.
Moreover, we studied the convergence properties of our LiMuon optimizer with exact SVD and Newton-Schulz steps, respectively. 
In particular, we proved that our LiMuon with Newton-Schulz obtains a lower sample  complexity of $O(\chi_{q}^3\epsilon^{-3})$ under a lower memory than $O(\chi_{q}^4\epsilon^{-4})$ of the Muon with Newton-Schulz~\citep{kim2026convergence}. Some numerical experimental results on pre-training
large models to demonstrate effectiveness of our LiMuon optimizer.

%
%
%
%

\section*{Acknowledgements}

We thank the anonymous reviewers for their helpful comments. This paper was partially supported by NSFC under Grant No. 62376125.

%
%

\section*{Impact Statement}
%
%
%

This paper presents work whose goal is to advance the field of Machine
Learning. There are many potential societal consequences of our work, none
which we feel must be specifically highlighted here.

\nocite{langley00}

\bibliography{LiMuon}
\bibliographystyle{icml2026}

\newpage
\appendix
\onecolumn

\section{Convergence Analysis of Our LiMuon \textcolor{blue}{with Exact SVD} }
In the section, we provide the convergence analysis for our LiMuon algorithm with exact SVD (\emph{using in orthogonalization process}) under some mild conditions.
We first give some useful lemmas.

\begin{lemma} \label{lem:A1}
For random variables $\xi_1, \cdots, \xi_N$ are independent
and its mean is zero, we have
 \begin{align}
  \E [ \|\xi_1 + \cdots +\xi_N\|^2 ] = \E [ \|\xi_1\|^2 + \cdots + \|\xi_N\|^2].
 \end{align}
\end{lemma}

\begin{lemma} \label{lem:A2}
For a random variable $\xi$, we have
 \begin{align}
  \E \|\xi-\E[\xi]\|^2 = \E\|\xi\|^2 - (\E[\xi])^2 \leq \E\|\xi\|^2.
 \end{align}
\end{lemma}

\begin{lemma}[Approximation error bound of RSVD~\citep{halko2011finding}] \label{lem:A3}
Let \(A \in \mathbb{R}^{m \times n} \) have singular values \( \nu_1 \geq \nu_2 \geq \cdots \). For a target rank \( \hat{r} \geq 2\) and oversampling parameter \( s\geq 2\), and  \(  \hat{r} + s \leq \min(m,n)\). Then the randomized SVD (Algorithm~\ref{alg:2}) produces an approximation \( \hat{A}\) such that
\begin{equation}\label{eq:rsvd_error}
\mathbb{E} \| A - \hat{A} \|_\F\leq \left(1 + \frac{\hat{r}}{s - 1} \right)^{\frac{1}{2}} (\sum_{j > \hat{r}} \nu_j^2)^{\frac{1}{2}}.
\end{equation}
\end{lemma}

\begin{lemma}\label{lem:A4}
The matrices $M_t$ and $\hat{M}_t$ are generalized from Algorithm~\ref{alg:1} or Algorithm~\ref{alg:3} , given $s\geq 2$, $\hat{r}\geq 2$ and $s+\hat{r}\leq r=\min(m,n)$, we have
\begin{align*}
    \|\hat{M}_t - M_t\|_\F\leq \gamma\|M_t\|_\F, \ \forall t\geq0
\end{align*}
where $\gamma=\tau\left(1 + \frac{\hat{r}}{s - 1} \right)^{\frac{1}{2}} $ and $\tau\in (0,1)$.
\end{lemma}

\begin{proof}
 Suppose matrix $M_t$ has singular values $\nu^t_1\geq\nu^t_2\geq\cdots$. By using 
 Lemma~\ref{lem:A3}, we have
 \begin{align*}
 \|\hat{M}_t- M_t\|_\F\leq \left(1 + \frac{\hat{r}}{s - 1} \right)^{\frac{1}{2}}(\sum_{j > \hat{r}} (\nu^t_j)^2)^{\frac{1}{2}} \leq\left(1 + \frac{\hat{r}}{s - 1} \right)^{\frac{1}{2}} \tau\|M_t\|_\F,
 \end{align*}
 where the last inequality holds by $(\sum_{j > \hat{r}} (\nu^t_j)^2)^{\frac{1}{2}} \leq \tau(\sum_{j\geq1}(\nu^t_j)^2)^{\frac{1}{2}}=\tau\|M_t\|_\F $ with $\tau\in (0,1)$. 
 Let $\gamma=\tau\left(1 + \frac{\hat{r}}{s - 1} \right)^{\frac{1}{2}}$,
 we have 
  \begin{align*}
 	\|\hat{M}_t- M_t\|_\F\leq \gamma\|M_t\|_\F.
 \end{align*}
 
\end{proof}

\begin{lemma}\label{lem:A5}
Under the Assumption~\ref{ass:s2}, we have for any $W,W'\in \R^{m\times n}$
\begin{align*}
    F(W') \leq F(W)+\langle \nabla F(W),W'-W\rangle+\frac{L_{0}+L_{1}\|\nabla F(W)\|_\F}{2}\|W'-W\|_\F^{2}.
\end{align*}
\end{lemma}
\begin{proof}
By using Assumption~\ref{ass:s2}, we have
\begin{align} \label{eq:A0}
     \|\nabla F(W)-\nabla F(W')\|_\F
    & \leq \sqrt{(L_0^2+L_1^2\|\nabla F(W)\|_\F^2)\|W-W'\|_\F^2} \nonumber \\
    & \leq (L_0+L_1\|\nabla F(W)\|_\F)\|W-W'\|_\F.
\end{align}
Since $F(W)$ is differentiable, by applying the fundamental theorem
of calculus to $G(t)= F(W+t(W'-W))$, we have
\begin{align}
F(W') & =F(W)+\int_{0}^{1}\langle \nabla F(W+t(W'-W)),W'-W\rangle\mathrm{d}t  \nonumber \\
 & =F(W)+\langle \nabla F(W),W'-W\rangle+\int_{0}^{1}\langle \nabla F(W+t(W'-W))- \nabla F(W),W'-W\rangle\mathrm{d}t  \nonumber \\
 & \overset{(i)}{\leq}F(W)+\langle \nabla F(W),W'-W\rangle+\int_{0}^{1}\| \nabla F(W+t(W'-W))- \nabla F(W)\|_\F\|W'-W\|_\F\mathrm{d}t \nonumber \\
 & \overset{(ii)}{\leq}F(W)+\langle \nabla F(W),W'-W\rangle+\int_{0}^{1}(L_{0}+L_{1}\| \nabla F(W)\|_\F)\|W'-W\|_\F^{2} t\mathrm{d}t  \nonumber \\
 & =F(W)+\langle \nabla F(W),W'-W\rangle+\frac{L_{0}+L_{1}\|\nabla F(W)\|_\F}{2}\|W'-W\|_\F^{2},
\end{align}
where the above inequality $(i)$ is by Cauchy-Schwarz inequality and the inequality $(ii)$ is due to
the above inequality~(\ref{eq:A0}).
\end{proof}

For notational simplicity, let $\zeta_t = M_t-\nabla F(W_t)$, $Z_t = \nabla F(W_t)-\nabla F(W_{t-1})-\big(\nabla f(W_t;\xi_t)- \nabla f(W_{t-1};\xi_t)\big)$ and $\delta_t =\nabla F(W_t)- \nabla f(W_t;\xi_t) $ for all $t\geq1$.

\subsection{Convergence Analysis of LiMuon Algorithm \textcolor{blue}{with Exact SVD and Option \#1 under Lipschitz Smoothness Assumption}}
In this subsection, we provide the convergence analysis of Algorithm~\ref{alg:1} with \textbf{Option} \#1 under Lipschitz smoothness assumption.

\begin{lemma} \label{lem:B1}
 Assume the stochastic gradient estimate $M_t$ be generated from Algorithm \ref{alg:1} under  \textbf{Option \#1}. Let $\zeta_t = M_t-\nabla F(W_t)$, $\eta_t=\eta$ and $\beta_t=\beta$ for all $t\geq 0$, under the above Assumption~\ref{ass:s1}, we have
 \begin{align}
 \E\|\zeta_t\|_\F \leq (1-\beta)^t\sigma + \sqrt{\frac{\beta}{2-\beta}}\sigma + \sqrt{\frac{r(1-\beta)^2}{(2-\beta)\beta}}L\eta.
 \end{align}
\end{lemma}

\begin{proof}
 By the definition of $M_t$ in Algorithm \ref{alg:1} under Option \#1, we have
 \begin{align}
  M_t - M_{t-1} = -\beta_tM_{t-1}+ \beta_t \nabla f(W_t;\xi_t) + (1-\beta_t)\big( \nabla f(W_t;\xi_t)- \nabla f(W_{t-1};\xi_t)\big).
 \end{align}
 Let $Z_t = \nabla f(W_t;\xi_t)- \nabla f(W_{t-1};\xi_t)-(\nabla F(W_t)-\nabla F(W_{t-1}))$,
 $\delta_t =\nabla F(W_t)- \nabla f(W_t;\xi_t) $, and $\beta_t=\beta$, we have
 \begin{align}
 \zeta_t & = \nabla F(W_t)- M_t \nonumber \\
 & = \nabla F(W_{t-1}) - M_{t-1} + \nabla F(W_t)-\nabla F(W_{t-1}) -(M_t-M_{t-1}) \nonumber \\
 & = (1-\beta)(\nabla F(W_{t-1}) - M_{t-1}) + \beta(\nabla F(W_t)- \nabla f(W_t;\xi_t))\nonumber \\
  & \quad \quad - (1-\beta)\big( \nabla f(W_t;\xi_t)- \nabla f(W_{t-1};\xi_t)-(\nabla F(W_t)-\nabla F(W_{t-1}))\big) \nonumber \\
  & = (1-\beta)\zeta_{t-1} + \beta\delta_t + (1-\beta)Z_t
 \end{align}
 By expanding the recursion, then we can obtain
 \begin{align}
 \zeta_t = (1-\beta)^t\zeta_{0} + \beta\sum_{s=1}^{t}(1-\beta)^{t-s}\delta_s +(1-\beta)\sum_{s=1}^{t}(1-\beta)^{t-s}Z_s.
 \end{align}

Since $M_{0}=\nabla f(W_0;\xi_0)$, we have
 \begin{align} \label{eq:A1}
 \E\|\zeta_{0}\|_\F = \E\|\nabla f(W_0;\xi_0)-\nabla F(W_0)\|_\F & = \sqrt{\big(\E\|\nabla f(W_0;\xi_0)-\nabla F(W_0)\|_\F\big)^2} \nonumber \\
 & \mathop{\leq}^{(i)} \sqrt{\E\|\nabla f(W_0;\xi_0)-\nabla F(W_0)\|_\F^2} \mathop{\leq}^{(ii)} \sqrt{\sigma^2} =\sigma,
 \end{align}
where the above inequality (i) is due to Jensen inequality, and the above inequality (ii) holds by Assumption~\ref{ass:v}.

Meanwhile, we have
 \begin{align} \label{eq:A2}
  \beta\E\|\sum_{s=1}^{t}(1-\beta)^{t-s}\delta_s\|_\F & = \sqrt{\beta^2\big(\E\big\|\sum_{s=1}^{t}(1-\beta)^{t-s}\delta_s \big\|_\F \big)^2}  \nonumber \\
  & \leq \sqrt{\beta^2\E\big\|\sum_{s=1}^{t}(1-\beta)^{t-s}\delta_s \big\|^2_\F} \nonumber \\
  & \mathop{=}^{(i)} \sqrt{\beta^2\sum_{s=1}^{t}(1-\beta)^{2(t-s)}\E\big\|\nabla F(W_s)- \nabla f(W_s;\xi_s)\big\|^2_\F} \nonumber \\
  & \leq\sqrt{ \beta^2\sum_{s=1}^{t}(1-\beta)^{2(t-s)} \sigma^2} \leq \sqrt{\frac{\beta^2}{1-(1-\beta)^2}\sigma^2} = \sqrt{\frac{\beta}{2-\beta}}\sigma,
 \end{align}
 where the above equality (i) holds by the above Lemma~\ref{lem:A1} on the fact that  $\{\xi_1,\cdots,\xi_t\}$ are independent random variables and $\E_{\xi_s}[\nabla F(W_s)- \nabla f(W_s;\xi_s)]=0$ for all $1\leq s\leq t$.

We also can obtain
 \begin{align} \label{eq:A3}
  (1-\beta)\E\|\sum_{s=1}^{t}(1-\beta)^{t-s}Z_s\|_\F & = \sqrt{(1-\beta)^2\big(\E\|\sum_{s=1}^{t}(1-\beta)^{t-s}Z_s\|_\F \big)^2} \nonumber \\
  & \leq  \sqrt{(1-\beta)^2\E\|\sum_{s=1}^{t}(1-\beta)^{t-s}Z_s\|^2_\F} \nonumber \\
  & \mathop{=}^{(i)} \sqrt{(1-\beta)^2\sum_{s=1}^{t}(1-\beta)^{2(t-s)}\E\|\nabla f(W_s;\xi_s)- \nabla f(W_{s-1};\xi_s)-(\nabla F(W_s)-\nabla F(W_{s-1}))\|^2_\F} \nonumber \\
  & \mathop{\leq}^{(ii)} \sqrt{(1-\beta)^2\sum_{s=1}^{t}(1-\beta)^{2(t-s)}\E\|\nabla f(W_s;\xi_s)- \nabla f(W_{s-1};\xi_s)\|^2_\F } \nonumber \\
   & \mathop{\leq}^{(iii)} \sqrt{(1-\beta)^2\sum_{s=1}^{t}(1-\beta)^{2(t-s)}L^2\E\|W_s- W_{s-1}\|^2_\F} \nonumber \\
   & \leq \sqrt{(1-\beta)^2\sum_{s=1}^{t}(1-\beta)^{2(t-s)}L^2\E\|\eta U_{t-1}V_{t-1}^\top\|^2_\F} \nonumber \\
   & \leq \sqrt{(1-\beta)^2\sum_{s=1}^{t}(1-\beta)^{2(t-s)}L^2\eta^2r} \nonumber \\
   & \leq \sqrt{(1-\beta)^2\frac{1}{1-(1-\beta)^2}L^2\eta^2r} = \sqrt{\frac{r(1-\beta)^2}{(2-\beta)\beta}}L\eta,
 \end{align}
 where the above equality (i) holds by the above Lemma~\ref{lem:A1} on the fact that$\{\xi_1,\cdots,\xi_t\}$ are independent random variables and $\E_{\xi_s}[\nabla f(W_s;\xi_s)- \nabla f(W_{s-1};\xi_s)-(\nabla F(W_s)-\nabla F(W_{s-1}))]=0$ for all $1\leq s\leq t$; the above inequality (ii) holds by the above lemma~\ref{lem:A2}, and the above inequality (iii) is due to Assumption~\ref{ass:s1}, and the second last inequality is due to $\|U_{t-1}V_{t-1}^\top\|\leq \sqrt{r}$.

 By using the above inequalities~(\ref{eq:A1}), (\ref{eq:A2}) and (\ref{eq:A3}), we have
 \begin{align}
 \E\|\zeta_t\|_\F
 & = \E\|(1-\beta)^t\zeta_{0} + \beta\sum_{s=1}^{t}(1-\beta)^{t-s}\delta_s -(1-\beta)\sum_{s=1}^{t}(1-\beta)^{t-s}Z_s\|_\F \nonumber \\
 & \leq (1-\beta)^t\E\|\zeta_{0}\|_\F + \beta\E\|\sum_{s=1}^{t}(1-\beta)^{t-s}\delta_s\|_\F +(1-\beta)\E\|\sum_{s=1}^{t}(1-\beta)^{t-s}Z_s\|_\F \nonumber \\
 & \leq (1-\beta)^t\sigma + \sqrt{\frac{\beta}{2-\beta}}\sigma + \sqrt{\frac{r(1-\beta)^2}{(2-\beta)\beta}}L\eta .
 \end{align}

\end{proof}

\begin{theorem} \label{th:B1}
Under the above Assumptions~\ref{ass:s1},~\ref{ass:v},~\ref{ass:l}, the sequence $\{W_t\}_{t=0}^T$ be generated
from Algorithm \ref{alg:1} with \textbf{Option} \#1. Let $\beta_t=\beta \in (0,1)$ and $\eta_t=\eta$ for all $t\geq 0$, we have
\begin{align}
 \frac{1}{T+1}\sum_{t=0}^{T}\E \|\nabla F(W_t)\|_* \leq \frac{F(W_0)- F^*}{T\eta}  + \frac{r L\eta}{2} + 2\sqrt{r}\big(\frac{\sigma}{T\beta} + \sqrt{\frac{\beta}{2-\beta}}\sigma + \sqrt{\frac{(1-\beta)^2}{(2-\beta)\beta}}L\eta r\big),
\end{align}
 where $r= \min(m,n)$.
\end{theorem}

\begin{proof}
Since $M_t=U_t\Sigma_tV_t^\top$, we have
\begin{align}
 \langle M_t , U_tV_t^\top \rangle & = \langle U_t\Sigma_tV_t^\top , U_tV_t^\top \rangle = \mbox{tr}\big((U_t\Sigma_tV_t^\top)^\top U_tV_t^\top\big)= \mbox{tr}\big(V_t\Sigma_tU_t^\top U_tV_t^\top\big) \nonumber \\
 & = \mbox{tr}\big(V_t\Sigma_tV_t^\top\big)=\mbox{tr}\big(V_t^\top V_t\Sigma_t\big)=\mbox{tr}\big(\Sigma_t\big)=\|M_t\|_*.
\end{align}

 By using Assumption~\ref{ass:s1}, i.e., $f$ is $L$ Frobenius norm Lipschitz smooth, we have
\begin{align}
    F(W_{t+1}) \leq & F(W_t)+ \langle\nabla F(W_t), W_{t+1}-W_t \rangle+\frac{L}{2}\|W_{t+1}-W_t \|_\F^2 \nonumber \\
    \leq & F(W_t)- \eta_t\langle\nabla F(W_t), U_tV_t^\top \rangle+\frac{L}{2}\eta_t^2\|U_tV_t^\top\|_\F^2 \nonumber \\
    \leq & F(W_t) - \eta_t\langle M_t, U_tV_t^\top \rangle + \frac{r L}{2}\eta_t^2 -\eta_t\langle \nabla F(W_t)-M_t , U_tV_t^\top\rangle \nonumber \\
    \leq& F(W_t)- \eta_t\langle M_t , U_tV_t^\top \rangle + \frac{rL}{2}\eta_t^2 + \eta_t\| \nabla F(W_t)-M_t \|_\F \|U_tV_t^\top\|_\F]  \nonumber  \\
    \mathop{\leq}^{(i)} & F(W_t)- \eta_t\|M_t\|_* + \frac{r L}{2}\eta_t^2 + \eta_t\sqrt{r}\|\nabla F(W_t)-M_t\|_\F \nonumber \\
    \leq & F(W_t)-  \eta_t \|\nabla F(W_t)\|_* + \frac{r L}{2}\eta_t^2 + 2\eta_t\sqrt{r}\|\nabla F(W_t)-M_t\|_\F,
\end{align}
where the above inequality (i) is due to $\langle M_t , U_tV_t^\top \rangle =\|M_t\|_*$ and $\|U_tV_t^\top\|_\F \leq \sqrt{r}$, and the last inequality holds by
$\|M_t\|_*\geq \|\nabla F(W_t)\|_* - \|\nabla F(W_t)-M_t\|_*\geq \|\nabla F(W_t)\|_* - \sqrt{r}\|\nabla F(W_t)-M_t\|_\F$ and $\|M_t\|_\F\leq \|M_t\|_*\leq \sqrt{r}\|M_t\|_\F$.

Let $\eta_t=\eta$, then we have
\begin{align}
    \E\|\nabla F(W_t)\|_* \leq & \E\big[\frac{F(W_t)- F(W_{t+1})}{\eta}  + \frac{r L\eta}{2} + 2\sqrt{r}\|\zeta_t\|_\F \big].
\end{align}

Thus we can obtain
 \begin{align}
    \frac{1}{T+1}\sum_{t=0}^{T}\E \|\nabla F(W_t)\|_* \leq & \E\big[ \frac{1}{T+1}\sum_{t=0}^{T}\frac{F(W_t)- F(W_{t+1})}{\eta}  + \frac{1}{T+1}\sum_{t=0}^{T} \frac{r L\eta}{2} + \frac{1}{T+1}\sum_{t=0}^{T}2\sqrt{r}\|\zeta_t\|_\F \big] \nonumber \\
    & \mathop{\leq}^{(i)} \frac{F(W_0)- F^*}{(T+1)\eta}  + \frac{r L\eta}{2} + \frac{1}{T+1}\sum_{t=0}^{T}2\sqrt{r}\E \|\zeta_t\|_\F \nonumber \\
    & \mathop{\leq}^{(ii)} \frac{F(W_0)- F^*}{(T+1)\eta}  + \frac{r L\eta}{2} + \frac{2\sqrt{r}}{T+1}\sum_{t=0}^{T}\big((1-\beta)^t\sigma + \sqrt{\frac{\beta}{2-\beta}}\sigma + \sqrt{\frac{r(1-\beta)^2}{(2-\beta)\beta}}L\eta\big) \nonumber \\
    & \leq \frac{F(W_0)- F^*}{T\eta}  + \frac{r L\eta}{2} + 2\sqrt{r}\big(\frac{\sigma}{T\beta} + \sqrt{\frac{\beta}{2-\beta}}\sigma + \sqrt{\frac{r(1-\beta)^2}{(2-\beta)\beta}}L\eta\big),
\end{align}
where the above inequality (i) is due to Assumption~\ref{ass:l}, and the above inequality (ii) holds by the above lemma~\ref{lem:B1}.

Let $\eta=O(\frac{1}{T^{2/3}})$ and $\beta=O(\frac{1}{T^{2/3}})$, we have
\begin{align}
    \frac{1}{T+1}\sum_{t=0}^{T} \E \|\nabla F(W_t)\|_*
    & \leq \frac{F(W_0)- F^*}{T\eta}  + \frac{r L\eta}{2} + 2\sqrt{r}\big(\frac{\sigma}{T\beta} + \sqrt{\frac{\beta}{2-\beta}}\sigma + \sqrt{\frac{r(1-\beta)^2}{(2-\beta)\beta}}L\eta\big) \nonumber \\
    & = O\big(\frac{1}{T^{1/3}}+ \frac{1}{T^{2/3}}+ \frac{1}{T^{1/3}}+ \frac{1}{T^{1/3}}+ \frac{1}{T^{1/3}}\big) =O\big(\frac{1}{T^{1/3}}\big)
\end{align}

\end{proof}

\subsection{Convergence Analysis of LiMuon Algorithm \textcolor{blue}{with Exact SVD and Option \#1 under Generalized Smoothness Assumption}}
In this subsection, we provide the convergence analysis of Algorithm~\ref{alg:1} with \textbf{Option} \#1 under \textbf{generalized smoothness} assumption.

\begin{lemma} \label{lem:C1}
 Assume the stochastic gradient estimate $M_t$ be generated from Algorithm \ref{alg:1} under  \textbf{Option \#1}. Let $\zeta_t = M_t-\nabla F(W_t)$, $\eta_t=\eta$ and $\beta_t=\beta$ for all $t\geq 0$, under the assumption~\ref{ass:s2}, we have
 \begin{align}
 \E\|\zeta_t\|_\F \leq (1-\beta)^t\sigma + \sqrt{\frac{\beta}{2-\beta}}\sigma + \sqrt{\frac{r(1-\beta)^2}{(2-\beta)\beta}}L_0\eta + L_1(1-\beta)\eta \sqrt{r}\sum_{s=1}^{t}(1-\beta)^{(t-s)}\E\|\nabla F(W_{s-1})\|_\F.
 \end{align}
\end{lemma}

\begin{proof}
By using Assumption~\ref{ass:s2}, we have
\begin{align} \label{eq:B1}
  &(1-\beta)\E\|\sum_{s=1}^{t}(1-\beta)^{t-s}Z_s\|_\F  \nonumber \\
  & = \sqrt{(1-\beta)^2\big(\E\|\sum_{s=1}^{t}(1-\beta)^{t-s}Z_s\|_\F \big)^2} \nonumber \\
  & \mathop{\leq}^{(i)}  \sqrt{(1-\beta)^2\E\|\sum_{s=1}^{t}(1-\beta)^{t-s}Z_s\|^2_\F} \nonumber \\
  & \mathop{=}^{(ii)} \sqrt{(1-\beta)^2\sum_{s=1}^{t}(1-\beta)^{2(t-s)}\E\|\nabla f(W_s;\xi_s)- \nabla f(W_{s-1};\xi_s)-(\nabla F(W_s)-\nabla F(W_{s-1}))\|^2_\F} \nonumber \\
  & \leq \sqrt{(1-\beta)^2\sum_{s=1}^{t}(1-\beta)^{2(t-s)}\E\|\nabla f(W_s;\xi_s)- \nabla f(W_{s-1};\xi_s)\|^2_\F } \nonumber \\
   &  \mathop{\leq}^{(iii)} \sqrt{(1-\beta)^2\sum_{s=1}^{t}(1-\beta)^{2(t-s)}(L_0^2+L_1^2(\E\|\nabla F(W_{s-1})\|_\F)^2)\|W_s- W_{s-1}\|^2_\F} \nonumber \\
   & \leq \sqrt{(1-\beta)^2\sum_{s=1}^{t}(1-\beta)^{2(t-s)}(L_0^2+L_1^2(\E\|\nabla F(W_{s-1})\|_\F)^2)\|\eta U_{t-1}V_{t-1}^\top\|^2_\F} \nonumber \\
   & \leq \sqrt{(1-\beta)^2\sum_{s=1}^{t}(1-\beta)^{2(t-s)}(L_0^2+L_1^2(\E\|\nabla F(W_{s-1})\|_\F)^2)\eta^2r},
 \end{align}
 where the above inequality (i) is due to Jensen inequality, and the above equality (ii) holds by the above Lemma~\ref{lem:A1} on the fact that$\{\xi_1,\cdots,\xi_t\}$ are independent random variables and $\E_{\xi_s}[\nabla f(W_s;\xi_s)- \nabla f(W_{s-1};\xi_s)-(\nabla F(W_s)-\nabla F(W_{s-1}))]=0$ for all $1\leq s\leq t$, and the above inequality (iii) is due to Assumption~\ref{ass:s2}.

By using the inequality $\sqrt{a+b}\leq \sqrt{a}+\sqrt{b}$ for all $a,b\geq0$, we have
 \begin{align} \label{eq:B2}
  &\sqrt{(1-\beta)^2\sum_{s=1}^{t}(1-\beta)^{2(t-s)}(L_0^2+L_1^2(\E\|\nabla F(W_{s-1})\|_\F)^2)\eta^2r} \nonumber \\
   & \leq \sqrt{(1-\beta)^2\sum_{s=1}^{t}(1-\beta)^{2(t-s)}L_0^2\eta^2r} + \sqrt{(1-\beta)^2\sum_{s=1}^{t}(1-\beta)^{2(t-s)}L_1^2(\E\|\nabla F(W_{s-1})\|_\F)^2\eta^2r} \nonumber \\
   & \leq \sqrt{(1-\beta)^2\frac{1}{1-(1-\beta)^2}L_0^2\eta^2r} + \sqrt{(1-\beta)^2\sum_{s=1}^{t}(1-\beta)^{2(t-s)}L_1^2(\E\|\nabla F(W_{s-1})\|_\F)^2\eta^2r} \nonumber \\
   & =  \sqrt{\frac{r(1-\beta)^2}{(2-\beta)\beta}}L_0\eta + \sqrt{(1-\beta)^2\sum_{s=1}^{t}(1-\beta)^{2(t-s)}L_1^2(\E\|\nabla F(W_{s-1})\|_\F)^2\eta^2r} \nonumber \\
   & \leq \sqrt{\frac{r(1-\beta)^2}{(2-\beta)\beta}}L_0\eta + L_1(1-\beta)\eta \sqrt{r}\sum_{s=1}^{t}(1-\beta)^{(t-s)}\E\|\nabla F(W_{s-1})\|_\F.
 \end{align}

 By using the above inequalities~(\ref{eq:B1}) and~(\ref{eq:B2}), we have
 \begin{align} \label{eq:B3}
  (1-\beta)\E\|\sum_{s=1}^{t}(1-\beta)^{t-s}Z_s\|_\F & \leq \sqrt{(1-\beta)^2\sum_{s=1}^{t}(1-\beta)^{2(t-s)}(L_0+L_1\E\|\nabla F(W_{s})\|_\F)^2\eta^2r} \nonumber \\
  & \leq \sqrt{\frac{r(1-\beta)^2}{(2-\beta)\beta}}L_0\eta  + L_1(1-\beta)\eta \sqrt{r}\sum_{s=1}^{t}(1-\beta)^{(t-s)}\E\|\nabla F(W_{s-1})\|_\F.
 \end{align}

From the above lemma~\ref{lem:B1}, we have
 \begin{align} \label{eq:B4}
 \E\|\zeta_{0}\|_\F \leq \sigma.
 \end{align}
We also have
 \begin{align} \label{eq:B5}
  \beta\E\|\sum_{s=1}^{t}(1-\beta)^{t-s}\delta_s\|_\F  \leq  \sqrt{\frac{\beta}{2-\beta}}\sigma.
 \end{align}

By using the above inequalities~(\ref{eq:B3}), (\ref{eq:B4}) and~(\ref{eq:B5}), we can obtain
\begin{align}
 \E\|\zeta_t\|_\F
 & = \E\|(1-\beta)^t\zeta_{0} + \beta\sum_{s=1}^{t}(1-\beta)^{t-s}\delta_s -(1-\beta)\sum_{s=1}^{t}(1-\beta)^{t-s}Z_s\|_\F \nonumber \\
 & \leq (1-\beta)^t\E\|\zeta_{0}\|_\F + \beta\E\|\sum_{s=1}^{t}(1-\beta)^{t-s}\delta_s\|_\F +(1-\beta)\E\|\sum_{s=1}^{t}(1-\beta)^{t-s}Z_s\|_\F \nonumber \\
 & \leq (1-\beta)^t\sigma + \sqrt{\frac{\beta}{2-\beta}}\sigma + \sqrt{\frac{r(1-\beta)^2}{(2-\beta)\beta}}L_0\eta  + L_1(1-\beta)\eta \sqrt{r}\sum_{s=1}^{t}(1-\beta)^{(t-s)}\E\|\nabla F(W_{s-1})\|_\F.
 \end{align}

\end{proof}

\begin{theorem} \label{th:C1}
Under the Assumptions~\ref{ass:s2},~\ref{ass:v},~\ref{ass:l}, the sequence $\{W_t\}_{t=0}^T$ be generated
from Algorithm \ref{alg:1} with \textbf{Option} \#1. Let $\beta_t=\beta \in (0,1)$ and $\eta_t=\eta \leq \min\big(\frac{1}{2L_1r},\frac{\beta}{8L_1r(1-\beta)}\big)$ for all $t\geq 0$, we have
\begin{align}
 \frac{1}{T+1}\sum_{t=0}^{T}\E \|\nabla F(W_t)\|_* \leq  \frac{2(F(W_0)- F^*)}{T\eta}  + r L_0\eta + \frac{4\sqrt{r}\sigma}{T\beta} + 4\sqrt{r}\sqrt{\frac{\beta}{2-\beta}}\sigma
    + 4L_0\eta r\sqrt{\frac{(1-\beta)^2}{(2-\beta)\beta}},
\end{align}
 where $r= \min(m,n)$.
\end{theorem}

\begin{proof}
 By using Assumption~\ref{ass:s2} and the above Lemma~\ref{lem:A5}, we have
\begin{align}
    F(W_{t+1}) \leq & F(W_t)+ \langle\nabla F(W_t), W_{t+1}-W_t \rangle+\frac{L_0+L_1\|\nabla F(W_t)\|_\F}{2}\|W_{t+1}-W_t \|_F^2 \nonumber \\
    \leq & F(W_t)- \eta_t\langle\nabla F(W_t), U_tV_t^\top \rangle+\frac{L_0+L_1\|\nabla F(W_t)\|_\F}{2}\eta_t^2\|U_tV_t^\top\|_\F^2 \nonumber \\
    \leq & F(W_t) - \eta_t\langle M_t, U_tV_t^\top \rangle + \frac{L_0+L_1\|\nabla F(W_t)\|_\F}{2}\eta_t^2 r -\eta_t\langle \nabla F(W_t)-M_t , U_tV_t^\top\rangle \nonumber \\
    \leq& F(W_t)- \eta_t\langle M_t , U_tV_t^\top \rangle + \frac{L_0+L_1\|\nabla F(W_t)\|_\F}{2}\eta_t^2 r + \eta_t\| \nabla F(W_t)-M_t \|_\F \|U_tV_t^\top\|_\F]  \nonumber  \\
    \mathop{\leq}^{(i)} & F(W_t)- \eta_t\|M_t\|_* + \frac{L_0+L_1\|\nabla F(W_t)\|_\F}{2}\eta_t^2r + \eta_t\sqrt{r}\|\nabla F(W_t)-M_t\|_\F \nonumber \\
    \leq & F(W_t)-  \eta_t \|\nabla F(W_t)\|_* + \frac{L_0+L_1\|\nabla F(W_t)\|_\F}{2}\eta_t^2r + 2\eta_t\sqrt{r}\|\nabla F(W_t)-M_t\|_\F,
\end{align}
where the above inequality (i) holds by $\langle M_t , U_tV_t^\top \rangle =\|M_t\|_*$ and $\|U_tV_t^\top\|_\F \leq \sqrt{r}$, and the last inequality is due to $\|M_t\|_*\geq \|\nabla F(W_t)\|_* - \|\nabla F(W_t)-M_t\|_*\geq \|\nabla F(W_t)\|_* - \sqrt{r}\|\nabla F(W_t)-M_t\|_\F$.

Let $\eta_t=\eta$, we have
\begin{align}
    \E\|\nabla F(W_t)\|_* & \leq \E\big[\frac{F(W_t)- F(W_{t+1})}{\eta}  + \frac{r L_0\eta}{2} +\frac{L_1\eta r}{2}\|\nabla F(W_t)\|_\F + 2\sqrt{r}\|\zeta_t\|_\F \big] \nonumber \\
   & \leq \E [\frac{F(W_t)- F(W_{t+1})}{\eta}]  + \frac{r L_0\eta}{2} +\frac{L_1\eta r}{2} \E\|\nabla F(W_t)\|_\F + 2\sqrt{r}(1-\beta)^t\sigma + 2\sqrt{r}\sqrt{\frac{\beta}{2-\beta}}\sigma \nonumber \\
   & \quad + \sqrt{\frac{(1-\beta)^2}{(2-\beta)\beta}}2L_0\eta r + 2L_1(1-\beta)\eta r\sum_{s=1}^{t}(1-\beta)^{(t-s)}\E\|\nabla F(W_{s})\|_F,
\end{align}
where the last inequality holds by the above lemma~\ref{lem:C1}.

Let $\nabla f(W_{-1})=0$, then we have
\begin{align}
    \frac{1}{T+1}\sum_{t=0}^{T}\E\|\nabla F(W_t)\|_*
   & \leq \frac{F(W_0)- F^*}{(T+1)\eta}  + \frac{r L_0\eta}{2} + \frac{L_1\eta r}{2}\frac{1}{T+1}\sum_{t=0}^{T}\E\|\nabla F(W_t)\|_\F + \frac{2\sqrt{r}\sigma}{T+1}\sum_{t=0}^{T}(1-\beta)^t + 2\sqrt{r}\sqrt{\frac{\beta}{2-\beta}}\sigma \nonumber \\
   & \quad + \sqrt{\frac{(1-\beta)^2}{(2-\beta)\beta}}2L_0\eta r + \frac{2L_1(1-\beta)\eta r }{T+1}\sum_{t=0}^{T}\sum_{s=1}^{t}(1-\beta)^{(t-s)}\E\|\nabla F(W_{s-1})\|_F \nonumber \\
   & \leq \frac{F(W_0)- F^*}{(T+1)\eta}  + \frac{r L_0\eta}{2} + \frac{L_1\eta r}{2}\frac{1}{T+1}\sum_{t=0}^{T}\E\|\nabla F(W_t)\|_\F + \frac{2\sqrt{r}\sigma}{(T+1)\beta} + 2\sqrt{r}\sqrt{\frac{\beta}{2-\beta}}\sigma \nonumber \\
   & \quad + \sqrt{\frac{(1-\beta)^2}{(2-\beta)\beta}}2L_0\eta r + \frac{2L_1(1-\beta)\eta r }{T+1}\sum_{s=0}^{T}\sum_{t=s}^{T}(1-\beta)^{(t-s)}\E\|\nabla F(W_{s-1})\|_F \nonumber \\
   & \leq \frac{F(W_0)- F^*}{(T+1)\eta}  + \frac{r L_0\eta}{2} + \frac{L_1\eta r}{2}\frac{1}{T+1}\sum_{t=0}^{T}\E \|\nabla F(W_t)\|_\F + \frac{2\sqrt{r}\sigma}{(T+1)\beta} + 2\sqrt{r}\sqrt{\frac{\beta}{2-\beta}}\sigma \nonumber \\
   & \quad + \sqrt{\frac{(1-\beta)^2}{(2-\beta)\beta}}2L_0\eta r + \frac{2L_1(1-\beta)\eta r }{\beta }\frac{1}{T+1}\sum_{t=0}^{T}\E\|\nabla F(W_{t})\|_F.
\end{align}

Let $0<\eta\leq \min\big(\frac{1}{2L_1r},\frac{\beta}{8L_1r(1-\beta)}\big)$, we have
\begin{align}
 \frac{L_1\eta r}{2} \leq \frac{1}{4}, \quad \frac{2L_1(1-\beta)\eta r}{\beta} \leq \frac{1}{4}.
\end{align}
Then we have
\begin{align}
    \frac{1}{T+1}\sum_{t=0}^{T}\E\|\nabla F(W_t)\|_*
   & \leq \frac{F(W_0)- F^*}{(T+1)\eta}  + \frac{r L_0\eta}{2} + \frac{2\sqrt{r}\sigma}{(T+1)\beta} + 2\sqrt{r}\sqrt{\frac{\beta}{2-\beta}}\sigma \nonumber \\
   & \quad + \sqrt{\frac{(1-\beta)^2}{(2-\beta)\beta}}2L_0\eta r + \frac{1}{2(T+1)}\sum_{t=0}^{T}\E\|\nabla F(W_{t})\|_F \nonumber \\
   & \leq  \frac{F(W_0)- F^*}{T\eta}  + \frac{r L_0\eta}{2} + \frac{2\sqrt{r}\sigma}{T\beta} + 2\sqrt{r}\sqrt{\frac{\beta}{2-\beta}}\sigma \nonumber \\
   & \quad + \sqrt{\frac{(1-\beta)^2}{(2-\beta)\beta}}2L_0\eta r + \frac{1}{2(T+1)}\sum_{t=0}^{T}\E\|\nabla F(W_{t})\|_*,
\end{align}
where the last inequality holds by $\|M_t\|_\F\leq \|M_t\|_*\leq \sqrt{r}\|M_t\|_\F$.

Then we can obtain
\begin{align}
    \frac{1}{T+1}\sum_{t=0}^{T}\E\|\nabla F(W_t)\|_*
   \leq  \frac{2(F(W_0)- F^*)}{T\eta}  + r L_0\eta + \frac{4\sqrt{r}\sigma}{T\beta} + 4\sqrt{r}\sqrt{\frac{\beta}{2-\beta}}\sigma
    + 4L_0\eta r\sqrt{\frac{(1-\beta)^2}{(2-\beta)\beta}}.
\end{align}

Let $\eta=O(\frac{1}{T^{2/3}})$ and $\beta=O(\frac{1}{T^{2/3}})$, we have
\begin{align}
    \frac{1}{T+1}\sum_{t=0}^{T}\E \|\nabla F(W_t)\|_*
    & \leq \frac{2(F(W_0)- F^*)}{T\eta}  + r L_0\eta + \frac{4\sqrt{r}\sigma}{T\beta} + 4\sqrt{r}\sqrt{\frac{\beta}{2-\beta}}\sigma + 4L_0\eta r\sqrt{\frac{(1-\beta)^2}{(2-\beta)\beta}} \nonumber \\
    & = O\big(\frac{1}{T^{1/3}}+ \frac{1}{T^{2/3}}+ \frac{1}{T^{1/3}}+ \frac{1}{T^{1/3}}+ \frac{1}{T^{1/3}}\big) =O\big(\frac{1}{T^{1/3}}\big).
\end{align}

\end{proof}

\subsection{Convergence Analysis of LiMuon Algorithm \textcolor{blue}{with Exact SVD and Option \#2 under Lipschitz Smoothness Assumption}}
In this subsection, we provide the convergence analysis of Algorithm~\ref{alg:1} with \textbf{Option} \#2 under Lipschitz smoothness assumption. For notational simplicity, let $Z_t = \nabla f(W_t;\xi_t)- \nabla f(W_{t-1};\xi_t)-(\nabla F(W_t)-\nabla F(W_{t-1}))$,
 $\delta_t =\nabla F(W_t)- \nabla f(W_t;\xi_t) $ and $\theta_t = M_{t}- \hat{M}_{t}$ for all $t\geq1$.

\begin{lemma} \label{lem:D1}
 Assume the stochastic gradient estimate $M_t$ be generated from Algorithm \ref{alg:1} under  \textbf{Option \#2}. Let $\zeta_t = M_t-\nabla F(W_t)$, $\eta_t=\eta$, $\beta_t=\beta$, $s\geq 2$, $\hat{r}\geq 2$, and $s+\hat{r}\leq r$. Under the Assumption~\ref{ass:s1}, we have
 \begin{align}
 \sum_{t=0}^{T}\E\|\zeta_t\|_\F
  \leq  \frac{\sigma}{\beta} + (T+1)\sqrt{\frac{\beta}{2-\beta}}\sigma + (T+1)\sqrt{\frac{r(1-\beta)^2}{(2-\beta)\beta}}L\eta  + \frac{\gamma(1-\beta)}{\beta} \sum_{t=0}^{T} \E\|M_t\|_\F ,
 \end{align}
 where $\gamma=\tau\left(1 + \frac{\hat{r}}{s - 1} \right)^{\frac{1}{2}}$ with $\tau\in(0,1)$.
\end{lemma}

\begin{proof}
 By the definition of $M_t$ in Algorithm \ref{alg:1} with Option \#2, we have
 \begin{align}
  M_t - \hat{M}_{t-1} = -\beta_t\hat{M}_{t-1}+ \beta_t \nabla f(W_t;\xi_t) + (1-\beta_t)\big( \nabla f(W_t;\xi_t)- \nabla f(W_{t-1};\xi_t)\big).
 \end{align}
 Let $\beta_t=\beta$, we have
 \begin{align}
 \zeta_t & = \nabla F(W_t)- M_t \nonumber \\
 & = \nabla F(W_{t-1}) - \hat{M}_{t-1} + \nabla F(W_t)-\nabla F(W_{t-1}) -(M_t-\hat{M}_{t-1}) \nonumber \\
  & = (1-\beta)(\nabla F(W_{t-1}) - \hat{M}_{t-1}) + \beta(\nabla F(W_t)- \nabla f(W_t;\xi_t)) \nonumber \\
  & \quad \quad - (1-\beta)\big( \nabla f(W_t;\xi_t)- \nabla f(W_{t-1};\xi_t)-(\nabla F(W_t)-\nabla F(W_{t-1}))\big) \nonumber \\
  & = (1-\beta)(\nabla F(W_{t-1}) - \hat{M}_{t-1}) + \beta\delta_t - (1-\beta)Z_t \nonumber \\
  & = (1-\beta) (\nabla F(W_{t-1}) -  M_{t-1} + M_{t-1}- \hat{M}_{t-1}) + \beta\delta_t - (1-\beta)Z_t \nonumber \\
  & = (1-\beta) (\nabla F(W_{t-1}) -  M_{t-1}) + (1-\beta)\theta_{t-1} + \beta\delta_t - (1-\beta)Z_t.
 \end{align}

 By expanding the recursion, then we can obtain
 \begin{align}
 \zeta_t = (1-\beta)^t\zeta_{0}  + (1-\beta)\sum_{s=1}^{t}(1-\beta)^{t-s}\theta_{s-1}+ \beta\sum_{s=1}^{t}(1-\beta)^{t-s}\delta_s -(1-\beta)\sum_{s=1}^{t}(1-\beta)^{t-s}Z_s.
 \end{align}
By using the above lemma~\ref{lem:A4}, we have
\begin{align} \label{eq:D1}
(1-\beta)\E\|\sum_{s=1}^{t}(1-\beta)^{t-s}\theta_{s-1}\|_\F & \leq (1-\beta)\sum_{s=1}^{t}(1-\beta)^{t-s}\E\|\theta_{s-1}\|_\F \nonumber \\
& \leq \gamma(1-\beta)\sum_{s=1}^{t}(1-\beta)^{t-s}\E\|M_{s-1}
\|_\F.
\end{align}

By using the proof of the above lemma~\ref{lem:B1}, we have
 \begin{align} \label{eq:D2}
 \E\|\zeta_{0}\|_\F \leq \sigma.
 \end{align}
Meanwhile, we have
 \begin{align} \label{eq:D3}
  \beta\E\|\sum_{s=1}^{t}(1-\beta)^{t-s}\delta_s\|_\F \leq  \sqrt{\frac{\beta}{2-\beta}}\sigma.
 \end{align}
We also have
 \begin{align} \label{eq:D4}
  (1-\beta)\E\|\sum_{s=1}^{t}(1-\beta)^{t-s}Z_s\|_\F  \leq  \sqrt{\frac{r(1-\beta)^2}{(2-\beta)\beta}}L\eta.
 \end{align}

 By using the above inequalities~(\ref{eq:D1}), (\ref{eq:D2}), (\ref{eq:D3}) and (\ref{eq:D4}), we have
  \begin{align}
 &\E\|\zeta_t\|_\F
 = \E\|(1-\beta)^t\zeta_{0}  + (1-\beta)\sum_{s=1}^{t}(1-\beta)^{t-s}\theta_{s-1}+ \beta\sum_{s=1}^{t}(1-\beta)^{t-s}\delta_s -(1-\beta)\sum_{s=1}^{t}(1-\beta)^{t-s}Z_s\|_\F \nonumber \\
 & \leq \E\|(1-\beta)^t\zeta_{0}\|_\F + (1-\beta)\E\|\sum_{s=1}^{t}(1-\beta)^{t-s}\theta_{s-1}\|_\F + \beta\E\|\sum_{s=1}^{t}(1-\beta)^{t-s}\delta_s\|_\F +(1-\beta)\E\|\sum_{s=1}^{t}(1-\beta)^{t-s}Z_s\|_\F \nonumber \\
 & \leq (1-\beta)^t\sigma + \sqrt{\frac{\beta}{2-\beta}}\sigma + \sqrt{\frac{r(1-\beta)^2}{(2-\beta)\beta}}L\eta  + \gamma(1-\beta)\sum_{s=1}^{t}(1-\beta)^{t-s}\E\|M_{s-1}\|_\F).
 \end{align}

 Let $M_{-1}=0$, then we have
  \begin{align}
 \sum_{t=0}^{T}\E\|\zeta_t\|_\F
 & \leq \sum_{t=0}^{T}\Big( (1-\beta)^t\sigma + \sqrt{\frac{\beta}{2-\beta}}\sigma + \sqrt{\frac{r(1-\beta)^2}{(2-\beta)\beta}}L\eta  + \gamma(1-\beta)\sum_{s=1}^{t}(1-\beta)^{t-s}\E\|M_{s-1}\|_\F \Big) \nonumber \\
 & \leq \frac{\sigma}{\beta} + (T+1)\sqrt{\frac{\beta}{2-\beta}}\sigma + (T+1)\sqrt{\frac{r(1-\beta)^2}{(2-\beta)\beta}}L\eta  + \gamma(1-\beta) \sum_{s=0}^{T}\sum_{t=s}^{T}(1-\beta)^{t-s} \E\|M_{s-1}\|_\F \nonumber \\
 & \leq \frac{\sigma}{\beta} + (T+1)\sqrt{\frac{\beta}{2-\beta}}\sigma + (T+1)\sqrt{\frac{r(1-\beta)^2}{(2-\beta)\beta}}L\eta  + \frac{\gamma(1-\beta)}{\beta} \sum_{t=0}^{T} \E\|M_t\|_\F .
 \end{align}

\end{proof}

\begin{theorem} \label{th:D1}
Under the Assumptions~\ref{ass:s1},~\ref{ass:v},~\ref{ass:l}, the sequence $\{W_t\}_{t=0}^T$ be generated
from Algorithm \ref{alg:1} with \textbf{Option} \#2. Let $\beta_t=\beta \in (0,1)$, $\eta_t=\eta$ for all $t\geq 0$, $s\geq 2$, $\hat{r}\geq 2$, $s+\hat{r}\leq r=\min(m,n)$, and $\gamma \leq \frac{\beta}{3\sqrt{r}(1-\beta)}$, we have
\begin{align}
 \frac{1}{T+1}\sum_{t=0}^{T}\E \|\nabla F(W_t)\|_* \leq \frac{2(F(W_0)- F^*)}{T\eta}  + r L\eta + 3\sqrt{r}\big(\frac{\sigma}{T\beta} + \sqrt{\frac{\beta}{2-\beta}}\sigma + \sqrt{\frac{r(1-\beta)^2}{(2-\beta)\beta}}L\eta\big),
\end{align}
 where $\gamma=\tau\left(1 + \frac{\hat{r}}{s - 1} \right)^{\frac{1}{2}}$ with $\tau\in(0,1)$.
\end{theorem}

\begin{proof}
 By using Assumption~\ref{ass:s1}, i.e., $f$ is $L$ Frobenius norm Lipschitz smooth, we have
\begin{align}
    F(W_{t+1}) \leq & F(W_t)+ \langle\nabla F(W_t), W_{t+1}-W_t \rangle+\frac{L}{2}\|W_{t+1}-W_t \|_\F^2 \nonumber \\
    \leq & F(W_t)- \eta_t\langle\nabla F(W_t), U_tV_t^\top \rangle+\frac{L}{2}\eta_t^2\|U_tV_t^\top\|_\F^2 \nonumber \\
    \leq & F(W_t) - \eta_t\langle M_t, U_tV_t^\top \rangle + \frac{r L}{2}\eta_t^2 -\eta_t\langle \nabla F(W_t)-M_t , U_tV_t^\top\rangle \nonumber \\
    \leq& F(W_t)- \eta_t\langle M_t , U_tV_t^\top \rangle + \frac{rL}{2}\eta_t^2 + \eta_t\| \nabla F(W_t)-M_t \|_\F \|U_tV_t^\top\|_\F]  \nonumber  \\
    \mathop{\leq}^{(i)} & F(W_t)- \eta_t\|M_t\|_* + \frac{r L}{2}\eta_t^2 + \eta_t\sqrt{r}\|\nabla F(W_t)-M_t\|_\F \nonumber \\
    \leq & F(W_t) -  \frac{\eta_t}{2} \|M_t\|_*-  \frac{\eta_t}{2} \|\nabla F(W_t)\|_* + \frac{r L}{2}\eta_t^2 + \frac{3}{2}\eta_t\sqrt{r}\|\nabla F(W_t)-M_t\|_\F,
\end{align}
where the above inequality (i) is due to $\langle M_t , U_tV_t^\top \rangle =\|M_t\|_*$ and $\|U_tV_t^\top\|_\F \leq \sqrt{r}$, and the last inequality holds by
$\|M_t\|_*\geq \|\nabla F(W_t)\|_* - \|\nabla F(W_t)-M_t\|_*\geq \|\nabla F(W_t)\|_* - \sqrt{r}\|\nabla F(W_t)-M_t\|_\F$.

Since $\eta_t=\eta$, we have
\begin{align}
    \E\|\nabla F(W_t)\|_* \leq & \E\big[\frac{2(F(W_t)- F(W_{t+1}))}{\eta}  -\|M_t\|_*  + r L\eta + 3\sqrt{r}\|\zeta_t\|_\F \big].
\end{align}
 Then we have
 \begin{align}
    & \frac{1}{T+1}\sum_{t=0}^{T}\E\|\nabla F(W_t)\|_* \nonumber \\
    & \leq \E[\frac{1}{T+1}\sum_{t=0}^{T}\frac{2(F(W_t)- F(W_{t+1}))}{\eta}] -\frac{1}{T+1}\sum_{t=0}^{T}\|M_t\|_* + \frac{1}{T+1}\sum_{t=0}^{T} r L\eta + \frac{1}{T+1}\sum_{t=0}^{T}3\sqrt{r}\E \|\zeta_t\|_\F \nonumber \\
    & \leq \frac{2(F(W_0)- F^*)}{(T+1)\eta}  -\frac{1}{T+1}\sum_{t=0}^{T}\|M_t\|_*+ r L\eta + \frac{1}{T+1}\sum_{t=0}^{T}3\sqrt{r}\E\|\zeta_t\|_\F \nonumber \\
    & \leq \frac{2(F(W_0)- F^*)}{(T+1)\eta}  -\frac{1}{T+1}\sum_{t=0}^{T}\|M_t\|_*+ r L\eta + \frac{3\sqrt{r}}{T+1}\Big( \frac{\sigma}{\beta} + (T+1)\sqrt{\frac{\beta}{2-\beta}}\sigma + (T+1)\sqrt{\frac{r(1-\beta)^2}{(2-\beta)\beta}}L\eta  \nonumber \\
    & \quad + \frac{\gamma(1-\beta)}{\beta} \sum_{t=0}^{T} \E\|M_t\|_\F \Big) \nonumber \\
    &\leq \frac{2(F(W_0)- F^*)}{(T+1)\eta}  -\frac{1}{T+1}\sum_{t=0}^{T}\|M_t\|_\F+ r L\eta+ 3\sqrt{r}\big(\frac{\sigma}{T\beta} + \sqrt{\frac{\beta}{2-\beta}}\sigma + \sqrt{\frac{r(1-\beta)^2}{(2-\beta)\beta}}L\eta \big) \nonumber \\
    &\quad + \frac{3\gamma\sqrt{r}(1-\beta)}{\beta} \frac{1}{T+1}\sum_{t=0}^{T} \E\|M_t\|_\F,
\end{align}
where the last inequality holds by $-\|M_t\|_*\leq -\|M_t\|_\F$.

Let $\gamma \leq \frac{\beta}{3\sqrt{r}(1-\beta)}$, we have
\begin{align}
    \frac{1}{T+1}\sum_{t=0}^{T}\E\|\nabla F(W_t)\|_* \leq \frac{2(F(W_0)- F^*)}{T\eta}  + rL\eta + 3\sqrt{r}\big(\frac{\sigma}{T\beta} + \sqrt{\frac{\beta}{2-\beta}}\sigma + \sqrt{\frac{r(1-\beta)^2}{(2-\beta)\beta}}L\eta \big) .
\end{align}

Let $\eta=O(\frac{1}{T^{2/3}})$ and $\beta=O(\frac{1}{T^{2/3}})$, we have
\begin{align}
    \frac{1}{T+1}\sum_{t=0}^{T}\E\|\nabla F(W_t)\|_*
    & \leq \frac{2(F(W_0)- F^*)}{T\eta}  + r L\eta + 3\sqrt{r}\big(\frac{\sigma}{T\beta} + \sqrt{\frac{\beta}{2-\beta}}\sigma + \sqrt{\frac{r(1-\beta)^2}{(2-\beta)\beta}}L\eta \big) \nonumber \\
    & = O\big(\frac{1}{T^{1/3}}+ \frac{1}{T^{2/3}}+ \frac{1}{T^{1/3}}+ \frac{1}{T^{1/3}}+ \frac{1}{T^{1/3}}\big) =O\big(\frac{1}{T^{1/3}}\big).
\end{align}

\end{proof}

\subsection{Convergence Analysis of LiMuon Algorithm \textcolor{blue}{with Exact SVD and Option \#2 under Generalized Smoothness Assumption}}
In this subsection, we provide the convergence analysis of Algorithm~\ref{alg:1} with \textbf{Option} \#2 under generalized smoothness assumption.

\begin{lemma} \label{lem:E1}
 Assume the stochastic gradient estimate $M_t$ be generated from Algorithm \ref{alg:1} under  \textbf{Option \#2}. Let $\zeta_t = M_t-\nabla F(W_t)$, $\eta_t=\eta$, $\beta_t=\beta$, $s\geq 2$, $\hat{r}\geq 2$, and $s+\hat{r}\leq r$. Under the Assumption~\ref{ass:s2}, we have
 \begin{align}
 \E\|\zeta_t\|_\F &\leq (1-\beta)^t\sigma + \sqrt{\frac{\beta}{2-\beta}}\sigma + L_0\eta \sqrt{\frac{r(1-\beta)^2}{(2-\beta)\beta}} + L_1\eta \sqrt{r}(1-\beta)\sum_{s=1}^{t}(1-\beta)^{(t-s)}\E\|\nabla F(W_{s-1})\|_\F \nonumber \\
 & \quad +\gamma(1-\beta)\sum_{s=1}^{t}(1-\beta)^{t-s}\E\|M_{s-1}\|_\F,
 \end{align}
 where $\gamma=\tau\left(1 + \frac{\hat{r}}{s - 1} \right)^{\frac{1}{2}}$ and $\tau\in(0,1)$.
\end{lemma}

\begin{proof}
 From the above lemma~\ref{lem:D1}, we have
 \begin{align}
 \zeta_t = (1-\beta)^t\zeta_{0}  + (1-\beta)\sum_{s=1}^{t}(1-\beta)^{t-s}\theta_{s-1}+ \beta\sum_{s=1}^{t}(1-\beta)^{t-s}\delta_s -(1-\beta)\sum_{s=1}^{t}(1-\beta)^{t-s}Z_s.
 \end{align}
 Then from both lemmas~\ref{lem:C1} and~\ref{lem:D1}, we have
 \begin{align}
 &\E\|\zeta_t\|_\F
= \E\|(1-\beta)^t\zeta_{0}  + (1-\beta)\sum_{s=1}^{t}(1-\beta)^{t-s}\theta_{s-1}+ \beta\sum_{s=1}^{t}(1-\beta)^{t-s}\delta_s -(1-\beta)\sum_{s=1}^{t}(1-\beta)^{t-s}Z_s\|_\F \nonumber \\
 & \leq \E\|(1-\beta)^t\zeta_{0}\|_\F + (1-\beta)\E\|\sum_{s=1}^{t}(1-\beta)^{t-s}\theta_{s-1}\|_\F + \beta\E\|\sum_{s=1}^{t}(1-\beta)^{t-s}\delta_s\|_\F +(1-\beta)\E\|\sum_{s=1}^{t}(1-\beta)^{t-s}Z_s\|_\F \nonumber \\
 & \leq (1-\beta)^t\sigma + \sqrt{\frac{\beta}{2-\beta}}\sigma + L_0\eta \sqrt{\frac{r(1-\beta)^2}{(2-\beta)\beta}} + L_1\eta \sqrt{r}(1-\beta)\sum_{s=1}^{t}(1-\beta)^{(t-s)}\E\|\nabla F(W_{s-1})\|_\F \nonumber \\
 & \quad +\gamma(1-\beta)\sum_{s=1}^{t}(1-\beta)^{t-s}\E\|M_{s-1}\|_\F.
 \end{align}

\end{proof}

\begin{theorem} \label{th:E1}
Under the Assumptions~\ref{ass:s2},~\ref{ass:v},~\ref{ass:l}, the sequence $\{W_t\}_{t=0}^T$ be generated
from Algorithm \ref{alg:1} with \textbf{Option} \#2. Let $\beta_t=\beta \in (0,1)$, and $\eta_t=\eta \leq \min\big(\frac{1}{4L_1r}, \frac{\beta}{12L_1r(1-\beta)} \big)$ for all $t\geq 0$, and $s\geq 2$, $\hat{r}\geq 2$, $s+\hat{r}\leq r$, and $\gamma \leq \frac{\beta}{3\sqrt{r}(1-\beta)}$, we have
\begin{align}
 \frac{1}{T+1}\sum_{t=0}^{T}\E \|\nabla F(W_t)\|_* \leq  \frac{4(F(W_0)- F^*)}{T\eta}  + 2r L_0\eta + \frac{6\sqrt{r}\sigma}{T\beta} + 6\sqrt{r}\sqrt{\frac{\beta}{2-\beta}}\sigma
    + 6L_0\eta r\sqrt{\frac{(1-\beta)^2}{(2-\beta)\beta}},
\end{align}
 where $\gamma=\tau\left(1 + \frac{\hat{r}}{s - 1} \right)^{\frac{1}{2}}$ and $\tau\in(0,1)$.
\end{theorem}

\begin{proof}
This proof basically follows the proof of Theorem~\ref{th:C1}.
 By using Assumption~\ref{ass:s2} and the above Lemma~\ref{lem:A5}, we have
\begin{align}
    F(W_{t+1}) \leq & F(W_t)+ \langle\nabla F(W_t), W_{t+1}-W_t \rangle+\frac{L_0+L_1\|\nabla F(W_t)\|_\F}{2}\|W_{t+1}-W_t \|_F^2 \nonumber \\
    \leq & F(W_t)- \eta_t\langle\nabla F(W_t), U_tV_t^\top \rangle+\frac{L_0+L_1\|\nabla F(W_t)\|_\F}{2}\eta_t^2\|U_tV_t^\top\|_\F^2 \nonumber \\
    \leq & F(W_t) - \eta_t\langle M_t, U_tV_t^\top \rangle + \frac{L_0+L_1\|\nabla F(W_t)\|_\F}{2}\eta_t^2 r -\eta_t\langle \nabla F(W_t)-M_t , U_tV_t^\top\rangle \nonumber \\
    \leq& F(W_t)- \eta_t\langle M_t , U_tV_t^\top \rangle + \frac{L_0+L_1\|\nabla F(W_t)\|_\F}{2}\eta_t^2 r + \eta_t\| \nabla F(W_t)-M_t \|_\F \|U_tV_t^\top\|_\F]  \nonumber  \\
    \mathop{\leq}^{(i)} & F(W_t)-  \eta_t\|M_t\|_* + \frac{L_0+L_1\|\nabla F(W_t)\|_\F}{2}\eta_t^2r + \eta_t\sqrt{r}\|\nabla F(W_t)-M_t\|_\F \nonumber \\
    \leq & F(W_t)-  \frac{\eta_t}{2}\|M_t\|_*-  \frac{\eta_t}{2} \|\nabla F(W_t)\|_* + \frac{L_0+L_1\|\nabla F(W_t)\|_\F}{2}\eta_t^2r + \frac{3}{2}\eta_t\sqrt{r}\|\nabla F(W_t)-M_t\|_\F,
\end{align}
where the above inequality (i) holds by $\langle M_t , U_tV_t^\top \rangle =\|M_t\|_*$ and $\|U_tV_t^\top\|_\F \leq \sqrt{r}$, and the last inequality is due to $\|M_t\|_*\geq \|\nabla F(W_t)\|_* - \|\nabla F(W_t)-M_t\|_*\geq \|\nabla F(W_t)\|_* - \sqrt{r}\|\nabla F(W_t)-M_t\|_\F$.

Let $\eta_t=\eta$, we have
\begin{align}
    \E\|\nabla F(W_t)\|_* & \leq \E\big[\frac{2(F(W_t)- F(W_{t+1}))}{\eta} -  \|M_t\|_* + r L_0\eta +L_1\eta r\|\nabla F(W_t)\|_\F + 3\sqrt{r}\|\zeta_t\|_\F \big] \nonumber \\
   & \leq \E [\frac{2(F(W_t)- F(W_{t+1}))}{\eta}]  -  \|M_t\|_* + r L_0\eta +L_1\eta r\|\nabla F(W_t)\|_\F  + 3\sqrt{r}(1-\beta)^t\sigma + 3\sqrt{r}\sqrt{\frac{\beta}{2-\beta}}\sigma \nonumber \\
   & \quad + 3L_0\eta r\sqrt{\frac{(1-\beta)^2}{(2-\beta)\beta}} + 3L_1\eta r(1-\beta)\sum_{s=1}^{t}(1-\beta)^{(t-s)}\E\|\nabla F(W_{s-1})\|_\F \nonumber \\
   & \quad +3\sqrt{r}\gamma(1-\beta)\sum_{s=1}^{t}(1-\beta)^{t-s}\E\|M_{s-1}\|_\F,
\end{align}
where the last inequality holds by the above lemma~\ref{lem:C1}.

Since $M_{-1}=0$, then we have
\begin{align}
   & \frac{1}{T+1}\sum_{t=0}^{T}\E\|\nabla F(W_t)\|_* \nonumber \\
   & \leq \frac{2(F(W_0)- F^*)}{(T+1)\eta} -\frac{1}{T+1}\sum_{t=0}^{T}\|M_t\|_* + r L_0\eta + L_1\eta r\frac{1}{T+1}\sum_{t=0}^{T}\E\|\nabla F(W_t)\|_\F + \frac{3\sqrt{r}\sigma}{T+1}\sum_{t=0}^{T}(1-\beta)^t + 3\sqrt{r}\sqrt{\frac{\beta}{2-\beta}}\sigma \nonumber \\
   & \quad + 3L_0\eta r\sqrt{\frac{(1-\beta)^2}{(2-\beta)\beta}} + 3L_1\eta r(1-\beta)\frac{1}{T+1}\sum_{t=0}^{T}\sum_{s=1}^{t}(1-\beta)^{(t-s)}\E\|\nabla F(W_{s-1})\|_\F \nonumber \\
   & \quad +3\sqrt{r}\gamma(1-\beta)\frac{1}{T+1}\sum_{t=0}^{T}\sum_{s=1}^{t}(1-\beta)^{t-s}\E\|M_{s-1}\|_\F \nonumber \\
   & \leq \frac{2(F(W_0)- F^*)}{(T+1)\eta} -\frac{1}{T+1}\sum_{t=0}^{T}\|M_t\|_* + r L_0\eta + L_1\eta r\frac{1}{T+1}\sum_{t=0}^{T}\E\|\nabla F(W_t)\|_\F + \frac{3\sqrt{r}\sigma}{(T+1)\beta} + 3\sqrt{r}\sqrt{\frac{\beta}{2-\beta}}\sigma \nonumber \\
   & \quad + 3L_0\eta r\sqrt{\frac{(1-\beta)^2}{(2-\beta)\beta}}  + \frac{3L_1\eta r(1-\beta)}{\beta}\frac{1}{T+1}\sum_{t=0}^{T}\E\|\nabla F(W_{t})\|_\F \nonumber \\
   & \quad +\frac{3\sqrt{r}\gamma(1-\beta)}{\beta}\frac{1}{T+1}\sum_{t=0}^{T}\E\|M_t\|_\F .
\end{align}

Let $0<\eta\leq \min\big(\frac{1}{4L_1r},\frac{\beta}{12L_1r(1-\beta)}\big)$ and $\gamma \leq \frac{\beta}{3\sqrt{r}(1-\beta)}$, we have
\begin{align}
 L_1\eta r \leq \frac{1}{4}, \quad \frac{3L_1\eta r(1-\beta)}{\beta} \leq \frac{1}{4}, \quad \frac{3\sqrt{r}\gamma(1-\beta)}{\beta} \leq 1.
\end{align}
Then we have
\begin{align}
    \frac{1}{T+1}\sum_{t=0}^{T}\E\|\nabla F(W_t)\|_*
   & \leq\frac{2(F(W_0)- F^*)}{(T+1)\eta}  + r L_0\eta + \frac{3\sqrt{r}\sigma}{(T+1)\beta} + 3\sqrt{r}\sqrt{\frac{\beta}{2-\beta}}\sigma \nonumber \\
   & \quad + 3L_0\eta r\sqrt{\frac{(1-\beta)^2}{(2-\beta)\beta}} + \frac{1}{2(T+1)}\sum_{t=0}^{T}\E\|\nabla F(W_t)\|_\F \nonumber \\
   & \leq  \frac{2(F(W_0)- F^*)}{(T+1)\eta}  + r L_0\eta + \frac{3\sqrt{r}\sigma}{(T+1)\beta} + 3\sqrt{r}\sqrt{\frac{\beta}{2-\beta}}\sigma \nonumber \\
   & \quad + 3L_0\eta r\sqrt{\frac{(1-\beta)^2}{(2-\beta)\beta}} + \frac{1}{2(T+1)}\sum_{t=0}^{T}\E\|\nabla F(W_t)\|_*,
\end{align}
where the first inequality holds by $\|M_t\|_\F\leq \|M_t\|_*\leq \sqrt{r}\|M_t\|_\F$.

Then we can obtain
\begin{align}
    \frac{1}{T+1}\sum_{t=0}^{T}\E\|\nabla F(W_t)\|_*
   \leq  \frac{4(F(W_0)- F^*)}{T\eta}  + 2r L_0\eta + \frac{6\sqrt{r}\sigma}{T\beta} + 6\sqrt{r}\sqrt{\frac{\beta}{2-\beta}}\sigma
    + 6L_0\eta r\sqrt{\frac{(1-\beta)^2}{(2-\beta)\beta}}.
\end{align}

Let $\eta=O(\frac{1}{T^{2/3}})$ and $\beta=O(\frac{1}{T^{2/3}})$, we have
\begin{align}
    \frac{1}{T+1}\sum_{t=0}^{T}\E \|\nabla F(W_t)\|_*
    & \leq \frac{4(F(W_0)- F^*)}{T\eta}  + 2r L_0\eta + \frac{6\sqrt{r}\sigma}{T\beta} + 6\sqrt{r}\sqrt{\frac{\beta}{2-\beta}}\sigma + 6L_0\eta r\sqrt{\frac{(1-\beta)^2}{(2-\beta)\beta}} \nonumber \\
    & = O\big(\frac{1}{T^{1/3}}+ \frac{1}{T^{2/3}}+ \frac{1}{T^{1/3}}+ \frac{1}{T^{1/3}}+ \frac{1}{T^{1/3}}\big) =O\big(\frac{1}{T^{1/3}}\big).
\end{align}

\end{proof}

\section{Convergence Analysis of Our LiMuon \textcolor{blue}{with Newton-Schulz} }
In the section, we provide the convergence analysis for our LiMuon algorithm with Newton-Schulz steps under some mild conditions.

\subsection{Convergence Analysis of LiMuon Algorithm \textcolor{blue}{with Newton-Schulz} and  Option \#1 under Lipschitz Smoothness Assumption}
In this subsection, we provide the convergence analysis of LiMuon algorithm with \textbf{Newton-Schulz} and \textbf{Option} \#1 under Lipschitz smoothness assumption.

\begin{lemma} (Lemma 6 of \citep{kim2026convergence}) \label{lem:F1}
	Under the above assumption~\ref{ass:ss1}, we have 
	\begin{align}
		F(W') \leq F(W) + \langle \nabla F(W),W'-W\rangle + \frac{L}{2}\|W'-W\|^2_{op}, \ \forall \  W,W' \in \R^{m\times n}. 
	\end{align}
\end{lemma}

\begin{lemma} \label{lem:F2}
	Assume the stochastic gradient estimate $M_t$ be generated from Algorithm \ref{alg:3} under  \textbf{Option \#1}. Let $\zeta_t = M_t-\nabla F(W_t)$, $\eta_t=\eta$ and $\beta_t=\beta$ for all $t\geq 0$, under the above Assumption~\ref{ass:ss1}, we have
	\begin{align}
		\E\|\zeta_t\|_* \leq \sqrt{r}(1-\beta)^t\sigma + \sqrt{r}\sqrt{\frac{\beta}{2-\beta}}\sigma + \sqrt{r}\sqrt{\frac{(1-\beta)^2}{(2-\beta)\beta}}L\eta (1+\varepsilon_{q}).
	\end{align}
\end{lemma}

\begin{proof}
This proof follows the above proof of Lemma~\ref{lem:B1}. By the definition of $M_t$ in Algorithm \ref{alg:3} under Option \#1, we have
\begin{align}
	M_t - M_{t-1} = -\beta_tM_{t-1}+ \beta_t \nabla f(W_t;\xi_t) + (1-\beta_t)\big( \nabla f(W_t;\xi_t)- \nabla f(W_{t-1};\xi_t)\big).
\end{align}
Let $Z_t = \nabla f(W_t;\xi_t)- \nabla f(W_{t-1};\xi_t)-(\nabla F(W_t)-\nabla F(W_{t-1}))$,
$\delta_t =\nabla F(W_t)- \nabla f(W_t;\xi_t) $, and $\beta_t=\beta$, we have
\begin{align}
	\zeta_t & = \nabla F(W_t)- M_t \nonumber \\
	& = \nabla F(W_{t-1}) - M_{t-1} + \nabla F(W_t)-\nabla F(W_{t-1}) -(M_t-M_{t-1}) \nonumber \\
	& = (1-\beta)(\nabla F(W_{t-1}) - M_{t-1}) + \beta(\nabla F(W_t)- \nabla f(W_t;\xi_t))\nonumber \\
	& \quad \quad - (1-\beta)\big( \nabla f(W_t;\xi_t)- \nabla f(W_{t-1};\xi_t)-(\nabla F(W_t)-\nabla F(W_{t-1}))\big) \nonumber \\
	& = (1-\beta)\zeta_{t-1} + \beta\delta_t + (1-\beta)Z_t
\end{align}
By expanding the recursion, then we can obtain
\begin{align}
	\zeta_t = (1-\beta)^t\zeta_{0} + \beta\sum_{s=1}^{t}(1-\beta)^{t-s}\delta_s +(1-\beta)\sum_{s=1}^{t}(1-\beta)^{t-s}Z_s.
\end{align}

Since $M_{0}=\nabla f(W_0;\xi_0)$, we have
\begin{align} \label{eq:F1}
	\E\|\zeta_{0}\|_\F = \E\|\nabla f(W_0;\xi_0)-\nabla F(W_0)\|_\F & = \sqrt{\big(\E\|\nabla f(W_0;\xi_0)-\nabla F(W_0)\|_\F\big)^2} \nonumber \\
	& \mathop{\leq}^{(i)} \sqrt{\E\|\nabla f(W_0;\xi_0)-\nabla F(W_0)\|_\F^2} \mathop{\leq}^{(ii)} \sqrt{\sigma^2} =\sigma,
\end{align}
where the above inequality (i) is due to Jensen inequality, and the above inequality (ii) holds by Assumption~\ref{ass:v}.

Meanwhile, we have
\begin{align} \label{eq:F2}
	\beta\E\|\sum_{s=1}^{t}(1-\beta)^{t-s}\delta_s\|_\F & = \sqrt{\beta^2\big(\E\big\|\sum_{s=1}^{t}(1-\beta)^{t-s}\delta_s \big\|_\F \big)^2}  \nonumber \\
	& \leq \sqrt{\beta^2\E\big\|\sum_{s=1}^{t}(1-\beta)^{t-s}\delta_s \big\|^2_\F} \nonumber \\
	& \mathop{=}^{(i)} \sqrt{\beta^2\sum_{s=1}^{t}(1-\beta)^{2(t-s)}\E\big\|\nabla F(W_s)- \nabla f(W_s;\xi_s)\big\|^2_\F} \nonumber \\
	& \leq\sqrt{ \beta^2\sum_{s=1}^{t}(1-\beta)^{2(t-s)} \sigma^2} \leq \sqrt{\frac{\beta^2}{1-(1-\beta)^2}\sigma^2} = \sqrt{\frac{\beta}{2-\beta}}\sigma,
\end{align}
where the above equality (i) holds by the above Lemma~\ref{lem:A1} on the fact that  $\{\xi_1,\cdots,\xi_t\}$ are independent random variables and $\E_{\xi_s}[\nabla F(W_s)- \nabla f(W_s;\xi_s)]=0$ for all $1\leq s\leq t$.

We also can obtain
\begin{align} \label{eq:F3}
	(1-\beta)\E\|\sum_{s=1}^{t}(1-\beta)^{t-s}Z_s\|_\F & = \sqrt{(1-\beta)^2\big(\E\|\sum_{s=1}^{t}(1-\beta)^{t-s}Z_s\|_\F \big)^2} \nonumber \\
	& \leq  \sqrt{(1-\beta)^2\E\|\sum_{s=1}^{t}(1-\beta)^{t-s}Z_s\|^2_\F} \nonumber \\
	& \mathop{=}^{(i)} \sqrt{(1-\beta)^2\sum_{s=1}^{t}(1-\beta)^{2(t-s)}\E\|\nabla f(W_s;\xi_s)- \nabla f(W_{s-1};\xi_s)-(\nabla F(W_s)-\nabla F(W_{s-1}))\|^2_\F} \nonumber \\
	& \mathop{\leq}^{(ii)} \sqrt{(1-\beta)^2\sum_{s=1}^{t}(1-\beta)^{2(t-s)}\E\|\nabla f(W_s;\xi_s)- \nabla f(W_{s-1};\xi_s)\|^2_\F } \nonumber \\
	& \leq \sqrt{(1-\beta)^2\sum_{s=1}^{t}(1-\beta)^{2(t-s)}\E\|\nabla f(W_s;\xi_s)- \nabla f(W_{s-1};\xi_s)\|^2_* } \nonumber \\
	& \mathop{\leq}^{(iii)} \sqrt{(1-\beta)^2\sum_{s=1}^{t}(1-\beta)^{2(t-s)}L^2\E\|W_s- W_{s-1}\|^2_{op}} \nonumber \\
	& = \sqrt{(1-\beta)^2\sum_{s=1}^{t}(1-\beta)^{2(t-s)}L^2\E\|\eta O_{t-1}\|^2_{op}} \nonumber \\
	& \leq \sqrt{(1-\beta)^2\sum_{s=1}^{t}(1-\beta)^{2(t-s)}L^2\eta^2(1+\varepsilon_{q})^2} \nonumber \\
	& \leq \sqrt{(1-\beta)^2\frac{1}{1-(1-\beta)^2}L^2\eta^2(1+\varepsilon_{q})^2} = \sqrt{\frac{(1-\beta)^2}{(2-\beta)\beta}}L\eta (1+\varepsilon_{q}),
\end{align}
where the above equality (i) holds by the above Lemma~\ref{lem:A1} on the fact that$\{\xi_1,\cdots,\xi_t\}$ are independent random variables and $\E_{\xi_s}[\nabla f(W_s;\xi_s)- \nabla f(W_{s-1};\xi_s)-(\nabla F(W_s)-\nabla F(W_{s-1}))]=0$ for all $1\leq s\leq t$; the above inequality (ii) holds by the above lemma~\ref{lem:A2}, and the above inequality (iii) is due to Assumption~\ref{ass:ss1}, and the second last inequality is due to $\|O_{t-1}\|_{op} \leq 1+\varepsilon_{q}$.

By using the above inequalities~(\ref{eq:F1}), (\ref{eq:F2}) and (\ref{eq:F3}), we have
\begin{align}
	\E\|\zeta_t\|_\F
	& = \E\|(1-\beta)^t\zeta_{0} + \beta\sum_{s=1}^{t}(1-\beta)^{t-s}\delta_s -(1-\beta)\sum_{s=1}^{t}(1-\beta)^{t-s}Z_s\|_\F \nonumber \\
	& \leq (1-\beta)^t\E\|\zeta_{0}\|_\F + \beta\E\|\sum_{s=1}^{t}(1-\beta)^{t-s}\delta_s\|_\F +(1-\beta)\E\|\sum_{s=1}^{t}(1-\beta)^{t-s}Z_s\|_\F \nonumber \\
	& \leq (1-\beta)^t\sigma + \sqrt{\frac{\beta}{2-\beta}}\sigma + \sqrt{\frac{(1-\beta)^2}{(2-\beta)\beta}}L\eta (1+\varepsilon_{q}).
\end{align}

Since $\|\zeta_t\|_* \leq \sqrt{r}\|\zeta_t\|_\F$, we have 
\begin{align}
	\E\|\zeta_t\|_* \leq \sqrt{r}\E\|\zeta_t\|_\F
	 \leq \sqrt{r}(1-\beta)^t\sigma + \sqrt{r}\sqrt{\frac{\beta}{2-\beta}}\sigma + \sqrt{r}\sqrt{\frac{(1-\beta)^2}{(2-\beta)\beta}}L\eta (1+\varepsilon_{q}).
\end{align}

\end{proof}

\begin{theorem} \label{th:F1}
	Under the above Assumptions~\ref{ass:ss1},~\ref{ass:v},~\ref{ass:l}, the sequence $\{W_t\}_{t=0}^T$ be generated
	from Algorithm \ref{alg:3} with \textbf{Option} \#1. Let $\beta_t=\beta \in (0,1)$ and $\eta_t=\eta$ for all $t\geq 0$, we have
	\begin{align}
		\frac{1}{T+1}\sum_{t=0}^{T}\E \|\nabla F(W_t)\|_* \leq \frac{F(W_0)- F^*}{T\eta(1-\varepsilon_{q})} + \frac{ L\eta(1+\varepsilon_{q})^2}{2(1-\varepsilon_{q})}  +\frac{2}{1-\varepsilon_{q}}\Big( \frac{\sqrt{r}\sigma}{T\beta} + \sqrt{r}\sqrt{\frac{\beta}{2-\beta}}\sigma + \sqrt{r}\sqrt{\frac{(1-\beta)^2}{(2-\beta)\beta}}L\eta (1+\varepsilon_{q})\Big),
	\end{align}
	where $\varepsilon_{q}\in (0,1)$ is a polar approximation error.
\end{theorem}

\begin{proof}

	By using the above Lemma~\ref{lem:F1},  we have
	\begin{align}
		F(W_{t+1}) \leq & F(W_t)+ \langle\nabla F(W_t), W_{t+1}-W_t \rangle+\frac{L}{2}\|W_{t+1}-W_t \|_{op}^2 \nonumber \\
		= & F(W_t)- \eta_t\langle\nabla F(W_t), O_t \rangle+\frac{L}{2}\eta_t^2\|O_t\|_{op}^2 \nonumber \\
		= & F(W_t) - \eta_t\langle M_t, O_t \rangle  -\eta_t\langle \nabla F(W_t)-M_t , O_t \rangle + \frac{L}{2}\eta_t^2 \|O_t\|_{op}^2 \nonumber \\
		\leq & F(W_t) - \eta_t\langle M_t, O_t \rangle  +\eta_t\| \nabla F(W_t)-M_t\|_* \|O_t\|_{op} + \frac{L}{2}\eta_t^2 \|O_t\|_{op}^2 \nonumber \\
		= & F(W_t) - \eta_t\langle M_t, P_t \rangle +\eta_t\langle M_t, P_t-O_t \rangle +\eta_t\| \nabla F(W_t)-M_t\|_* \|O_t\|_{op} + \frac{L}{2}\eta_t^2 \|O_t\|_{op}^2 \nonumber \\
		= & F(W_t) - \eta_t\| M_t\|_* +\eta_t\langle M_t, P_t-O_t \rangle +\eta_t\| \nabla F(W_t)-M_t\|_* \|O_t\|_{op} + \frac{L}{2}\eta_t^2 \|O_t\|_{op}^2 \nonumber \\
		\leq & F(W_t) - \eta_t\| M_t\|_* +\eta_t\ M_t\|_* \|P_t-O_t \|_{op} +\eta_t\| \nabla F(W_t)-M_t\|_* \|O_t\|_{op} + \frac{L}{2}\eta_t^2 \|O_t\|_{op}^2,
	\end{align}
	where the second inequality is due to $\langle M_t , U_tV_t^\top \rangle =\|M_t\|_*$.
	
	Since $P_t=U_tV_t^\top$, $\varepsilon_{q,t}=\|O_t-P_t\|_{op}$ with $\varepsilon_{q}=\sup_{t} \varepsilon_{q,t}$, we have 
	\begin{align}
		\|P_t\|_{op}=1, \quad \|O_t\|_{op} \leq 1+\varepsilon_{q,t}\leq 1+\varepsilon_{q},
	\end{align}
	where $\varepsilon_{q}\in (0,1)$.
	Thus, we can obtain 
	\begin{align}
		F(W_{t+1}) 
		 & \leq F(W_t) - \eta_t\| M_t\|_* +\eta_t\ M_t\|_* \|P_t-O_t \|_{op} +\eta_t\| \nabla F(W_t)-M_t\|_* \|O_t\|_{op} + \frac{L}{2}\eta_t^2 \|O_t\|_{op}^2 \nonumber \\
		& \leq F(W_t) - \eta_t(1-\varepsilon_{q})\| M_t\|_* + \eta_t(1+\varepsilon_{q})\|\nabla F(W_t)-M_t\|_* + \frac{L}{2}\eta_t^2 (1+\varepsilon_{q})^2 \nonumber \\
		& \leq F(W_t) - \eta_t(1-\varepsilon_{q})\| \nabla F(W_t)\|_* + 2\eta_t\|\nabla F(W_t)-M_t\|_* + \frac{L}{2}\eta_t^2 (1+\varepsilon_{q})^2 , 
	\end{align}
	where the last inequality holds by
	$\|M_t\|_*\geq \|\nabla F(W_t)\|_* - \|\nabla F(W_t)-M_t\|_*$.
	
	Let $\eta_t=\eta$ and $\zeta_t=\nabla F(W_t)-M_t$, then we have
	\begin{align}
		\E\|\nabla F(W_t)\|_* \leq & \E\big[\frac{F(W_t)- F(W_{t+1})}{\eta(1-\varepsilon_{q})}  + \frac{2}{1-\varepsilon_{q}}\|\zeta_t\|_*  + \frac{ L\eta(1+\varepsilon_{q})^2}{2(1-\varepsilon_{q})}\big].
	\end{align}
	
	Thus we can obtain
	\begin{align}
		&\frac{1}{T+1}\sum_{t=0}^{T}\E \|\nabla F(W_t)\|_* \nonumber \\
		& \leq  \E\big[ \frac{1}{T+1}\sum_{t=0}^{T}\frac{F(W_t)- F(W_{t+1})}{\eta(1-\varepsilon_{q})}  + \frac{1}{T+1}\sum_{t=0}^{T}\frac{2}{1-\varepsilon_{q}}\|\zeta_t\|_*  + \frac{1}{T+1}\sum_{t=0}^{T} \frac{r L\eta(1+\varepsilon_{q})^2}{2(1-\varepsilon_{q})}\big] \nonumber \\
		& \mathop{\leq}^{(i)} \frac{F(W_0)- F^*}{(T+1)\eta(1-\varepsilon_{q})} + \frac{ L\eta(1+\varepsilon_{q})^2}{2(1-\varepsilon_{q})}  + \frac{1}{T+1}\sum_{t=0}^{T}\frac{2}{1-\varepsilon_{q}}\|\zeta_t\|_* \nonumber \\
		& \mathop{\leq}^{(ii)} \frac{F(W_0)- F^*}{(T+1)\eta(1-\varepsilon_{q})} + \frac{ L\eta(1+\varepsilon_{q})^2}{2(1-\varepsilon_{q})}  +   \frac{1}{T+1}\sum_{t=0}^{T}\frac{2}{1-\varepsilon_{q}}\big(\sqrt{r}(1-\beta)^t\sigma + \sqrt{r}\sqrt{\frac{\beta}{2-\beta}}\sigma + \sqrt{r}\sqrt{\frac{(1-\beta)^2}{(2-\beta)\beta}}L\eta (1+\varepsilon_{q})\big) \nonumber \\
		& \leq \frac{F(W_0)- F^*}{T\eta(1-\varepsilon_{q})} + \frac{ L\eta(1+\varepsilon_{q})^2}{2(1-\varepsilon_{q})}  +\frac{2}{1-\varepsilon_{q}}\Big( \frac{\sqrt{r}\sigma}{T\beta} + \sqrt{r}\sqrt{\frac{\beta}{2-\beta}}\sigma + \sqrt{r}\sqrt{\frac{(1-\beta)^2}{(2-\beta)\beta}}L\eta (1+\varepsilon_{q})\Big),
	\end{align}
	where the above inequality (i) is due to Assumption~\ref{ass:l}, and the above inequality (ii) holds by the above lemma~\ref{lem:B1}.
	
	Let $\eta=O(\frac{1}{T^{2/3}})$,  $\beta=O(\frac{1}{T^{2/3}})$, $L=O(1)$ and $\sigma=O(1)$, we have
	\begin{align}
		& \frac{1}{T+1}\sum_{t=0}^{T} \E \|\nabla F(W_t)\|_* \nonumber \\
		& \leq \frac{F(W_0)- F^*}{T\eta(1-\varepsilon_{q})} + \frac{ L\eta(1+\varepsilon_{q})^2}{2(1-\varepsilon_{q})}  +\frac{2}{1-\varepsilon_{q}}\Big( \frac{\sqrt{r}\sigma}{T\beta} + \sqrt{r}\sqrt{\frac{\beta}{2-\beta}}\sigma + \sqrt{r}\sqrt{\frac{(1-\beta)^2}{(2-\beta)\beta}}L\eta (1+\varepsilon_{q})\Big) \nonumber \\
		& = O\big(\frac{1}{(1-\varepsilon_{q})T^{1/3}}+ \frac{1}{(1-\varepsilon_{q})T^{2/3}}+ \frac{1}{(1-\varepsilon_{q})T^{1/3}}+ \frac{1}{(1-\varepsilon_{q})T^{1/3}}+ \frac{1}{(1-\varepsilon_{q})T^{1/3}}\big) =O\big(\frac{1}{(1-\varepsilon_{q})T^{1/3}}\big)
	\end{align}

\end{proof}

\subsection{Convergence Analysis of LiMuon Algorithm \textcolor{blue}{ with Newton-Schulz} and Option \#1 under Generalized Smoothness Assumption}
In this subsection, we provide the convergence analysis of LiMuon algorithm with \textbf{Newton-Schulz} and \textbf{Option} \#1 under \textbf{generalized smoothness} assumption. 

\begin{lemma}\label{lem:G0}
	Under the Assumption~\ref{ass:ss2}, we have for any $W,W'\in \R^{m\times n}$
	\begin{align*}
		F(W') \leq F(W)+\langle \nabla F(W),W'-W\rangle+\frac{L_{0}+L_{1}\|\nabla F(W)\|_*}{2}\|W'-W\|_{op}^{2}.
	\end{align*}
\end{lemma}

\begin{proof}
	By using Assumption~\ref{ass:ss2}, we have
	\begin{align} \label{eq:G0}
		\|\nabla F(W)-\nabla F(W')\|_\F
		& \leq \sqrt{(L_0^2+L_1^2\|\nabla F(W)\|_*^2)\|W-W'\|_{op}^2} \nonumber \\
		& \leq (L_0+L_1\|\nabla F(W)\|_*)\|W-W'\|_{op}.
	\end{align}
	Since $F(W)$ is differentiable, by applying the fundamental theorem
	of calculus to $G(t)= F(W+t(W'-W))$, we have
	\begin{align}
		F(W') & =F(W)+\int_{0}^{1}\langle \nabla F(W+t(W'-W)),W'-W\rangle\mathrm{d}t  \nonumber \\
		& =F(W)+\langle \nabla F(W),W'-W\rangle+\int_{0}^{1}\langle \nabla F(W+t(W'-W))- \nabla F(W),W'-W\rangle\mathrm{d}t  \nonumber \\
		& \overset{(i)}{\leq}F(W)+\langle \nabla F(W),W'-W\rangle+\int_{0}^{1}\| \nabla F(W+t(W'-W))- \nabla F(W)\|_*\|W'-W\|_{op} \mathrm{d}t \nonumber \\
		& \overset{(ii)}{\leq}F(W)+\langle \nabla F(W),W'-W\rangle+\int_{0}^{1}(L_{0}+L_{1}\| \nabla F(W)\|_*)\|W'-W\|_{op}^{2} t\mathrm{d}t  \nonumber \\
		& =F(W)+\langle \nabla F(W),W'-W\rangle+\frac{L_{0}+L_{1}\|\nabla F(W)\|_*}{2}\|W'-W\|_{op}^{2},
	\end{align}
	where the above inequality $(i)$ is by Cauchy-Schwarz inequality and the inequality $(ii)$ is due to
	the above inequality~(\ref{eq:G0}).
\end{proof}

\begin{lemma} \label{lem:G1}
	Assume the stochastic gradient estimate $M_t$ be generated from Algorithm \ref{alg:3} under  \textbf{Option \#1}. Let $\zeta_t = M_t-\nabla F(W_t)$, $\eta_t=\eta$ and $\beta_t=\beta$ for all $t\geq 0$, under the assumption~\ref{ass:ss2}, we have
	\begin{align}
		\E\|\zeta_t\|_*
		& \leq \sqrt{r}(1-\beta)^t\sigma + \sqrt{r}\sqrt{\frac{\beta}{2-\beta}}\sigma + \sqrt{r}\sqrt{\frac{(1-\beta)^2}{(2-\beta)\beta}}L_0\eta (1+\varepsilon_{q}) \nonumber \\
		&\quad + L_1(1-\beta)\eta \sqrt{r}  (1+\varepsilon_{q})\sum_{s=1}^{t}(1-\beta)^{(t-s)}\E\|\nabla F(W_{s-1})\|_*.
	\end{align}
\end{lemma}

\begin{proof}
	By using Assumption~\ref{ass:ss2}, we have
	\begin{align} \label{eq:G1}
		&(1-\beta)\E\|\sum_{s=1}^{t}(1-\beta)^{t-s}Z_s\|_\F  \nonumber \\
		& = \sqrt{(1-\beta)^2\big(\E\|\sum_{s=1}^{t}(1-\beta)^{t-s}Z_s\|_\F \big)^2} \nonumber \\
		& \mathop{\leq}^{(i)}  \sqrt{(1-\beta)^2\E\|\sum_{s=1}^{t}(1-\beta)^{t-s}Z_s\|^2_\F} \nonumber \\
		& \mathop{=}^{(ii)} \sqrt{(1-\beta)^2\sum_{s=1}^{t}(1-\beta)^{2(t-s)}\E\|\nabla f(W_s;\xi_s)- \nabla f(W_{s-1};\xi_s)-(\nabla F(W_s)-\nabla F(W_{s-1}))\|^2_\F} \nonumber \\
		& \leq \sqrt{(1-\beta)^2\sum_{s=1}^{t}(1-\beta)^{2(t-s)}\E\|\nabla f(W_s;\xi_s)- \nabla f(W_{s-1};\xi_s)\|^2_\F } \nonumber \\
		& \leq \sqrt{(1-\beta)^2\sum_{s=1}^{t}(1-\beta)^{2(t-s)}\E\|\nabla f(W_s;\xi_s)- \nabla f(W_{s-1};\xi_s)\|^2_* } \nonumber \\
		&  \mathop{\leq}^{(iii)} \sqrt{(1-\beta)^2\sum_{s=1}^{t}(1-\beta)^{2(t-s)}(L_0^2+L_1^2(\E\|\nabla F(W_{s-1})\|_*)^2)\|W_s- W_{s-1}\|^2_{op}} \nonumber \\
		& =\sqrt{(1-\beta)^2\sum_{s=1}^{t}(1-\beta)^{2(t-s)}(L_0^2+L_1^2(\E\|\nabla F(W_{s-1})\|_*)^2)\|\eta O_{t-1}\|^2_{op}} \nonumber \\
		& \leq \sqrt{(1-\beta)^2\sum_{s=1}^{t}(1-\beta)^{2(t-s)}(L_0^2+L_1^2(\E\|\nabla F(W_{s-1})\|_*)^2)\eta^2(1+\varepsilon_{q})^2},
	\end{align}
	where the above inequality (i) is due to Jensen inequality, and the above equality (ii) holds by the above Lemma~\ref{lem:A1} on the fact that$\{\xi_1,\cdots,\xi_t\}$ are independent random variables and $\E_{\xi_s}[\nabla f(W_s;\xi_s)- \nabla f(W_{s-1};\xi_s)-(\nabla F(W_s)-\nabla F(W_{s-1}))]=0$ for all $1\leq s\leq t$, and the above inequality (iii) is due to Assumption~\ref{ass:ss2}.
	
	By using the inequality $\sqrt{a+b}\leq \sqrt{a}+\sqrt{b}$ for all $a,b\geq0$, we have
	\begin{align} \label{eq:G2}
		&\sqrt{(1-\beta)^2\sum_{s=1}^{t}(1-\beta)^{2(t-s)}(L_0^2+L_1^2(\E\|\nabla F(W_{s-1})\|_*)^2)\eta^2(1+\varepsilon_{q})^2} \nonumber \\
		& \leq \sqrt{(1-\beta)^2\sum_{s=1}^{t}(1-\beta)^{2(t-s)}L_0^2\eta^2(1+\varepsilon_{q})^2} + \sqrt{(1-\beta)^2\sum_{s=1}^{t}(1-\beta)^{2(t-s)}L_1^2(\E\|\nabla F(W_{s-1})\|_*)^2\eta^2(1+\varepsilon_{q})^2} \nonumber \\
		& \leq \sqrt{(1-\beta)^2\frac{1}{1-(1-\beta)^2}L_0^2\eta^2(1+\varepsilon_{q})^2} + \sqrt{(1-\beta)^2\sum_{s=1}^{t}(1-\beta)^{2(t-s)}L_1^2(\E\|\nabla F(W_{s-1})\|_*)^2\eta^2(1+\varepsilon_{q})^2} \nonumber \\
		& =  \sqrt{\frac{(1-\beta)^2}{(2-\beta)\beta}}L_0\eta (1+\varepsilon_{q}) + \sqrt{(1-\beta)^2\sum_{s=1}^{t}(1-\beta)^{2(t-s)}L_1^2(\E\|\nabla F(W_{s-1})\|_*)^2\eta^2r^2} \nonumber \\
		& \leq \sqrt{\frac{(1-\beta)^2}{(2-\beta)\beta}}L_0\eta (1+\varepsilon_{q}) + L_1(1-\beta)\eta (1+\varepsilon_{q})\sum_{s=1}^{t}(1-\beta)^{(t-s)}\E\|\nabla F(W_{s-1})\|_*.
	\end{align}
	
	By using the above inequalities~(\ref{eq:G1}) and~(\ref{eq:G2}), we have
	\begin{align} \label{eq:G3}
		(1-\beta)\E\|\sum_{s=1}^{t}(1-\beta)^{t-s}Z_s\|_\F & \leq \sqrt{(1-\beta)^2\sum_{s=1}^{t}(1-\beta)^{2(t-s)}(L_0+L_1\E\|\nabla F(W_{s})\|_*)^2\eta^2(1+\varepsilon_{q})^2} \nonumber \\
		& \leq \sqrt{\frac{(1-\beta)^2}{(2-\beta)\beta}}L_0\eta (1+\varepsilon_{q}) + L_1(1-\beta)\eta (1+\varepsilon_{q})\sum_{s=1}^{t}(1-\beta)^{(t-s)}\E\|\nabla F(W_{s-1})\|_*.
	\end{align}
	
	From the above lemma~\ref{lem:B1}, we have
	\begin{align} \label{eq:G4}
		\E\|\zeta_{0}\|_\F \leq \sigma.
	\end{align}
	We also have
	\begin{align} \label{eq:G5}
		\beta\E\|\sum_{s=1}^{t}(1-\beta)^{t-s}\delta_s\|_\F  \leq  \sqrt{\frac{\beta}{2-\beta}}\sigma.
	\end{align}
	
	By using the above inequalities~(\ref{eq:G3}), (\ref{eq:G4}) and~(\ref{eq:G5}), we can obtain
	\begin{align}
		\E\|\zeta_t\|_\F
		& = \E\|(1-\beta)^t\zeta_{0} + \beta\sum_{s=1}^{t}(1-\beta)^{t-s}\delta_s -(1-\beta)\sum_{s=1}^{t}(1-\beta)^{t-s}Z_s\|_\F \nonumber \\
		& \leq (1-\beta)^t\E\|\zeta_{0}\|_\F + \beta\E\|\sum_{s=1}^{t}(1-\beta)^{t-s}\delta_s\|_\F +(1-\beta)\E\|\sum_{s=1}^{t}(1-\beta)^{t-s}Z_s\|_\F \nonumber \\
		& \leq (1-\beta)^t\sigma + \sqrt{\frac{\beta}{2-\beta}}\sigma + \sqrt{\frac{(1-\beta)^2}{(2-\beta)\beta}}L_0\eta (1+\varepsilon_{q}) + L_1(1-\beta)\eta (1+\varepsilon_{q})\sum_{s=1}^{t}(1-\beta)^{(t-s)}\E\|\nabla F(W_{s-1})\|_*.
	\end{align}

Since $\|\zeta_t\|_* \leq \sqrt{r}\|\zeta_t\|_\F$, we have 
  \begin{align}
		\E\|\zeta_t\|_*
		& \leq \sqrt{r}(1-\beta)^t\sigma + \sqrt{r}\sqrt{\frac{\beta}{2-\beta}}\sigma + \sqrt{r}\sqrt{\frac{(1-\beta)^2}{(2-\beta)\beta}}L_0\eta (1+\varepsilon_{q}) \nonumber \\
		&\quad + L_1(1-\beta)\eta \sqrt{r}  (1+\varepsilon_{q})\sum_{s=1}^{t}(1-\beta)^{(t-s)}\E\|\nabla F(W_{s-1})\|_*.
	\end{align}
	
\end{proof}

\begin{theorem} \label{th:G1}
	Under the Assumptions~\ref{ass:ss2},~\ref{ass:v},~\ref{ass:l}, the sequence $\{W_t\}_{t=0}^T$ be generated
	from Algorithm \ref{alg:3} with \textbf{Option} \#1. Let $\beta_t=\beta \in (0,1)$ and $\eta_t=\eta \leq \min\big(\frac{1-\varepsilon_{q}}{2L_1(1+\varepsilon_{q})^2},\frac{\beta(1-\varepsilon_{q})}{8L_1\sqrt{r}(1+\varepsilon_{q})(1-\beta)}\big)$ for all $t\geq 0$, we have
	\begin{align}
		\frac{1}{T+1}\sum_{t=0}^{T}\E \|\nabla F(W_t)\|_* & \leq \frac{2((W_0)- F^*)}{(T+1)\eta(1-\varepsilon_{q})}   + \frac{L_0\eta(1+\varepsilon_{q})^2}{1-\varepsilon_{q}} \nonumber \\
		&\quad + \frac{4}{1-\varepsilon_{q}}\Big(\frac{\sqrt{r}\sigma}{T\beta} + \sqrt{r}\sqrt{\frac{\beta}{2-\beta}}\sigma + \sqrt{r}\sqrt{\frac{(1-\beta)^2}{(2-\beta)\beta}}L_0\eta (1+\varepsilon_{q})\Big),
	\end{align}
	where $\varepsilon_{q}\in (0,1)$ is a polar approximation error.
\end{theorem}

\begin{proof}
	By using the above Lemma~\ref{lem:G0}, we have
	\begin{align}
		F(W_{t+1}) \leq & F(W_t)+ \langle\nabla F(W_t), W_{t+1}-W_t \rangle+\frac{L_0+L_1\|\nabla F(W_t)\|_*}{2}\|W_{t+1}-W_t \|_{op}^2 \nonumber \\
		\leq & F(W_t)- \eta_t\langle\nabla F(W_t), O_t \rangle+\frac{L_0+L_1\|\nabla F(W_t)\|_*}{2}\eta_t^2\|O_t\|_{op}^2 \nonumber \\
		\leq & F(W_t) - \eta_t\langle M_t, O_t \rangle  -\eta_t\langle \nabla F(W_t)-M_t , O_t\rangle + \frac{L_0+L_1\|\nabla F(W_t)\|_*}{2}\eta_t^2 \|O_t\|_{op} \nonumber \\
		= & F(W_t) - \eta_t\langle M_t, P_t \rangle + \eta_t\langle M_t, P_t-O_t \rangle -\eta_t\langle \nabla F(W_t)-M_t , O_t\rangle + \frac{L_0+L_1\|\nabla F(W_t)\|_*}{2}\eta_t^2 \|O_t\|^2_{op} \nonumber \\
		\leq & F(W_t)-  \eta_t \|M_t\|_* + \eta_t\|M_t\|_* \|P_t-O_t \|_{op} +\eta_t\|\nabla F(W_t)-M_t\|_* \|O_t\|_{op} \nonumber \\
		&\quad + \frac{L_0+L_1\|\nabla F(W_t)\|_*}{2}\eta_t^2 \|O_t\|^2_{op},
	\end{align}
	where the last inequality holds by $\langle M_t , P_t \rangle=\langle M_t , U_tV_t^\top \rangle =\|M_t\|_*$.
	
	Since $P_t=U_tV_t^\top$, $\varepsilon_{q,t}=\|O_t-P_t\|_{op}$ with $\varepsilon_{q}=\sup_{t} \varepsilon_{q,t}$, we have 
	\begin{align}
		\|P_t\|_{op}=1, \quad \|O_t\|_{op} \leq 1+\varepsilon_{q,t}\leq 1+\varepsilon_{q},
	\end{align}
	where $\varepsilon_{q}\in (0,1)$.
	Thus, we can obtain 
	\begin{align}
		F(W_{t+1}) \leq  & F(W_t)-  \eta_t \|M_t\|_* + \eta_t\|M_t\|_* \|P_t-O_t \|_{op} +\eta_t\|\nabla F(W_t)-M_t\|_* \|O_t\|_{op} \nonumber \\
		&\quad + \frac{L_0+L_1\|\nabla F(W_t)\|_*}{2}\eta_t^2 \|O_t\|^2_{op} \nonumber \\
		\leq & F(W_t)-  \eta_t \|M_t\|_* + \eta_t\|M_t\|_* \varepsilon_{q} +\eta_t\|\nabla F(W_t)-M_t\|_* (1+\varepsilon_{q}) \nonumber \\
		&\quad + \frac{L_0+L_1\|\nabla F(W_t)\|_*}{2}\eta_t^2 (1+\varepsilon_{q})^2 \nonumber \\
		\leq & F(W_t)- \eta_t (1-\varepsilon_{q})\|\nabla F(W_t)\|_* +2\eta_t\|\nabla F(W_t)-M_t\|_*   + \frac{L_0+L_1\|\nabla F(W_t)\|_*}{2}\eta_t^2 (1+\varepsilon_{q})^2, 
	\end{align}
 where the last inequality holds by $\|M_t\|_*\geq \|\nabla F(W_t)\|_* - \|\nabla F(W_t)-M_t\|_*$.
	
	Let $\eta_t=\eta$ and $\zeta_t=\nabla F(W_t)-M_t$, we have
	\begin{align}
		\E\|\nabla F(W_t)\|_* & \leq \E\big[\frac{F(W_t)- F(W_{t+1})}{\eta(1-\varepsilon_{q})}  + \frac{L_0\eta(1+\varepsilon_{q})^2}{2(1-\varepsilon_{q})} +\frac{L_1\eta (1+\varepsilon_{q})^2}{2(1-\varepsilon_{q})}\|\nabla F(W_t)\|_* + \frac{2}{1-\varepsilon_{q}}\|\zeta_t\|_* \big] \nonumber \\
		& \leq \E [\frac{F(W_t)- F(W_{t+1})}{\eta(1-\varepsilon_{q})} ] + \frac{L_0\eta(1+\varepsilon_{q})^2}{2(1-\varepsilon_{q})} +\frac{L_1\eta (1+\varepsilon_{q})^2}{2(1-\varepsilon_{q})}\E \|\nabla F(W_t)\|_* \nonumber \\
		& \quad + \frac{2}{1-\varepsilon_{q}}\Big(\sqrt{r}(1-\beta)^t\sigma + \sqrt{r}\sqrt{\frac{\beta}{2-\beta}}\sigma + \sqrt{r}\sqrt{\frac{(1-\beta)^2}{(2-\beta)\beta}}L_0\eta (1+\varepsilon_{q}) \nonumber \\
		&\qquad + L_1(1-\beta)\eta \sqrt{r}  (1+\varepsilon_{q})\sum_{s=1}^{t}(1-\beta)^{(t-s)}\E\|\nabla F(W_{s-1})\|_*\Big),
	\end{align}
	where the last inequality holds by the above lemma~\ref{lem:G1}.
	
	Let $\nabla f(W_{-1})=0$, then we have
	\begin{align} \label{eq:G7}
		\frac{1}{T+1}\sum_{t=0}^{T}\E\|\nabla F(W_t)\|_*
		& \leq \frac{F(W_0)- F^*}{(T+1)\eta(1-\varepsilon_{q})}   + \frac{L_0\eta(1+\varepsilon_{q})^2}{2(1-\varepsilon_{q})} +\frac{L_1\eta (1+\varepsilon_{q})^2}{2(1-\varepsilon_{q})}\frac{1}{T+1}\sum_{t=0}^{T}\E \|\nabla F(W_t)\|_*  \nonumber \\
		&  \quad + \frac{2}{1-\varepsilon_{q}}\frac{1}{T+1}\sum_{t=0}^{T}\Big(\sqrt{r}(1-\beta)^t\sigma + \sqrt{r}\sqrt{\frac{\beta}{2-\beta}}\sigma + \sqrt{r}\sqrt{\frac{(1-\beta)^2}{(2-\beta)\beta}}L_0\eta (1+\varepsilon_{q}) \nonumber \\
		&\qquad + L_1(1-\beta)\eta \sqrt{r}  (1+\varepsilon_{q})\sum_{s=1}^{t}(1-\beta)^{(t-s)}\E\|\nabla F(W_{s-1})\|_*\Big) \nonumber \\
		& \leq \frac{F(W_0)- F^*}{(T+1)\eta(1-\varepsilon_{q})}   + \frac{L_0\eta(1+\varepsilon_{q})^2}{2(1-\varepsilon_{q})} +\frac{L_1\eta (1+\varepsilon_{q})^2}{2(1-\varepsilon_{q})}\frac{1}{T+1}\sum_{t=0}^{T}\E\|\nabla F(W_t)\|_*  \nonumber \\
		&  \quad + \frac{2}{1-\varepsilon_{q}}\Big(\frac{\sqrt{r}\sigma}{(T+1)\beta} + \sqrt{r}\sqrt{\frac{\beta}{2-\beta}}\sigma + \sqrt{r}\sqrt{\frac{(1-\beta)^2}{(2-\beta)\beta}}L_0\eta (1+\varepsilon_{q}) \nonumber \\
		&\qquad + L_1(1-\beta)\eta \sqrt{r}  (1+\varepsilon_{q})\frac{1}{T+1}\sum_{s=0}^{T}\sum_{t=s}^{T}(1-\beta)^{(t-s)}\E\|\nabla F(W_{s-1})\|_* \Big) \nonumber \\
		& \leq \frac{F(W_0)- F^*}{(T+1)\eta(1-\varepsilon_{q})}   + \frac{L_0\eta(1+\varepsilon_{q})^2}{2(1-\varepsilon_{q})} +\frac{L_1\eta (1+\varepsilon_{q})^2}{2(1-\varepsilon_{q})}\frac{1}{T+1}\sum_{t=0}^{T}\E \|\nabla F(W_t)\|_*  \nonumber \\
		&  \quad + \frac{2}{1-\varepsilon_{q}}\Big(\frac{\sqrt{r}\sigma}{T\beta} + \sqrt{r}\sqrt{\frac{\beta}{2-\beta}}\sigma + \sqrt{r}\sqrt{\frac{(1-\beta)^2}{(2-\beta)\beta}}L_0\eta (1+\varepsilon_{q})\Big) \nonumber \\
		&\qquad + \frac{2L_1(1-\beta)\eta \sqrt{r}  (1+\varepsilon_{q})}{\beta(1-\varepsilon_{q})}\frac{1}{T+1}\sum_{t=0}^{T}\E\|\nabla F(W_t)\|_*, 
	\end{align}
	
	Let $0<\eta\leq \min\big(\frac{1-\varepsilon_{q}}{2L_1(1+\varepsilon_{q})^2},\frac{\beta(1-\varepsilon_{q})}{8L_1\sqrt{r}(1+\varepsilon_{q})(1-\beta)}\big)$, we have
	\begin{align}
		\frac{L_1\eta (1+\varepsilon_{q})^2}{2(1-\varepsilon_{q})} \leq \frac{1}{4}, \quad \frac{2L_1(1-\beta)\eta \sqrt{r}  (1+\varepsilon_{q})}{\beta(1-\varepsilon_{q})} \leq \frac{1}{4}.
	\end{align}
	By using the above inequality~(\ref{eq:G7}), then we have
	\begin{align}
		\frac{1}{T+1}\sum_{t=0}^{T}\E\|\nabla F(W_t)\|_*
		& \leq \frac{F(W_0)- F^*}{(T+1)\eta(1-\varepsilon_{q})}   + \frac{L_0\eta(1+\varepsilon_{q})^2}{2(1-\varepsilon_{q})} +\frac{1}{2(T+1)}\sum_{t=0}^{T}\|\nabla F(W_t)\|_*  \nonumber \\
		&  \quad + \frac{2}{1-\varepsilon_{q}}\Big(\frac{\sqrt{r}\sigma}{T\beta} + \sqrt{r}\sqrt{\frac{\beta}{2-\beta}}\sigma + \sqrt{r}\sqrt{\frac{(1-\beta)^2}{(2-\beta)\beta}}L_0\eta (1+\varepsilon_{q})\Big).
	\end{align}
	Thus, we can obtain
	\begin{align}
		\frac{1}{T+1}\sum_{t=0}^{T}\E\|\nabla F(W_t)\|_*
		& \leq  \frac{2((W_0)- F^*)}{(T+1)\eta(1-\varepsilon_{q})}   + \frac{L_0\eta(1+\varepsilon_{q})^2}{1-\varepsilon_{q}} \nonumber \\
		&\quad + \frac{4}{1-\varepsilon_{q}}\Big(\frac{\sqrt{r}\sigma}{T\beta} + \sqrt{r}\sqrt{\frac{\beta}{2-\beta}}\sigma + \sqrt{r}\sqrt{\frac{(1-\beta)^2}{(2-\beta)\beta}}L_0\eta (1+\varepsilon_{q})\Big).
	\end{align}
	
	Let $\eta=O(\frac{1}{T^{2/3}})$, $\beta=O(\frac{1}{T^{2/3}})$, $L_0=O(1)$ and $\sigma=O(1)$, we have
	\begin{align}
		\frac{1}{T+1}\sum_{t=0}^{T}\E \|\nabla F(W_t)\|_*
		& \leq \frac{2((W_0)- F^*)}{(T+1)\eta(1-\varepsilon_{q})}   + \frac{L_0\eta(1+\varepsilon_{q})^2}{1-\varepsilon_{q}} \nonumber \\
		&\quad + \frac{4}{1-\varepsilon_{q}}\Big(\frac{\sqrt{r}\sigma}{T\beta} + \sqrt{r}\sqrt{\frac{\beta}{2-\beta}}\sigma + \sqrt{r}\sqrt{\frac{(1-\beta)^2}{(2-\beta)\beta}}L_0\eta (1+\varepsilon_{q})\Big) \nonumber \\
		& = O\big(\frac{1}{(1-\varepsilon_{q})T^{1/3}}\big).
	\end{align}

\end{proof}

\subsection{Convergence Analysis of LiMuon Algorithm with \textcolor{blue}{Newton-Schulz} and Option \#2 under Lipschitz Smoothness Assumption}
In this subsection, we provide the convergence analysis of LiMuon algorithm with \textbf{Newton-Schulz} and Option \#2 under Lipschitz smoothness assumption. For notational simplicity, let $Z_t = \nabla f(W_t;\xi_t)- \nabla f(W_{t-1};\xi_t)-(\nabla F(W_t)-\nabla F(W_{t-1}))$,
$\delta_t =\nabla F(W_t)- \nabla f(W_t;\xi_t) $ and $\theta_t = M_{t}- \hat{M}_{t}$ for all $t\geq1$.

\begin{lemma} \label{lem:H1}
	Assume the stochastic gradient estimate $M_t$ be generated from Algorithm \ref{alg:3} under  \textbf{Option \#2}. Let $\zeta_t = M_t-\nabla F(W_t)$, $\eta_t=\eta$, $\beta_t=\beta$,$s\geq 2$, $\hat{r}\geq 2$, and $s+\hat{r}\leq r$. Under the assumption~\ref{ass:ss1}, we have
	\begin{align}
		\sum_{t=0}^{T}\E\|\zeta_t\|_*
		\leq \frac{\sqrt{r}\sigma}{\beta} + (T+1)\sqrt{r}\sqrt{\frac{\beta}{2-\beta}}\sigma + (T+1)\sqrt{r}\sqrt{\frac{(1-\beta)^2}{(2-\beta)\beta}}L\eta (1+\varepsilon_{q}) + \frac{\sqrt{r}\gamma(1-\beta)}{\beta} \sum_{t=0}^{T} \E\|M_t\|_\F ,
	\end{align}
	where $\gamma=\tau\left(1 + \frac{\hat{r}}{s - 1} \right)^{\frac{1}{2}}$ with $\tau\in(0,1)$.
\end{lemma}

\begin{proof}
	By the definition of $M_t$ in Algorithm \ref{alg:3} with Option \#2, we have
	\begin{align}
		M_t - \hat{M}_{t-1} = -\beta_t\hat{M}_{t-1}+ \beta_t \nabla f(W_t;\xi_t) + (1-\beta_t)\big( \nabla f(W_t;\xi_t)- \nabla f(W_{t-1};\xi_t)\big).
	\end{align}
	Let $\beta_t=\beta$, we have
	\begin{align}
		\zeta_t & = \nabla F(W_t)- M_t \nonumber \\
		& = \nabla F(W_{t-1}) - \hat{M}_{t-1} + \nabla F(W_t)-\nabla F(W_{t-1}) -(M_t-\hat{M}_{t-1}) \nonumber \\
		& = (1-\beta)(\nabla F(W_{t-1}) - \hat{M}_{t-1}) + \beta(\nabla F(W_t)- \nabla f(W_t;\xi_t)) \nonumber \\
		& \quad \quad - (1-\beta)\big( \nabla f(W_t;\xi_t)- \nabla f(W_{t-1};\xi_t)-(\nabla F(W_t)-\nabla F(W_{t-1}))\big) \nonumber \\
		& = (1-\beta)(\nabla F(W_{t-1}) - \hat{M}_{t-1}) + \beta\delta_t - (1-\beta)Z_t \nonumber \\
		& = (1-\beta) (\nabla F(W_{t-1}) -  M_{t-1} + M_{t-1}- \hat{M}_{t-1}) + \beta\delta_t - (1-\beta)Z_t \nonumber \\
		& = (1-\beta) (\nabla F(W_{t-1}) -  M_{t-1}) + (1-\beta)\theta_{t-1} + \beta\delta_t - (1-\beta)Z_t.
	\end{align}
	
	By expanding the recursion, then we can obtain
	\begin{align}
		\zeta_t = (1-\beta)^t\zeta_{0}  + (1-\beta)\sum_{s=1}^{t}(1-\beta)^{t-s}\theta_{s-1}+ \beta\sum_{s=1}^{t}(1-\beta)^{t-s}\delta_s -(1-\beta)\sum_{s=1}^{t}(1-\beta)^{t-s}Z_s.
	\end{align}
	By using the above lemma~\ref{lem:A4}, we have
	\begin{align} \label{eq:H1}
		(1-\beta)\E\|\sum_{s=1}^{t}(1-\beta)^{t-s}\theta_{s-1}\|_\F & \leq (1-\beta)\sum_{s=1}^{t}(1-\beta)^{t-s}\E\|\theta_{s-1}\|_\F \nonumber \\
		& \leq \gamma(1-\beta)\sum_{s=1}^{t}(1-\beta)^{t-s}\E\|M_{s-1}
		\|_\F.
	\end{align}
	
	By using the proof of the above lemma~\ref{lem:F2}, we have
	\begin{align} \label{eq:H2}
		\E\|\zeta_{0}\|_\F \leq \sigma.
	\end{align}
	Meanwhile, we have
	\begin{align} \label{eq:H3}
		\beta\E\|\sum_{s=1}^{t}(1-\beta)^{t-s}\delta_s\|_\F \leq  \sqrt{\frac{\beta}{2-\beta}}\sigma.
	\end{align}
	We also have
	\begin{align} \label{eq:H4}
		(1-\beta)\E\|\sum_{s=1}^{t}(1-\beta)^{t-s}Z_s\|_\F  \leq  \sqrt{\frac{(1-\beta)^2}{(2-\beta)\beta}}L\eta (1+\varepsilon_{q}).
	\end{align}
	
	By using the above inequalities~(\ref{eq:H1}), (\ref{eq:H2}), (\ref{eq:H3}) and (\ref{eq:H4}), we have
	\begin{align}
		&\E\|\zeta_t\|_\F
		= \E\|(1-\beta)^t\zeta_{0}  + (1-\beta)\sum_{s=1}^{t}(1-\beta)^{t-s}\theta_{s-1}+ \beta\sum_{s=1}^{t}(1-\beta)^{t-s}\delta_s -(1-\beta)\sum_{s=1}^{t}(1-\beta)^{t-s}Z_s\|_\F \nonumber \\
		& \leq \E\|(1-\beta)^t\zeta_{0}\|_\F + (1-\beta)\E\|\sum_{s=1}^{t}(1-\beta)^{t-s}\theta_{s-1}\|_\F + \beta\E\|\sum_{s=1}^{t}(1-\beta)^{t-s}\delta_s\|_\F +(1-\beta)\E\|\sum_{s=1}^{t}(1-\beta)^{t-s}Z_s\|_\F \nonumber \\
		& \leq (1-\beta)^t\sigma + \sqrt{\frac{\beta}{2-\beta}}\sigma + \sqrt{\frac{(1-\beta)^2}{(2-\beta)\beta}}L\eta (1+\varepsilon_{q}) + \gamma(1-\beta)\sum_{s=1}^{t}(1-\beta)^{t-s}\E\|M_{s-1}\|_\F).
	\end{align}
	
	Let $M_{-1}=0$, then we have
	\begin{align}
		\sum_{t=0}^{T}\E\|\zeta_t\|_\F
		& \leq \sum_{t=0}^{T}\Big( (1-\beta)^t\sigma + \sqrt{\frac{\beta}{2-\beta}}\sigma + \sqrt{\frac{(1-\beta)^2}{(2-\beta)\beta}}L\eta (1+\varepsilon_{q}) + \gamma(1-\beta)\sum_{s=1}^{t}(1-\beta)^{t-s}\E\|M_{s-1}\|_\F \Big) \nonumber \\
		& \leq \frac{\sigma}{\beta} + (T+1)\sqrt{\frac{\beta}{2-\beta}}\sigma + (T+1)\sqrt{\frac{(1-\beta)^2}{(2-\beta)\beta}}L\eta(1+\varepsilon_{q}) + \gamma(1-\beta) \sum_{s=0}^{T}\sum_{t=s}^{T}(1-\beta)^{t-s} \E\|M_{s-1}\|_\F \nonumber \\
		& \leq \frac{\sigma}{\beta} + (T+1)\sqrt{\frac{\beta}{2-\beta}}\sigma + (T+1)\sqrt{\frac{(1-\beta)^2}{(2-\beta)\beta}}L\eta (1+\varepsilon_{q}) + \frac{\gamma(1-\beta)}{\beta} \sum_{t=0}^{T} \E\|M_t\|_\F .
	\end{align}
	
	Since $\|\zeta_t\|_* \leq \sqrt{r}\|\zeta_t\|_\F$, we have 
	\begin{align}
		\sum_{t=0}^{T}\E\|\zeta_t\|_*
	\leq \frac{\sqrt{r}\sigma}{\beta} + (T+1)\sqrt{r}\sqrt{\frac{\beta}{2-\beta}}\sigma + (T+1)\sqrt{r}\sqrt{\frac{(1-\beta)^2}{(2-\beta)\beta}}L\eta (1+\varepsilon_{q}) + \frac{\sqrt{r}\gamma(1-\beta)}{\beta} \sum_{t=0}^{T} \E\|M_t\|_\F .
	\end{align}
	
\end{proof}

\begin{theorem} \label{th:H1}
	Under the Assumptions~\ref{ass:ss1},~\ref{ass:v},~\ref{ass:l}, the sequence $\{W_t\}_{t=0}^T$ be generated
	from Algorithm \ref{alg:3} with \textbf{Option} \#2. Let $\beta_t=\beta \in (0,1)$, $\eta_t=\eta$ for all $t\geq 0$, and $s\geq 2$, $\hat{r}\geq 2$, $s+\hat{r}\leq r$, and $\gamma \leq  \frac{\beta(1-\varepsilon_{q})}{(3+\varepsilon_{q})\sqrt{r}(1-\beta)}$, we have
	\begin{align}
		\frac{1}{T+1}\sum_{t=0}^{T}\E \|\nabla F(W_t)\|_* &\leq \frac{2(F(W_0)- F^*)}{T\eta(1-\varepsilon_{q})}  + \frac{\eta L(1+ \varepsilon_{q})^2}{1-\varepsilon_{q}} + \frac{3+\varepsilon_{q}}{1-\varepsilon_{q}}\Big( \frac{\sqrt{r}\sigma}{T\beta} \nonumber \\
		& \quad  + \sqrt{r}\sqrt{\frac{\beta}{2-\beta}}\sigma + \sqrt{r}\sqrt{\frac{(1-\beta)^2}{(2-\beta)\beta}}L\eta (1+\varepsilon_{q})  \Big),
	\end{align}
	where $\gamma=\tau\left(1 + \frac{\hat{r}}{s - 1} \right)^{\frac{1}{2}}$ with $\tau\in(0,1)$.
\end{theorem}

\begin{proof}
	
By using the above Lemma~\ref{lem:F1},  we have
	\begin{align}
		F(W_{t+1}) \leq & F(W_t)+ \langle\nabla F(W_t), W_{t+1}-W_t \rangle+\frac{L}{2}\|W_{t+1}-W_t \|_{op}^2 \nonumber \\
		= & F(W_t)- \eta_t\langle\nabla F(W_t), O_t \rangle+\frac{L}{2}\eta_t^2\|O_t\|_{op}^2 \nonumber \\
		= & F(W_t) - \eta_t\langle M_t, O_t \rangle  -\eta_t\langle \nabla F(W_t)-M_t , O_t \rangle + \frac{L}{2}\eta_t^2 \|O_t\|_{op}^2 \nonumber \\
		\leq & F(W_t) - \eta_t\langle M_t, O_t \rangle  +\eta_t\| \nabla F(W_t)-M_t\|_* \|O_t\|_{op} + \frac{L}{2}\eta_t^2 \|O_t\|_{op}^2 \nonumber \\
		= & F(W_t) - \eta_t\langle M_t, P_t \rangle +\eta_t\langle M_t, P_t-O_t \rangle +\eta_t\| \nabla F(W_t)-M_t\|_* \|O_t\|_{op} + \frac{L}{2}\eta_t^2 \|O_t\|_{op}^2 \nonumber \\
		= & F(W_t) - \eta_t\| M_t\|_* +\eta_t\langle M_t, P_t-O_t \rangle +\eta_t\| \nabla F(W_t)-M_t\|_* \|O_t\|_{op} + \frac{L}{2}\eta_t^2 \|O_t\|_{op}^2 \nonumber \\
		\leq & F(W_t) - \eta_t\| M_t\|_* +\eta_t\| M_t\|_* \|P_t-O_t \|_{op} +\eta_t\| \nabla F(W_t)-M_t\|_* \|O_t\|_{op} + \frac{L}{2}\eta_t^2 \|O_t\|_{op}^2,
	\end{align}
	where the second inequality is due to $\langle M_t , U_tV_t^\top \rangle =\|M_t\|_*$.
	
	Since $P_t=U_tV_t^\top$, $\varepsilon_{q,t}=\|O_t-P_t\|_{op}$ with $\varepsilon_{q}=\sup_{t} \varepsilon_{q,t}$, we have 
	\begin{align}
		\|P_t\|_{op}=1, \quad \|O_t\|_{op} \leq 1+\varepsilon_{q,t}\leq 1+\varepsilon_{q},
	\end{align}
	where $\varepsilon_{q}\in (0,1)$.
	Then we can obtain 
	\begin{align}
		F(W_{t+1}) 
		\leq &  F(W_t) - \eta_t\| M_t\|_* +\eta_t\| M_t\|_* \|P_t-O_t \|_{op} +\eta_t\| \nabla F(W_t)-M_t\|_* \|O_t\|_{op} + \frac{L}{2}\eta_t^2 \|O_t\|_{op}^2 \nonumber \\
		\leq & F(W_t) - \eta_t(1-\varepsilon_{q})\| M_t\|_*  +\eta_t(1+\varepsilon_{q})\| \nabla F(W_t)-M_t\|_*  + \frac{L}{2}\eta_t^2 (1+\varepsilon_{q})^2 \nonumber \\
		\leq & F(W_t) - \frac{\eta_t}{2}(1-\varepsilon_{q})\| M_t\|_* -\frac{\eta_t}{2}(1-\varepsilon_{q})\| \nabla F(W_t)\|_* +\frac{\eta_t}{2}(3+\varepsilon_{q})\| \nabla F(W_t)-M_t\|_*  \nonumber \\
		& + \frac{L}{2}\eta_t^2 (1+\varepsilon_{q})^2,
	\end{align}
	where the last inequality holds by
	$\|M_t\|_*\geq \|\nabla F(W_t)\|_* - \|\nabla F(W_t)-M_t\|_*$. 
	
	Let $\eta_t=\eta$ and $\zeta_t=\nabla F(W_t)-M_t$, then we have
	\begin{align}
		\E\|\nabla F(W_t)\|_* \leq & \E\big[\frac{2(F(W_t)- F(W_{t+1}))}{\eta(1-\varepsilon_{q})}  -\|M_t\|_*  + \frac{\eta L(1+ \varepsilon_{q})^2}{1-\varepsilon_{q}} + \frac{3+\varepsilon_{q}}{1-\varepsilon_{q}}\|\zeta_t\|_* \big].
	\end{align}
	Then we have
	\begin{align}
		& \frac{1}{T+1}\sum_{t=0}^{T}\E\|\nabla F(W_t)\|_* \nonumber \\
		& \leq \E[\frac{1}{T+1}\sum_{t=0}^{T}\frac{2(F(W_t)- F(W_{t+1}))}{\eta(1-\varepsilon_{q})} -\frac{1}{T+1}\sum_{t=0}^{T}\|M_t\|_* + \frac{\eta L(1+ \varepsilon_{q})^2}{1-\varepsilon_{q}} + \frac{1}{T+1}\sum_{t=0}^{T}\frac{3+\varepsilon_{q}}{1-\varepsilon_{q}} \|\zeta_t\|_* ]\nonumber \\
		& \leq \frac{2(F(W_0)- F^*)}{(T+1)\eta(1-\varepsilon_{q})}  -\frac{1}{T+1}\sum_{t=0}^{T}\E \|M_t\|_*+ \frac{\eta L(1+ \varepsilon_{q})^2}{1-\varepsilon_{q}} + \frac{1}{T+1}\sum_{t=0}^{T}\frac{3+\varepsilon_{q}}{1-\varepsilon_{q}}\E \|\zeta_t\|_* \nonumber \\
		& \leq \frac{2(F(W_0)- F^*)}{(T+1)\eta(1-\varepsilon_{q})}  -\frac{1}{T+1}\sum_{t=0}^{T}\E\|M_t\|_*+ \frac{\eta L(1+ \varepsilon_{q})^2}{1-\varepsilon_{q}} + \frac{1}{T+1}\frac{3+\varepsilon_{q}}{1-\varepsilon_{q}}\Big( \frac{\sqrt{r}\sigma}{\beta} + (T+1)\sqrt{r}\sqrt{\frac{\beta}{2-\beta}}\sigma \nonumber \\
		& \quad + (T+1)\sqrt{r}\sqrt{\frac{(1-\beta)^2}{(2-\beta)\beta}}L\eta (1+\varepsilon_{q}) + \frac{\sqrt{r}\gamma(1-\beta)}{\beta} \sum_{t=0}^{T} \E\|M_t\|_\F \Big) \nonumber \\
		&\leq \frac{2(F(W_0)- F^*)}{(T+1)\eta(1-\varepsilon_{q})}  -\frac{1}{T+1}\sum_{t=0}^{T}\E\|M_t\|_*+ \frac{\eta L(1+ \varepsilon_{q})^2}{1-\varepsilon_{q}} + \frac{3+\varepsilon_{q}}{1-\varepsilon_{q}}\Big( \frac{\sqrt{r}\sigma}{(T+1)\beta} + \sqrt{r}\sqrt{\frac{\beta}{2-\beta}}\sigma \nonumber \\
		& \quad + \sqrt{r}\sqrt{\frac{(1-\beta)^2}{(2-\beta)\beta}}L\eta (1+\varepsilon_{q}) + \frac{\sqrt{r}\gamma(1-\beta)}{\beta} \frac{1}{T+1} \sum_{t=0}^{T} \E\|M_t\|_* \Big),
	\end{align}
	where the second last inequality holds by the above lemma~\ref{lem:H1}, and the last inequality is due to $\|M_t\|_\F \leq \|M_t\|_*$.
	
	Let $\gamma \leq \frac{\beta(1-\varepsilon_{q})}{(3+\varepsilon_{q})\sqrt{r}(1-\beta)}$, we have
	\begin{align}
		\frac{1}{T+1}\sum_{t=0}^{T}\E\|\nabla F(W_t)\|_* &\leq \frac{2(F(W_0)- F^*)}{(T+1)\eta(1-\varepsilon_{q})}  + \frac{\eta L(1+ \varepsilon_{q})^2}{1-\varepsilon_{q}} + \frac{3+\varepsilon_{q}}{1-\varepsilon_{q}}\Big( \frac{\sqrt{r}\sigma}{(T+1)\beta} \nonumber \\
		& \quad  + \sqrt{r}\sqrt{\frac{\beta}{2-\beta}}\sigma + \sqrt{r}\sqrt{\frac{(1-\beta)^2}{(2-\beta)\beta}}L\eta (1+\varepsilon_{q})  \Big) \nonumber \\
		&\leq \frac{2(F(W_0)- F^*)}{T\eta(1-\varepsilon_{q})}  + \frac{\eta L(1+ \varepsilon_{q})^2}{1-\varepsilon_{q}} + \frac{3+\varepsilon_{q}}{1-\varepsilon_{q}}\Big( \frac{\sqrt{r}\sigma}{T\beta} \nonumber \\
		& \quad  + \sqrt{r}\sqrt{\frac{\beta}{2-\beta}}\sigma + \sqrt{r}\sqrt{\frac{(1-\beta)^2}{(2-\beta)\beta}}L\eta (1+\varepsilon_{q})  \Big).
	\end{align}
	
	Let $\eta=O(\frac{1}{T^{2/3}})$ and $\beta=O(\frac{1}{T^{2/3}})$, we have
	\begin{align}
		\frac{1}{T+1}\sum_{t=0}^{T}\E\|\nabla F(W_t)\|_*
		& \leq \frac{2(F(W_0)- F^*)}{T\eta(1-\varepsilon_{q})}  + \frac{\eta L(1+ \varepsilon_{q})^2}{1-\varepsilon_{q}} + \frac{3+\varepsilon_{q}}{1-\varepsilon_{q}}\Big( \frac{\sqrt{r}\sigma}{T\beta} \nonumber \\
		& \quad  + \sqrt{r}\sqrt{\frac{\beta}{2-\beta}}\sigma + \sqrt{r}\sqrt{\frac{(1-\beta)^2}{(2-\beta)\beta}}L\eta (1+\varepsilon_{q})  \Big) \nonumber \\
		&  = O\big(\frac{1}{(1-\varepsilon_{q})T^{1/3}}\big).
	\end{align}

\end{proof}

\subsection{Convergence Analysis of LiMuon Algorithm with \textcolor{blue}{Newton-Schulz} and  Option \#2 under Generalized Smoothness Assumption}
In this subsection, we provide the convergence analysis of LiMuon algorithm with \textbf{Newton-Schulz} and \textbf{Option} \#2 under generalized smoothness assumption.

\begin{lemma} \label{lem:I1}
	Assume the stochastic gradient estimate $M_t$ be generated from Algorithm \ref{alg:3} under  \textbf{Option \#2}. Let $\zeta_t = M_t-\nabla F(W_t)$, $\eta_t=\eta$,  $\beta_t=\beta$, $s\geq 2$, $\hat{r}\geq 2$, and  $s+\hat{r}\leq r$. Under the assumption~\ref{ass:ss2}, we have
	\begin{align}
		\E\|\zeta_t\|_* & \leq \sqrt{r}(1-\beta)^t\sigma + \sqrt{r}\sqrt{\frac{\beta}{2-\beta}}\sigma + \sqrt{r}\sqrt{\frac{(1-\beta)^2}{(2-\beta)\beta}}L_0\eta (1+\varepsilon_{q}) \nonumber \\
		& \quad + \sqrt{r}L_1(1-\beta)\eta (1+\varepsilon_{q})\sum_{s=1}^{t}(1-\beta)^{(t-s)}\E\|\nabla F(W_{s-1})\|_* +\sqrt{r}\gamma(1-\beta)\sum_{s=1}^{t}(1-\beta)^{t-s}\E\|M_{s-1}\|_*,
	\end{align}
	where $\gamma=\tau\left(1 + \frac{\hat{r}}{s - 1} \right)^{\frac{1}{2}}$ and $\tau\in(0,1)$.
\end{lemma}

\begin{proof}
	From the above lemma~\ref{lem:H1}, we have
	\begin{align}
		\zeta_t = (1-\beta)^t\zeta_{0}  + (1-\beta)\sum_{s=1}^{t}(1-\beta)^{t-s}\theta_{s-1}+ \beta\sum_{s=1}^{t}(1-\beta)^{t-s}\delta_s -(1-\beta)\sum_{s=1}^{t}(1-\beta)^{t-s}Z_s.
	\end{align}
	Then from both lemmas~\ref{lem:G1} and~\ref{lem:H1}, we have
	\begin{align}
		&\E\|\zeta_t\|_\F
		= \E\|(1-\beta)^t\zeta_{0}  + (1-\beta)\sum_{s=1}^{t}(1-\beta)^{t-s}\theta_{s-1}+ \beta\sum_{s=1}^{t}(1-\beta)^{t-s}\delta_s -(1-\beta)\sum_{s=1}^{t}(1-\beta)^{t-s}Z_s\|_\F \nonumber \\
		& \leq \E\|(1-\beta)^t\zeta_{0}\|_\F + (1-\beta)\E\|\sum_{s=1}^{t}(1-\beta)^{t-s}\theta_{s-1}\|_\F + \beta\E\|\sum_{s=1}^{t}(1-\beta)^{t-s}\delta_s\|_\F +(1-\beta)\E\|\sum_{s=1}^{t}(1-\beta)^{t-s}Z_s\|_\F \nonumber \\
		& \leq (1-\beta)^t\sigma + \sqrt{\frac{\beta}{2-\beta}}\sigma + \sqrt{\frac{(1-\beta)^2}{(2-\beta)\beta}}L_0\eta (1+\varepsilon_{q}) + L_1(1-\beta)\eta (1+\varepsilon_{q})\sum_{s=1}^{t}(1-\beta)^{(t-s)}\E\|\nabla F(W_{s-1})\|_* \nonumber \\
		& \quad +\gamma(1-\beta)\sum_{s=1}^{t}(1-\beta)^{t-s}\E\|M_{s-1}\|_\F.
	\end{align}
	
	Since $\|\zeta_t\|_* \leq \sqrt{r}\|\zeta_t\|_\F$, we have 
	\begin{align}
		\E\|\zeta_t\|_*
		& \leq \sqrt{r}(1-\beta)^t\sigma + \sqrt{r}\sqrt{\frac{\beta}{2-\beta}}\sigma + \sqrt{r}\sqrt{\frac{(1-\beta)^2}{(2-\beta)\beta}}L_0\eta (1+\varepsilon_{q}) \nonumber \\
		& \quad + \sqrt{r}L_1(1-\beta)\eta (1+\varepsilon_{q})\sum_{s=1}^{t}(1-\beta)^{(t-s)}\E\|\nabla F(W_{s-1})\|_* +\sqrt{r}\gamma(1-\beta)\sum_{s=1}^{t}(1-\beta)^{t-s}\E\|M_{s-1}\|_\F \nonumber \\
		& \leq \sqrt{r}(1-\beta)^t\sigma + \sqrt{r}\sqrt{\frac{\beta}{2-\beta}}\sigma + \sqrt{r}\sqrt{\frac{(1-\beta)^2}{(2-\beta)\beta}}L_0\eta (1+\varepsilon_{q}) \nonumber \\
		& \quad + \sqrt{r}L_1(1-\beta)\eta (1+\varepsilon_{q})\sum_{s=1}^{t}(1-\beta)^{(t-s)}\E\|\nabla F(W_{s-1})\|_* +\sqrt{r}\gamma(1-\beta)\sum_{s=1}^{t}(1-\beta)^{t-s}\E\|M_{s-1}\|_*,
	\end{align}
	where the last inequality is due to $\|M_{s-1}\|_\F \leq \|M_{s-1}\|_*$. 
\end{proof}

\begin{theorem} \label{th:I1}
	Under the Assumptions~\ref{ass:ss2},~\ref{ass:v},~\ref{ass:l}, the sequence $\{W_t\}_{t=0}^T$ be generated
	from Algorithm \ref{alg:3} with \textbf{Option} \#2. Let $\beta_t=\beta \in (0,1)$, and $\eta_t=\eta \leq \min\big(\frac{1-\varepsilon_{q}}{4L_1(1+\varepsilon_{q})^2},\frac{(1-\varepsilon_{q})\beta}{4L_1\sqrt{r}(1-\beta)(1+\varepsilon_{q})(3+\varepsilon_{q})}\big)$ for all $t\geq 0$, and $s\geq 2$, $\hat{r}\geq 2$, $s+\hat{r}\leq r$, and $\gamma \leq \frac{(1-\varepsilon_{q})\beta}{(3+\varepsilon_{q})\sqrt{r}(1-\beta)}$, we have
	\begin{align}
		 \frac{1}{T+1}\sum_{t=0}^{T}\E \|\nabla F(W_t)\|_*
		& \leq \frac{4(F(W_0)- F^*)}{(T+1)\eta(1-\varepsilon_{q})}  + \frac{2L_0\eta(1+\varepsilon_{q})^2}{1-\varepsilon_{q}}   \nonumber \\
		& \quad + \frac{2(3+\varepsilon_{q})}{1-\varepsilon_{q}}\Big(\frac{\sqrt{r}\sigma}{(T+1)\beta} + \sqrt{r}\sqrt{\frac{\beta}{2-\beta}}\sigma + \sqrt{r}\sqrt{\frac{(1-\beta)^2}{(2-\beta)\beta}}L_0\eta (1+\varepsilon_{q}) \Big) ,
	\end{align}
	where $\gamma=\tau\left(1 + \frac{\hat{r}}{s - 1} \right)^{\frac{1}{2}}$ and $\tau\in(0,1)$.
\end{theorem}

\begin{proof}
	This proof basically follows the proof of Theorem~\ref{th:G1}.
	By using the above Lemma~\ref{lem:G0}, we have
	\begin{align}
		F(W_{t+1}) \leq & F(W_t)+ \langle\nabla F(W_t), W_{t+1}-W_t \rangle+\frac{L_0+L_1\|\nabla F(W_t)\|_*}{2}\|W_{t+1}-W_t \|_{op}^2 \nonumber \\
		\leq & F(W_t)- \eta_t\langle\nabla F(W_t), O_t \rangle+\frac{L_0+L_1\|\nabla F(W_t)\|_*}{2}\eta_t^2\|O_t\|_{op}^2 \nonumber \\
		\leq & F(W_t) - \eta_t\langle M_t, O_t \rangle  -\eta_t\langle \nabla F(W_t)-M_t , O_t\rangle + \frac{L_0+L_1\|\nabla F(W_t)\|_*}{2}\eta_t^2 \|O_t\|_{op} \nonumber \\
		= & F(W_t) - \eta_t\langle M_t, P_t \rangle + \eta_t\langle M_t, P_t-O_t \rangle -\eta_t\langle \nabla F(W_t)-M_t , O_t\rangle + \frac{L_0+L_1\|\nabla F(W_t)\|_*}{2}\eta_t^2 \|O_t\|^2_{op} \nonumber \\
		\leq & F(W_t)-  \eta_t \|M_t\|_* + \eta_t\|M_t\|_* \|P_t-O_t \|_{op} +\eta_t\|\nabla F(W_t)-M_t\|_* \|O_t\|_{op} \nonumber \\
		&\quad + \frac{L_0+L_1\|\nabla F(W_t)\|_*}{2}\eta_t^2 \|O_t\|^2_{op},
	\end{align}
	where the last inequality holds by $\langle M_t , P_t \rangle=\langle M_t , U_tV_t^\top \rangle =\|M_t\|_*$.
	
	Since $P_t=U_tV_t^\top$, $\varepsilon_{q,t}=\|O_t-P_t\|_{op}$ with $\varepsilon_{q}=\sup_{t} \varepsilon_{q,t}$, we have 
	\begin{align}
		\|P_t\|_{op}=1, \quad \|O_t\|_{op} \leq 1+\varepsilon_{q,t}\leq 1+\varepsilon_{q},
	\end{align}
	where $\varepsilon_{q}\in (0,1)$.
	Thus, we can obtain 
	\begin{align}
		F(W_{t+1}) \leq  & F(W_t)-  \eta_t \|M_t\|_* + \eta_t\|M_t\|_* \|P_t-O_t \|_{op} +\eta_t\|\nabla F(W_t)-M_t\|_* \|O_t\|_{op} \nonumber \\
		&\quad + \frac{L_0+L_1\|\nabla F(W_t)\|_*}{2}\eta_t^2 \|O_t\|^2_{op} \nonumber \\
		\leq & F(W_t)-  \eta_t \|M_t\|_* + \eta_t\|M_t\|_* \varepsilon_{q} +\eta_t\|\nabla F(W_t)-M_t\|_* (1+\varepsilon_{q}) \nonumber \\
		&\quad + \frac{L_0+L_1\|\nabla F(W_t)\|_*}{2}\eta_t^2 (1+\varepsilon_{q})^2 \nonumber \\
		\leq & F(W_t)- \frac{\eta_t}{2} (1-\varepsilon_{q})\|M_t\|_*- \frac{\eta_t}{2} (1-\varepsilon_{q})\|\nabla F(W_t)\|_* +\frac{\eta_t}{2}(3+\varepsilon_{q})\|\nabla F(W_t)-M_t\|_*  \nonumber \\
		& + \frac{L_0+L_1\|\nabla F(W_t)\|_*}{2}\eta_t^2 (1+\varepsilon_{q})^2, 
	\end{align}
	where the last inequality holds by $\|M_t\|_*\geq \|\nabla F(W_t)\|_* - \|\nabla F(W_t)-M_t\|_*$.
	
	Let $\eta_t=\eta$ and $\zeta_t=\nabla F(W_t)-M_t$, we have
	\begin{align}
		\E\|\nabla F(W_t)\|_* & \leq \E\big[\frac{2(F(W_t)- F(W_{t+1}))}{\eta(1-\varepsilon_{q})} - \|M_t\|_*  + \frac{L_0\eta(1+\varepsilon_{q})^2}{1-\varepsilon_{q}} +\frac{L_1\eta (1+\varepsilon_{q})^2}{1-\varepsilon_{q}}\|\nabla F(W_t)\|_* + \frac{3+\varepsilon_{q}}{1-\varepsilon_{q}}\|\zeta_t\|_* \big] \nonumber \\
		& \leq \E\big[\frac{2(F(W_t)- F(W_{t+1}))}{\eta(1-\varepsilon_{q})} \big] - \|M_t\|_*  + \frac{L_0\eta(1+\varepsilon_{q})^2}{1-\varepsilon_{q}} +\frac{L_1\eta (1+\varepsilon_{q})^2}{1-\varepsilon_{q}}\E\|\nabla F(W_t)\|_*  \nonumber \\
		& \quad + \frac{3+\varepsilon_{q}}{1-\varepsilon_{q}}\Big(\sqrt{r}(1-\beta)^t\sigma + \sqrt{r}\sqrt{\frac{\beta}{2-\beta}}\sigma + \sqrt{r}\sqrt{\frac{(1-\beta)^2}{(2-\beta)\beta}}L_0\eta (1+\varepsilon_{q}) \nonumber \\
		& \quad + \sqrt{r}L_1(1-\beta)\eta (1+\varepsilon_{q})\sum_{s=1}^{t}(1-\beta)^{(t-s)}\E\|\nabla F(W_{s-1})\|_* +\sqrt{r}\gamma(1-\beta)\sum_{s=1}^{t}(1-\beta)^{t-s}\E\|M_{s-1}\|_*\Big),
	\end{align}
	where the last inequality holds by the above lemma~\ref{lem:I1}.

	Since $M_{-1}=0$, then we have
	\begin{align}
		& \frac{1}{T+1}\sum_{t=0}^{T}\E\|\nabla F(W_t)\|_* \nonumber \\
		& \leq \frac{2(F(W_0)- F^*)}{(T+1)\eta(1-\varepsilon_{q})} -\frac{1}{T+1}\sum_{t=0}^{T}\|M_t\|_* + \frac{L_0\eta(1+\varepsilon_{q})^2}{1-\varepsilon_{q}} +\frac{L_1\eta (1+\varepsilon_{q})^2}{1-\varepsilon_{q}}\frac{1}{T+1}\sum_{t=0}^{T}\E\|\nabla F(W_t)\|_*  \nonumber \\
		& \quad + \frac{3+\varepsilon_{q}}{1-\varepsilon_{q}}\frac{1}{T+1}\sum_{t=0}^{T}\Big(\sqrt{r}(1-\beta)^t\sigma + \sqrt{r}\sqrt{\frac{\beta}{2-\beta}}\sigma + \sqrt{r}\sqrt{\frac{(1-\beta)^2}{(2-\beta)\beta}}L_0\eta (1+\varepsilon_{q}) \nonumber \\
		& \quad + \sqrt{r}L_1(1-\beta)\eta (1+\varepsilon_{q})\sum_{s=1}^{t}(1-\beta)^{(t-s)}\E\|\nabla F(W_{s-1})\|_* +\sqrt{r}\gamma(1-\beta)\sum_{s=1}^{t}(1-\beta)^{t-s}\E\|M_{s-1}\|_*\Big) \nonumber \\
		& \leq \frac{2(F(W_0)- F^*)}{(T+1)\eta(1-\varepsilon_{q})} -\frac{1}{T+1}\sum_{t=0}^{T}\|M_t\|_* + \frac{L_0\eta(1+\varepsilon_{q})^2}{1-\varepsilon_{q}} +\frac{L_1\eta (1+\varepsilon_{q})^2}{1-\varepsilon_{q}}\frac{1}{T+1}\sum_{t=0}^{T}\E\|\nabla F(W_t)\|_* \nonumber \\
		& \quad + \frac{3+\varepsilon_{q}}{1-\varepsilon_{q}}\Big(\frac{\sqrt{r}\sigma}{(T+1)\beta} + \sqrt{r}\sqrt{\frac{\beta}{2-\beta}}\sigma + \sqrt{r}\sqrt{\frac{(1-\beta)^2}{(2-\beta)\beta}}L_0\eta (1+\varepsilon_{q}) \nonumber \\
		& \quad + \frac{\sqrt{r}L_1(1-\beta)\eta (1+\varepsilon_{q})}{\beta}\frac{1}{T+1}\sum_{t=0}^{T}\E\|\nabla F(W_t)\|_* +\frac{\sqrt{r}\gamma(1-\beta)}{\beta}\frac{1}{T+1}\sum_{t=0}^{T}\E\|M_t\|_*\Big).
	\end{align}
	
	Let $0<\eta\leq \min\big(\frac{1-\varepsilon_{q}}{4L_1(1+\varepsilon_{q})^2},\frac{(1-\varepsilon_{q})\beta}{4L_1\sqrt{r}(1-\beta)(1+\varepsilon_{q})(3+\varepsilon_{q})}\big)$ and $\gamma \leq \frac{(1-\varepsilon_{q})\beta}{(3+\varepsilon_{q})\sqrt{r}(1-\beta)}$, we have
	\begin{align}
		\frac{L_1\eta (1+\varepsilon_{q})^2}{1-\varepsilon_{q}} \leq \frac{1}{4}, \quad \frac{(3+\varepsilon_{q})\sqrt{r}L_1(1-\beta)\eta (1+\varepsilon_{q})}{(1-\varepsilon_{q})\beta} \leq \frac{1}{4}, \quad \frac{(3+\varepsilon_{q})\sqrt{r}\gamma(1-\beta)}{(1-\varepsilon_{q})\beta} \leq 1.
	\end{align}
	Then we have
	\begin{align} \label{eq:I1}
		\frac{1}{T+1}\sum_{t=0}^{T}\E\|\nabla F(W_t)\|_*
		& \leq \frac{2(F(W_0)- F^*)}{(T+1)\eta(1-\varepsilon_{q})}  + \frac{L_0\eta(1+\varepsilon_{q})^2}{1-\varepsilon_{q}}  + \frac{1}{2(T+1)}\sum_{t=0}^{T}\E\|\nabla F(W_t)\|_* \nonumber \\
		& \quad + \frac{3+\varepsilon_{q}}{1-\varepsilon_{q}}\Big(\frac{\sqrt{r}\sigma}{(T+1)\beta} + \sqrt{r}\sqrt{\frac{\beta}{2-\beta}}\sigma + \sqrt{r}\sqrt{\frac{(1-\beta)^2}{(2-\beta)\beta}}L_0\eta (1+\varepsilon_{q}) \Big).
	\end{align}
	By rewrite the above inequality~(\ref{eq:I1}), we can obtain
	\begin{align}
		\frac{1}{T+1}\sum_{t=0}^{T}\E\|\nabla F(W_t)\|_*
		& \leq  \frac{4(F(W_0)- F^*)}{(T+1)\eta(1-\varepsilon_{q})}  + \frac{2L_0\eta(1+\varepsilon_{q})^2}{1-\varepsilon_{q}}   \nonumber \\
		& \quad + \frac{2(3+\varepsilon_{q})}{1-\varepsilon_{q}}\Big(\frac{\sqrt{r}\sigma}{(T+1)\beta} + \sqrt{r}\sqrt{\frac{\beta}{2-\beta}}\sigma + \sqrt{r}\sqrt{\frac{(1-\beta)^2}{(2-\beta)\beta}}L_0\eta (1+\varepsilon_{q}) \Big).
	\end{align}
	
	Let $\eta=O(\frac{1}{T^{2/3}})$,  $\beta=O(\frac{1}{T^{2/3}})$, $L_0=O(1)$ and $\sigma=O(1)$, we have
	\begin{align}
		\frac{1}{T+1}\sum_{t=0}^{T}\E \|\nabla F(W_t)\|_*
		& \leq \frac{4(F(W_0)- F^*)}{(T+1)\eta(1-\varepsilon_{q})}  + \frac{2L_0\eta(1+\varepsilon_{q})^2}{1-\varepsilon_{q}}   \nonumber \\
		& \quad + \frac{2(3+\varepsilon_{q})}{1-\varepsilon_{q}}\Big(\frac{\sqrt{r}\sigma}{(T+1)\beta} + \sqrt{r}\sqrt{\frac{\beta}{2-\beta}}\sigma + \sqrt{r}\sqrt{\frac{(1-\beta)^2}{(2-\beta)\beta}}L_0\eta (1+\varepsilon_{q}) \Big) \nonumber \\
		&  =O\big(\frac{1}{(1-\varepsilon_{q})T^{1/3}}\big).
	\end{align}

\end{proof}

\section{ Hyper-parameters in Experiments and Additional Experimental Results } \label{sec:hypara}
\begin{figure*}[ht]
	\centering
	\subfigure{\includegraphics[width=0.32\textwidth]{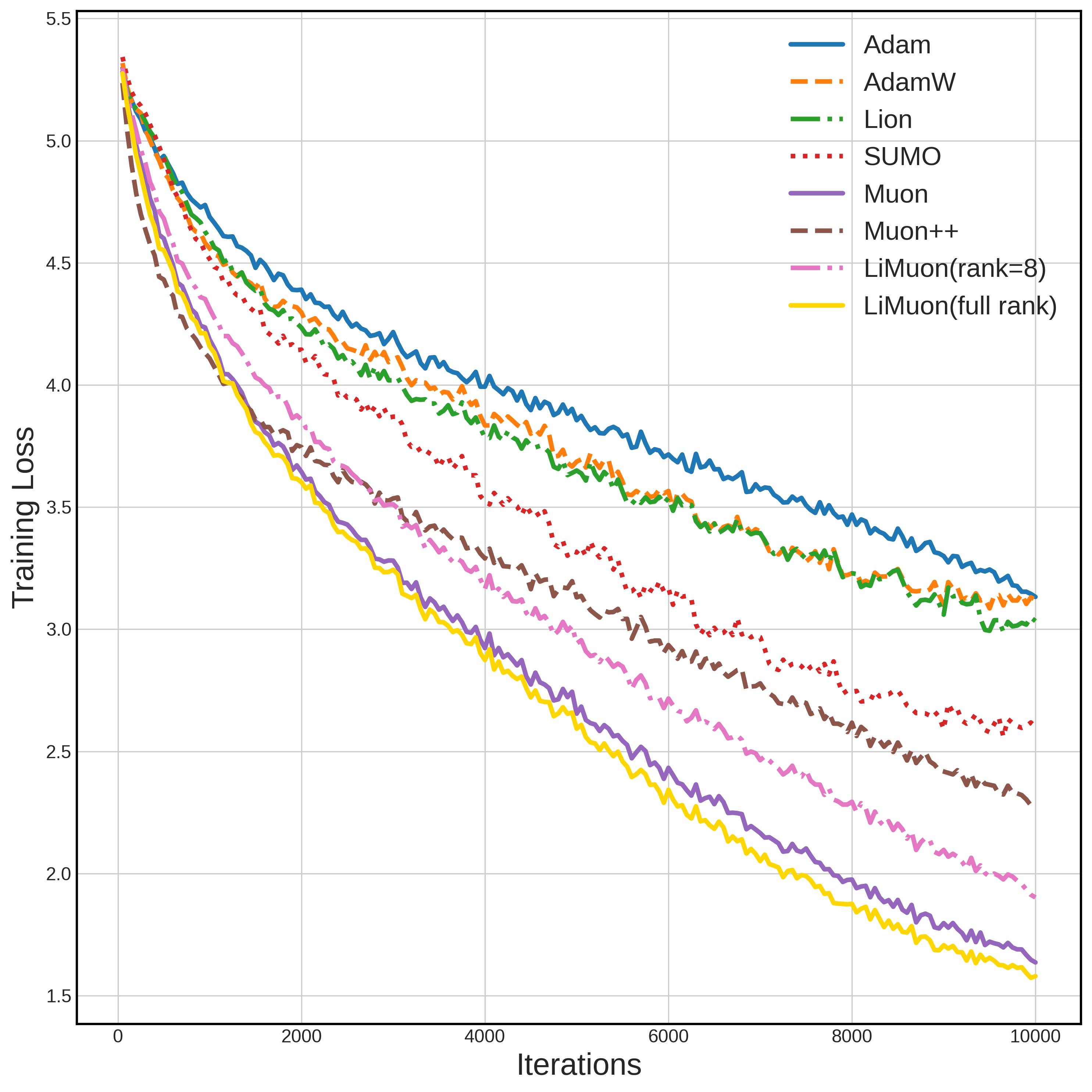}}
	\hfill
	\subfigure{\includegraphics[width=0.32\textwidth]{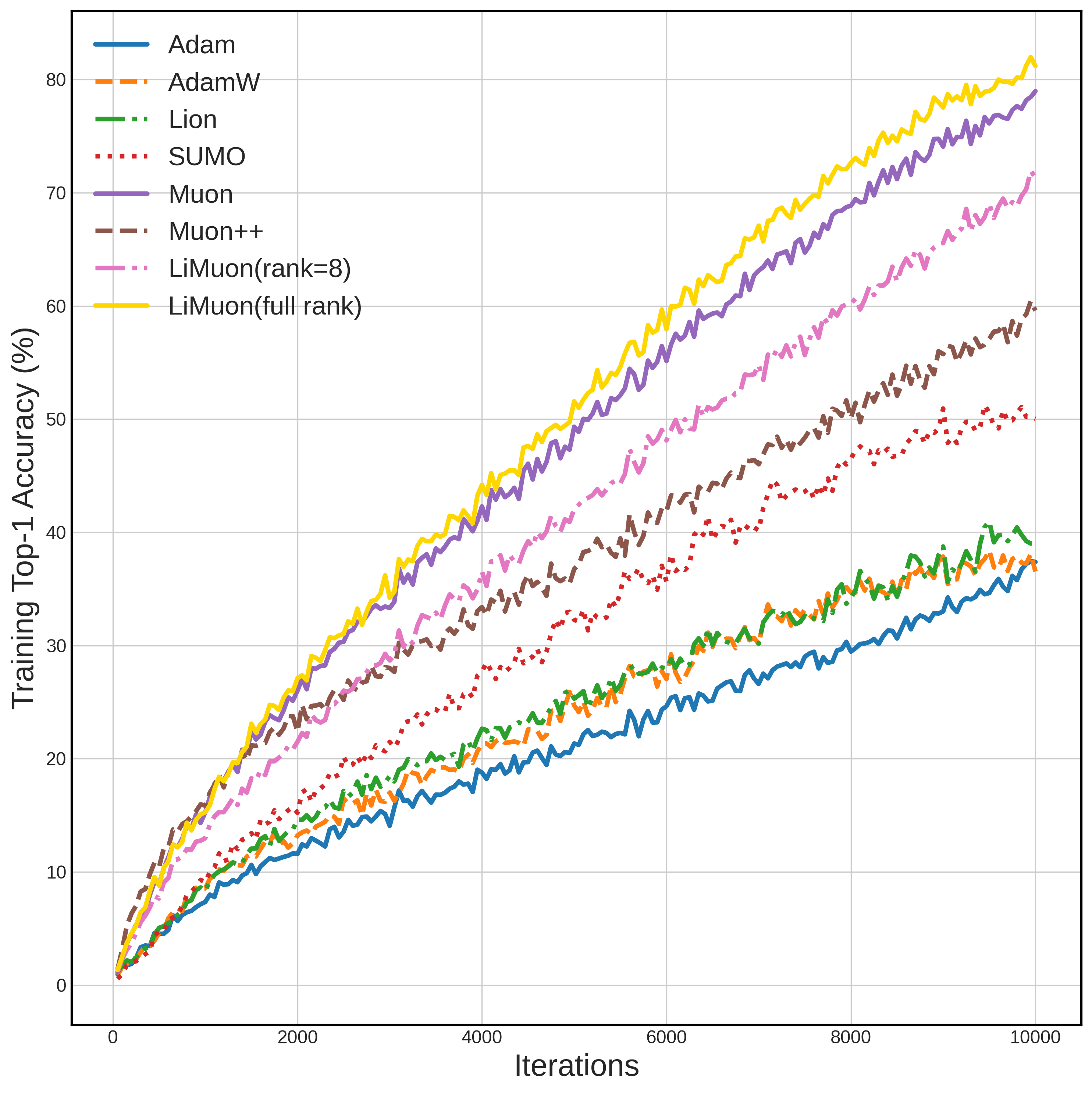}}
	\hfill
	\subfigure{\includegraphics[width=0.32\textwidth]{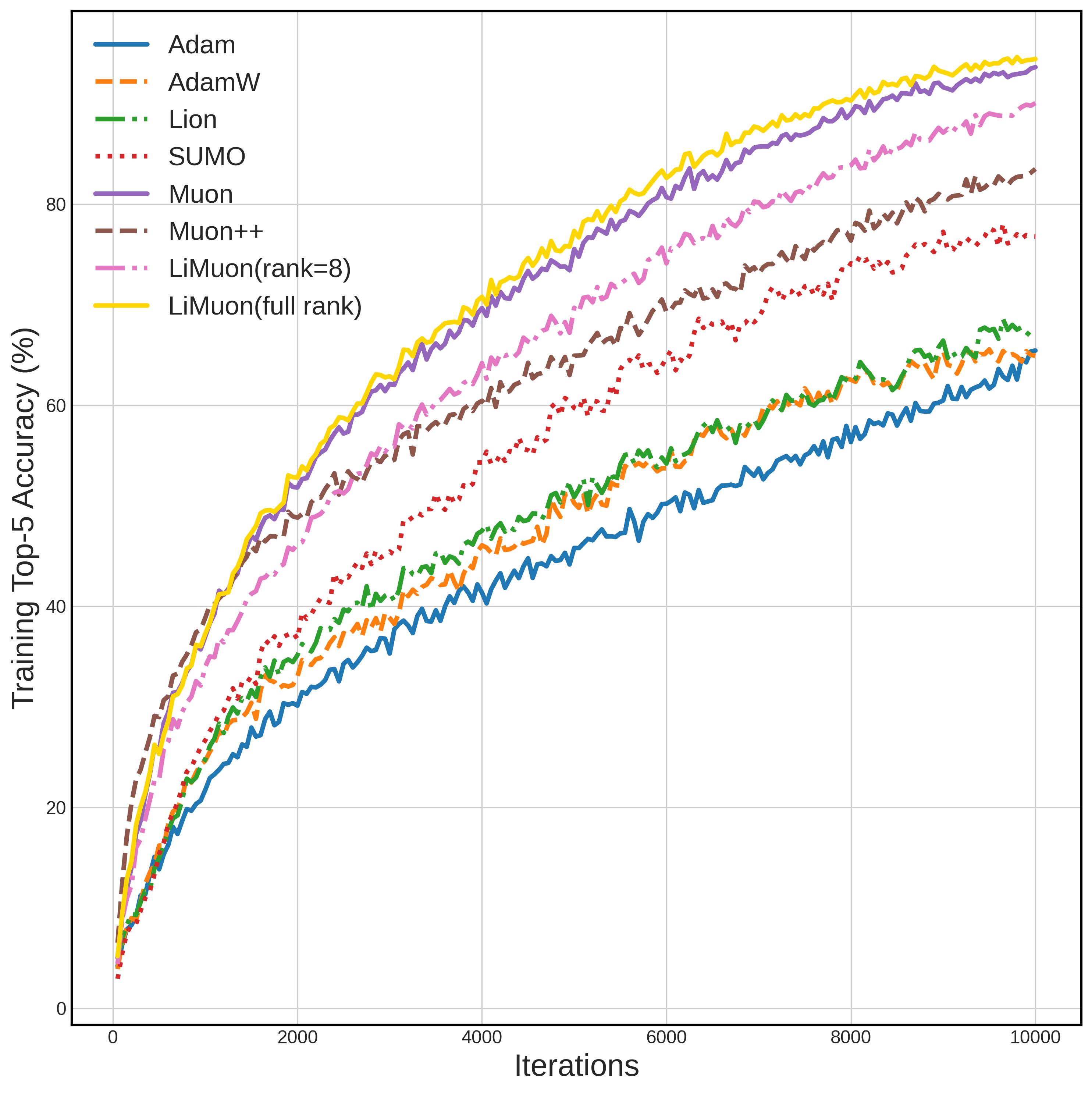}}
	\hfill
	\subfigure{\includegraphics[width=0.32\textwidth]{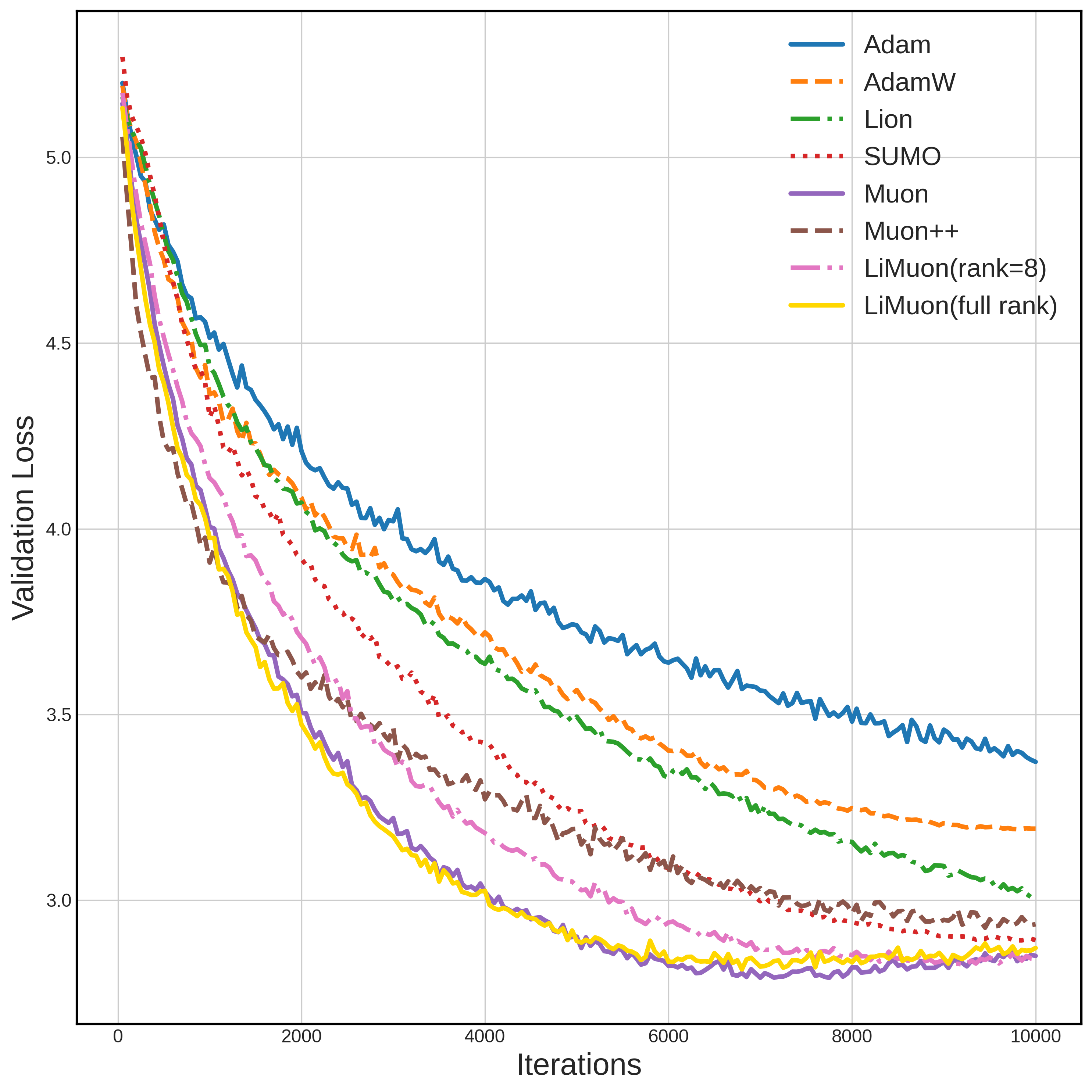}}
	\hfill
	\subfigure{\includegraphics[width=0.32\textwidth]{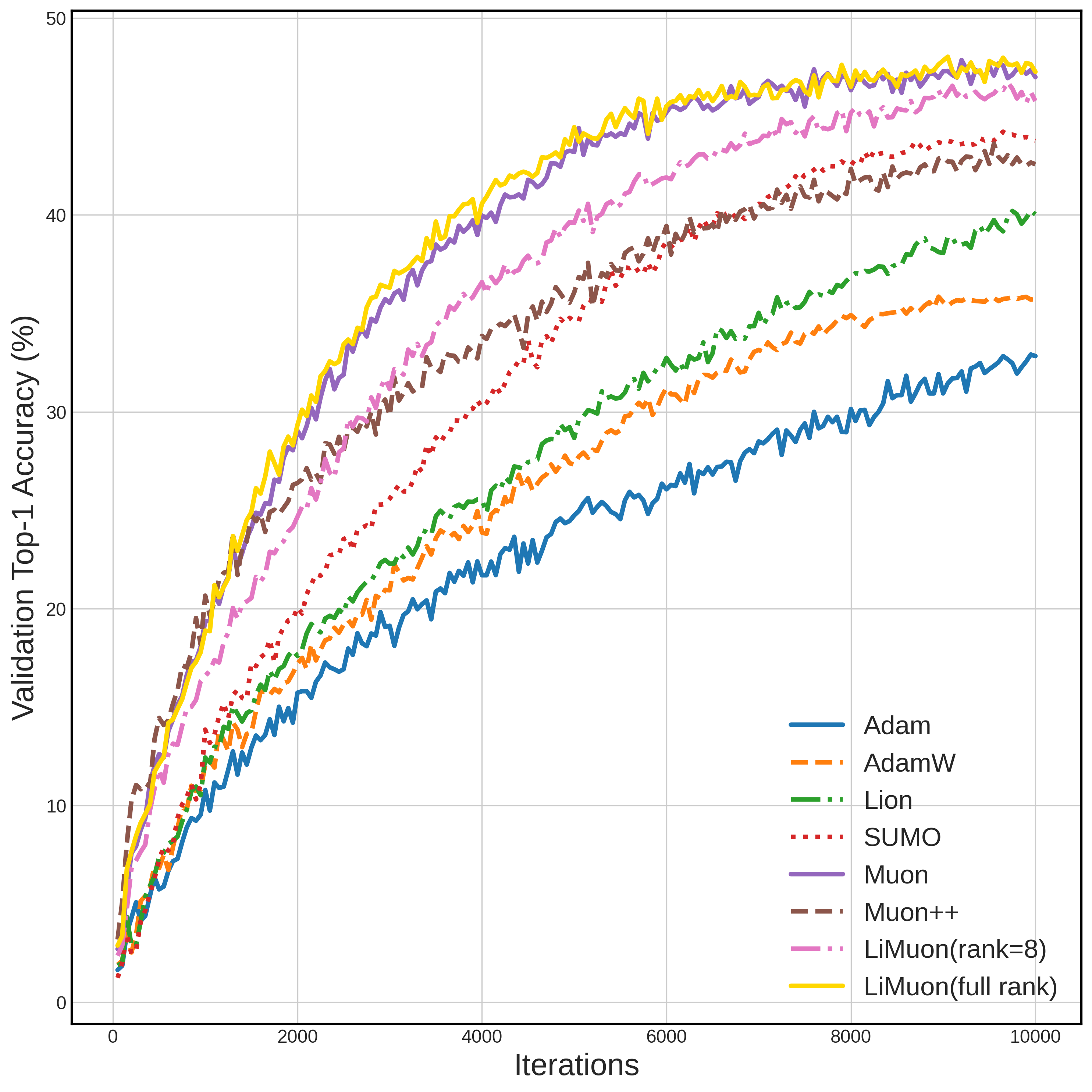}}
	\hfill
	\subfigure{\includegraphics[width=0.32\textwidth]{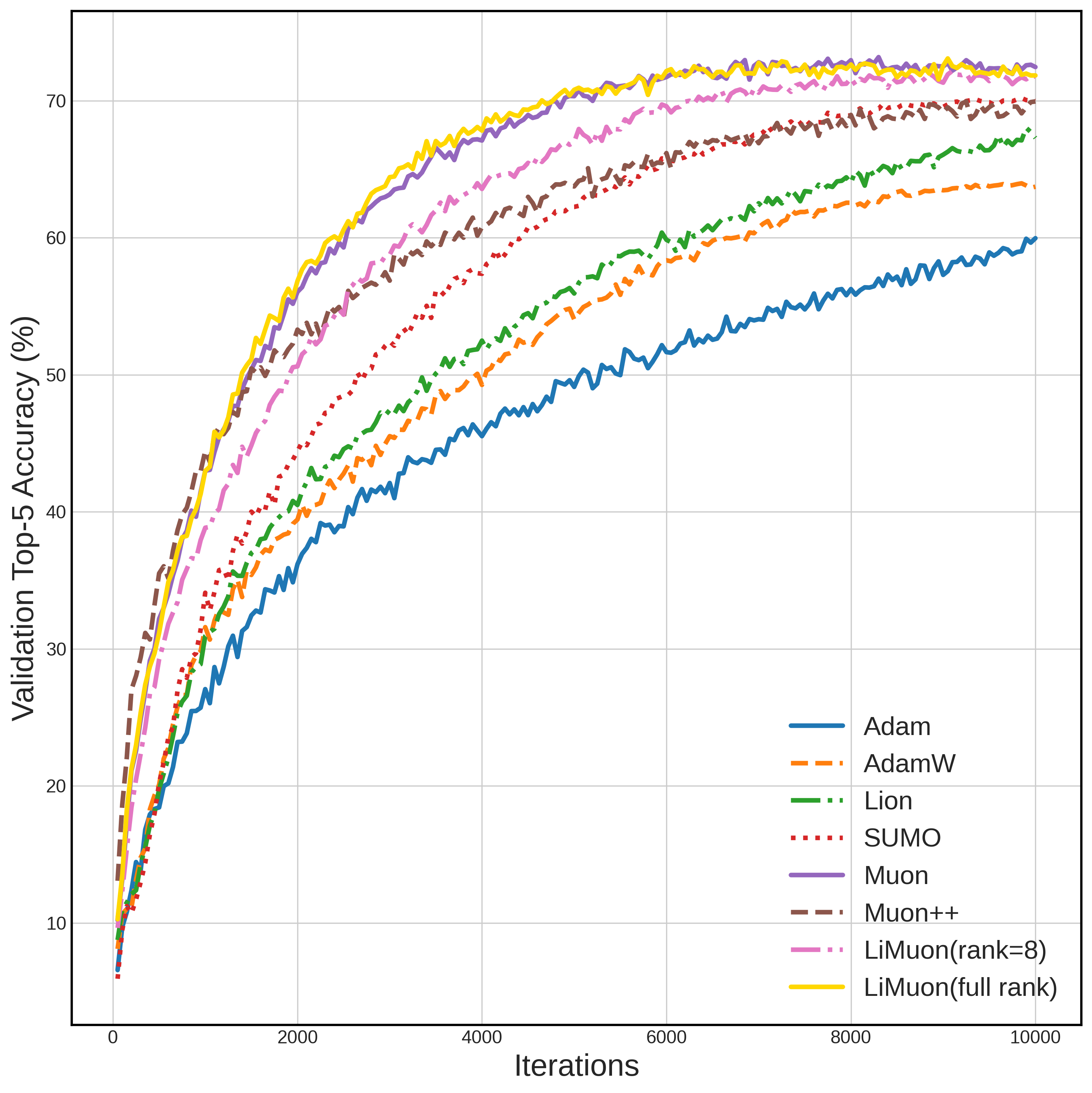}}
	\hfill
	\caption{Performance comparison at ViT }
	\label{fig:vit1}
\end{figure*}

In this section, we provide hyper-parameters used in the experiments, and 
some additional experimental results. Tables~\ref{tab:mamba-para},~\ref{tab:qwen-para} and~\ref{tab:vit-para} show the hyper-parameters used for Mamba-130M, Qwen2.5-0.5B and ViT, respectively. Table~\ref{tab:para-sr} provides hyper-parameter search ranges for all algorithms. Figure~\ref{fig:vit1} shows performance comparison at ViT.

\begin{table}[t]
	\centering
	\caption{Hyperparameters for Mamba-130M.}
	\label{tab:mamba-para}
	\small
	\begin{tabular}{l|c|c|c|c|c|c}
		\hline
		Optimizer & LR & $(\beta_1,\beta_2)$ & $\beta_t(momentum)$ & Weight decay & Rank $\hat r$ & Other Params \\
		\hline
		Adam   & 1e-3  & (0.9, 0.999) & --   & 0    & -- & -- \\
		AdamW  & 1e-3  & (0.9, 0.999) & --   & 0.01 & -- & -- \\
		Lion   & 3e-4  & (0.9, 0.99)  & --   & 0.03 & -- & -- \\
		\hline
		Muon   & 0.02  & --           & 0.95 & 0.01   & -- & -- \\
		Muon++ & 0.02  & --           & 0.95($\mu$)   & 0.001   & -- & clipping $M=3$ \\
		SUMO   & 0.002 & --           & 0.95   & 0.01   & 8  & $K=200$, $\gamma=1.1$ \\
		\hline
		LiMuon        & 0.02 & -- & 0.05 & 0.01 & 4    & $s=5$ \\
		LiMuon        & 0.02 & -- & 0.05 & 0.01 & 8    & $s=5$ \\
		LiMuon        & 0.02 & -- & 0.05 & 0.01 & 16   & $s=5$ \\
		LiMuon (full) & 0.02 & -- & 0.05 & 0.01 & full & -- \\
		\hline
	\end{tabular}
\end{table}


\begin{table}[t]
	\centering
	\caption{Hyperparameters for Qwen2.5-0.5B.}
	\label{tab:qwen-para}
	\small
	\begin{tabular}{l|c|c|c|c|c|c}
		\hline
		Optimizer & LR & $(\beta_1,\beta_2)$ & $\beta_t(momentum)$ & Weight decay & Rank $\hat r$ & Other \\
		\hline
		Adam   & 6e-4  & (0.9, 0.95) & --   & 0    & -- & -- \\
		AdamW  & 6e-4  & (0.9, 0.95) & --   & 0.01 & -- & -- \\
		Lion   & 1e-4  & (0.9, 0.99) & --   & 0.05 & -- & -- \\
		\hline
		Muon   & 0.02  & --          & 0.95 & 0.01   & -- & -- \\
		Muon++ & 0.02  & --          & 0.95($\mu$)   & 0.005   & -- & clipping $M=6$ \\
		SUMO   & 0.003 & --          & 0.95   & 0.001   & 8  & $K=100$, $\gamma=1.1$ \\
		\hline
		LiMuon        & 0.02 & -- & 0.05 & 0.01 & 4    & $s=10$ \\
		LiMuon        & 0.02 & -- & 0.05 & 0.01 & 8    & $s=10$ \\
		LiMuon        & 0.02 & -- & 0.05 & 0.01 & 16   & $s=10$ \\
		LiMuon (full) & 0.02 & -- & 0.05 & 0.01 & full & -- \\
		\hline
	\end{tabular}
\end{table}

\begin{table}[t]
	\centering
	\caption{Hyperparameters for ViT.}
	\label{tab:vit-para}
	\small
	\begin{tabular}{l|c|c|c|c|c|c}
		\hline
		Optimizer & LR & $(\beta_1,\beta_2)$ & $\beta_t(momentum)$ & Weight decay & Rank $\hat r$ & Other \\
		\hline
		Adam   & 3e-4  & (0.9, 0.999) & --   & 0 & -- & -- \\
		AdamW  & 3e-4  & (0.9, 0.999) & --   & 0.05 & -- & -- \\
		Lion   & 6e-5  & (0.9, 0.99)  & --   & 0.05 & -- & -- \\
		\hline
		Muon   & 0.005 & --           & 0.95 & 0.01   & -- & -- \\
		Muon++ & 0.01  & --           & 0.95($\mu)$   & 0.001   & -- & clipping $M=3$ \\
		SUMO   & 0.002 & --           & 0.95   & 0.05   & 8  & $K= 200$, $\gamma=1.1$ \\
		\hline
		LiMuon        & 0.005 & -- & 0.05 & 0.05 & 4    & $s=5$ \\
		LiMuon        & 0.005 & -- & 0.05 & 0.05 & 8    & $s=5$ \\
		LiMuon        & 0.005 & -- & 0.05 & 0.01 & 16   & $s=5$ \\
		LiMuon (full) & 0.005 & -- & 0.05 & 0.01 & full & -- \\
		\hline
	\end{tabular}
\end{table}

\begin{table}[t]
	\centering
	\caption{Hyperparameter Search Ranges.}
	\label{tab:para-sr}
	\small
	\begin{tabular}{lc}
		\hline
		Hyperparameter & Search range \\
		\hline
		\quad Learning rate \emph{(Adam / AdamW)}       & \{1e-4, 3e-4, 6e-4, 1e-3\} \\
		\quad Learning rate \emph{(Lion)}                & \{3e-5, 6e-5, 1e-4, 3e-4,1e-3\} \\
		\quad Learning rate \emph{(Muon / Muon++ / LiMuon)} & \{2e-3, 5e-3, 1e-2, 2e-2,5e-2\} \\
		\quad Learning rate \emph{(SUMO)}               & \{5e-4, 1e-3, 2e-3, 3e-3, 5e-3\} \\
		\hline
		Weight decay                                     & \{0, 1e-3, 5e-3, 1e-2, 3e-2, 5e-2\} \\
		\hline
		Gradient clipping threshold $M$ \emph{(Muon++)} & \{1, 2, 3, 4, 6\} \\
		\hline
		Rank $\hat{r}$ \emph{(LiMuon)}           & \{4, 8, 16\} \\
		Step size $s$ \emph{(LiMuon)}                    & \{5, 10\} \\
		Update period $K$ \emph{(SUMO)}                  & \{50, 100, 200\} \\
		\hline
	\end{tabular}
\end{table}

\end{document}